# MACHINE LEARNING FOR SCIENTIFIC VISUALIZATION: ENSEMBLE DATA ANALYSIS

HAMID GADIROV

Cover: The illustration shows a 3D rendering of a Nyx simulation sample from cosmological structure formation. It visualizes the density field of dark matter in an expanding universe, produced with the `yt` toolkit.

Machine Learning
for Scientific Visualization:
Ensemble Data Analysis

Hamid Gadirov
PhD Thesis

university of groningen

# Machine Learning for Scientific Visualization: Ensemble Data Analysis

**PhD thesis**

to obtain the degree of PhD at the
University of Groningen
on the authority of the
Rector Magnificus Prof. J.M.A. Scherpen
and in accordance with
the decision by the College of Deans.

This thesis will be defended in public on

Tuesday 14 October 2025 at 16.15 hours

by

**Hamid Gadirov**

born on 8 July 1996
in Baku, Azerbaijan

**Supervisor**
Prof. J. B. T. M. Roerdink

**Co-supervisor**
Dr. S. D. Frey

**Assessment Committee**
Prof. C. Wang
Prof. A. C. Telea
Prof. J. Kosinka

*I am not what happened to me, I am what I choose to become.*

— Carl Jung

# ABSTRACT

Scientific simulations and experimental measurements produce vast amounts of spatio-temporal data, but extracting meaningful insights from such data remains a challenge due to its high dimensionality, complex structures, and missing information. Traditional analysis techniques often struggle with these issues, motivating the need for more robust, data-driven approaches. This dissertation explores deep learning methodologies to enhance the analysis and visualization of spatio-temporal scientific ensembles, focusing on dimensionality reduction, flow estimation, and temporal interpolation. First, we address the challenge of high-dimensional data representation by investigating autoencoder-based dimensionality reduction for scientific ensembles. We evaluate the stability of projection metrics under partial labeling and introduce a Pareto-efficient selection strategy to identify optimal autoencoder variants, ensuring expressive and reliable low-dimensional embeddings. Next, we present FLINT, a deep learning model designed for high-quality flow estimation and temporal interpolation in both flow-supervised and flow-unsupervised settings. FLINT reconstructs missing velocity fields and generates high-fidelity temporal interpolants for scalar fields across 2D+time and 3D+time ensembles, without requiring domain-specific assumptions or extensive fine-tuning. To further improve adaptability and generalization, we introduce HyperFLINT, a novel hypernetwork-based approach that dynamically conditions on simulation parameters to estimate flow fields and interpolate scalar data. This parameter-aware adaptation enables more accurate reconstructions across diverse scientific domains, even in cases of sparse or incomplete data. By addressing key challenges in scientific data analysis, this dissertation advances deep learning techniques for scientific visualization, providing scalable, adaptable, and high-quality solutions for interpreting complex spatio-temporal ensembles.

# SAMENVATTING

Wetenschappelijke simulaties en experimentele metingen produceren enorme hoeveelheden spatiotemporele data, maar het afleiden van zinvolle inzichten uit dergelijke data blijft een uitdaging vanwege de hoge dimensionaliteit, complexe structuren en ontbrekende informatie. Traditionele analysetechnieken worstelen vaak met deze problemen, wat de behoefte aan robuustere, datagestuurde benaderingen motiveert. Dit proefschrift onderzoekt *deep learning*-methodologieën om de analyse en visualisatie van spatiotemporele wetenschappelijke ensembles te verbeteren, met de nadruk op dimensionaliteitsreductie, stromingsschatting en temporele interpolatie. Ten eerste pakken we de uitdaging van hoogdimensionale datarepresentatie aan door autoencoder-gebaseerde dimensionaliteitsreductie voor wetenschappelijke ensembles te onderzoeken. We evalueren de stabiliteit van projectiemetrieken onder gedeeltelijke labeling en introduceren een Pareto-efficiënte selectiestrategie om optimale autoencodervarianten te identificeren, wat zorgt voor expressieve en betrouwbare laagdimensionale inbeddingen. Vervolgens presenteren we FLINT, een *deep learning*-model dat is ontworpen voor hoogwaardige stromingsschatting en temporele interpolatie in omstandigheden zowel met als zonder stromings-supervisie. FLINT reconstrueert ontbrekende snelheidsvelden en genereert zeer nauwkeurige temporele interpolanten voor scalaire velden in 2D+tijd- en 3D+tijd-ensembles, zonder dat domeinspecifieke aannames of uitgebreide fijnafstemming nodig zijn. Om de aanpasbaarheid en generalisatie verder te verbeteren, introduceren we HyperFLINT, een nieuwe, op hypernetwerken gebaseerde aanpak die dynamische voorwaarden stelt aan simulatieparameters om stromingsvelden te schatten en scalaire data te interpoleren. Deze parameter-inclusieve aanpassing maakt nauwkeurigere reconstructies mogelijk in diverse wetenschappelijke domeinen, zelfs in gevallen van schaarse of onvolledige data. Door belangrijke uitdagingen in wetenschappelijke data-analyse aan te pakken, bevordert dit proefschrift *deep learning*-technieken voor wetenschappelijke visualisatie en biedt het schaalbare, aanpasbare en hoogwaardige oplossingen voor de interpretatie van complexe spatiotemporele ensembles.

# PUBLICATIONS

This thesis is based on the following publications:

- H. Gadirov, G. Tkachev, T. Ertl, and S. Frey. Evaluation and selection of autoencoders for expressive dimensionality reduction of spatial ensembles. In *International Symposium on Visual Computing. Springer*, 2021, pp. 222–234.
- H. Gadirov, J. B. T. M. Roerdink, and S. Frey. FLINT: Learning-based flow estimation and temporal interpolation for scientific ensemble visualization. *IEEE Transactions on Visualization and Computer Graphics* (2025).
- H. Gadirov, Qi Wu, D. Bauer, K.-L. Ma, J. B. T. M. Roerdink, and S. Frey. HyperFLINT: Hypernetwork-based Flow Estimation and Temporal Interpolation for Scientific Ensemble Visualization. *Computer Graphics Forum* (2025).

Additionally, contributed as a co-author to the following projects:

- D. Bauer, Qi Wu, H. Gadirov, and K.-L. Ma. GSCache: Real-Time Radiance Caching for Volume Path Tracing using 3D Gaussian Splatting. *IEEE Transactions on Visualization and Computer Graphics* (2025).
- Z. Yin, H. Gadirov, J. Kosinka, and S. Frey. ENTIRE: Learning-based Volume Rendering Time Prediction. *arXiv preprint arXiv:2501.2501* (2025).

# CONTENTS

# 1 INTRODUCTION

Advancements in computing and data acquisition technologies have led to a rapid increase in the size and complexity of scientific datasets. From high-resolution simulations in fluid dynamics to large-scale astrophysical models and experimental imaging, scientific data often consist of measurements or simulation outputs defined over spatial or spatio-temporal domains, typically represented as multidimensional fields such as scalar, vector, or tensor quantities distributed over 2D or 3D grids evolving in time.

In many scientific applications, individual datasets are not sufficient to capture variability or uncertainty. Instead, researchers rely on scientific ensembles—collections of simulation runs or repeated experimental measurements—where each member represents a variation in input parameters, initial conditions, or stochastic influences. These ensembles are critical for understanding complex natural and engineered systems, enabling sensitivity analysis, uncertainty quantification, and robust predictions.

While ensembles offer significant potential for new scientific insights, they also present unique challenges, including handling vast data volumes, understanding parameter dependencies, and extracting meaningful patterns from high-dimensional structures (Wang et al., 2018). A fundamental challenge in ensemble analysis arises from the nature of scientific data itself: scientific ensembles exhibit spatial, temporal, and parameter-dependent variability. Each ensemble member evolves over time, making it difficult to track changes, identify trends, and draw meaningful comparisons. Traditional visualization techniques and statistical methods struggle to effectively process these datasets due to their scale and complexity. To address these limitations, machine learning (ML) has emerged as a powerful tool, enabling automated analysis, dimensionality reduction, and reconstruction of missing or incomplete information.

Dimensionality reduction (DR) techniques, such as principal component analysis (PCA), t-distributed stochastic neighbor embedding (t-SNE), and uniform manifold approximation and projection (UMAP), have been widely applied to scientific datasets to reveal patterns and similarities across ensemble members (Kehrer and Hauser, 2013). However, these conventional techniques often struggle to preserve the complex spatial and temporal structures inherent in scientific ensembles—especially in high-dimensional and sparse data contexts (Liu et al., 2016). To address these limitations, deep learning approaches—particularly autoencoders—have emerged as powerful alter-natives, capable of learning expressive and compact representations that better retain semantic and structural information. Autoencoders have also shown promise in related tasks, such as predicting volume rendering times (Yin et al., 2025), where learned embeddings capture key features to estimate computational cost. These applications highlight the versatility and growing role of autoencoder architectures in scientific visualization.

To address the challenges of preserving structure and generalizing across sparse, high-dimensional ensemble data, we propose a tailored dimensionality reduction framework based on autoencoder architectures. Building upon our prior evaluation of DR techniques in scientific visualization (Gadirov et al., 2021), we conduct a systematic study of multiple autoencoder variants and analyze their impact on projection quality across a range of spatial ensemble datasets. We explore how autoencoders balance compression and reconstruction while maintaining meaningful spatial structures, and we introduce a guided selection strategy based on partially labeled data. This enables robust, expressive low-dimensional embeddings and supports Pareto-efficient model selection, addressing key questions of model stability and generalization across ensemble data.

Technological advancements now make it possible to capture and simulate time-dependent processes at unprecedented resolutions, leading to vast spatio-temporal datasets that enable detailed exploration of parameter variations and stochastic effects. However, the sheer volume of generated data often exceeds storage capacities, necessitating selective preservation of timesteps or variables (Childs et al., 2019). Additionally, experimental data is often constrained by measurement limitations, preventing full observation of all relevant dynamics. Reconstructing missing information post-hoc is therefore critical for enabling effective visualization and analysis.

Spatio-temporal ensemble simulations are increasingly used across various scientific domains, including fluid dynamics, astrophysics, climate modeling, and biomedical imaging. These simulations often consist of multiple runs with varying parameters, generating high-dimensional datasets that require specialized methods for processing and visualization (Wang et al., 2018). While preserving all timesteps and variables in a high-resolution simulation would be ideal, storage and computational constraints often necessitate aggressive data reduction (Childs et al., 2019). Similarly, experimental datasets are limited by what can be physically measured or recorded, further complicating analysis. A key challenge in ensemble data visualization is reconstructing missing information, such as scalar fields (e.g., density, temperature) or vector fields (e.g., velocity, optical flow), to improve the completeness and interpretability of the data.

To address these challenges, in this thesis we propose FLINT (Gadirov et al., 2025a) (learning-based flow estimation and temporal interpola-tion), a deep learning-based framework for reconstructing scalar fields and estimating flow fields in 2D and 3D scientific datasets. FLINT introduces a student-teacher architecture designed to produce high-quality density interpolation and flow estimation. This is particularly valuable for applications such as flow visualization (Jänicke et al., 2011), optimal timestep selection (Frey and Ertl, 2017), and ensemble member comparison (Tkachev et al., 2021). By leveraging machine learning, FLINT enables temporal super-resolution and optical flow estimation, even when flow data is unavailable due to storage limitations or experimental constraints.

However, a method like FLINT does not explicitly incorporate ensemble parameters, limiting its ability to generalize across different simulation conditions. To overcome this limitation, we propose HyperFLINT (Gadirov et al., 2025b), an extension that integrates hypernetworks to dynamically incorporate ensemble parameters. By generating weights conditioned on simulation parameters, HyperFLINT adapts to diverse simulation conditions, improving both flow estimation and scalar-field interpolation while enabling parameter space exploration. Unlike conventional methods, which often require retraining for new datasets, HyperFLINT offers a flexible and generalizable approach for large-scale scientific ensembles.

## 1.1 RESEARCH QUESTIONS

This dissertation addresses the following research questions:

- **RQ1:** How can autoencoders be leveraged for effective dimensionality reduction while preserving essential spatial and temporal structures in scientific ensembles?
- **RQ2:** How can deep learning methods estimate flow fields and learn the underlying dynamics of physical processes from spatiotemporal ensembles, especially in the absence of explicit flow supervision?
- **RQ3:** How can we design deep learning models that enable experimental or simulation parameter-aware learning, and what impact does this dynamic adaptation have on flow estimation and interpolation quality?
- **RQ4:** What underlying patterns can deep learning reveal in scientific ensemble data that may be overlooked by conventional data analysis and visualization techniques?

## 1.2 SUMMARY OF CONTRIBUTIONS

This dissertation presents novel methodologies for analyzing and visualizing spatio-temporal scientific ensembles using deep learning. Our main contributions include:

- **Autoencoder-based Dimensionality Reduction:** We conduct a comprehensive study of several autoencoder variants for expressive dimensionality reduction across diverse scientific ensembles. We evaluate the stability of projection metrics under partially labeled data and propose a Pareto-efficient selection strategy to identify the most suitable autoencoder variant for a given dataset.
- **FLINT—Flow Estimation and Temporal Interpolation:** We introduce a deep learning-based method for high-quality flow estimation in scenarios where (1) flow information is partially available (*flow-supervised*, e.g., simulations where flow is partly omitted due to storage constraints) or (2) entirely unavailable (*flow-unsupervised*, e.g., experimental image sequences captured via cameras that measure scalar fields such as density or luminance). Our method also produces high-fidelity temporal interpolants for scalar fields, such as density or luminance, and operates effectively on both 2D+time and 3D+time ensembles without requiring domain-specific assumptions, complex pre-training, or fine-tuning on simplified datasets.
- **HyperFLINT—Hypernetwork-Based Flow Estimation and Interpolation:** We propose the first approach that employs a hypernetwork to adaptively estimate flow fields and generate state-of-the-art temporal interpolations of density fields for scientific ensembles. By dynamically conditioning on simulation parameters, HyperFLINT generates parameter-aware weights, enhancing model quality and performance by adapting to varying simulation conditions. This results in significantly improved flow estimation and scalar field interpolation, even in scenarios with sparse or incomplete data, making the method highly versatile for analyzing complex spatio-temporal ensembles across diverse scientific domains.

These contributions collectively advance the field of scientific visualization and deep learning for ensemble data analysis, offering robust and flexible solutions for flow estimation, interpolation, and parameter-driven exploration.

## 1.3 OUTLINE

This manuscript is structured as follows:

In the upcoming Chapter 2, we provide background, introducing the fundamental concepts and methodologies underlying the techniques explored throughout this work. This includes discussions on neural networks, convolutional architectures, autoencoders, hypernetworks, dimensionality reduction techniques, and ensemble visualization. This chapter lays the foundation for addressing **RQ1**, **RQ2**, **RQ3**, and **RQ4** by exploring how deep learning methods extract meaningful representations from scientific ensemble data.

Following this, in Chapter 3, we focus on evaluating and selecting autoencoder configurations for expressive dimensionality reduction of spatial ensembles. We systematically assess how different autoencoder variants impact the quality of low-dimensional projections and propose a guided selection strategy based on partially labeled data. This chapter directly addresses **RQ1** and **RQ4** by exploring how autoencoders balance compression and reconstruction quality while preserving essential structures in ensemble data.

Next, in Chapter 4, we investigate how convolutional networks, incorporating autoencoder architectures, can be leveraged for flow estimation and temporal interpolation in 2D and 3D ensemble datasets. This chapter presents FLINT, a model designed to jointly predict flow fields and interpolate scalar fields, improving temporal resolution while maintaining structural consistency. This chapter answers **RQ2** by demonstrating how convolutional networks estimate flow fields and learn the governing dynamics of physical processes from ensemble simulations. Furthermore, it contributes to **RQ4** by uncovering how the model implicitly learns spatial and temporal patterns that may be overlooked by conventional methods.

Building upon FLINT, we introduce HyperFLINT in Chapter 5, an extension that integrates hypernetworks to further improve accuracy and expand the model's capabilities. Beyond flow estimation and interpolation, HyperFLINT introduces a parameter-aware adaptation mechanism, allowing the model to dynamically adjust based on ensemble parameters. This not only enhances prediction accuracy but also enables parameter space exploration, providing deeper insights into scientific simulations. By leveraging hypernetworks, HyperFLINT addresses limitations in previous approaches, making it a powerful tool for analyzing complex ensemble datasets. This chapter directly addresses **RQ3** by demonstrating how hypernetworks enable simulation parameter-aware learning and enhance model performance. Specifically, we show that HyperFLINT, through the use of hypernetworks, improves both flow estimation and temporal interpolation, enabling more accurate and generalizable reconstructions across diverse scientific domains. It also supports **RQ4** by revealing correlations between ensemble parameters and the dynamics of the underlying data.

Finally, in Chapter 6, we conclude with a summary of research results and contributions, reflecting on the key findings of this work and out-lining potential directions for future research. This chapter discusses the broader implications of our findings, addressing **RQ4**, particularly in terms of the underlying patterns deep learning can reveal, the role of data diversity in ensemble visualization, and the generalization of deep learning models across scientific domains.

# 2 BACKGROUND

*In this chapter, we introduce the fundamental concepts and methodologies underlying the techniques described in this thesis. We begin by explaining the foundations of neural networks, detailing their structure and learning mechanisms. Next, we provide an overview of convolutional neural networks (CNNs) and their role in feature extraction and hierarchical representation learning. We then introduce transformers, discussing their impact in large language models (LLMs) and vision transformers (ViTs), highlighting their advantages in capturing long-range dependencies and contextual information. Following this, we explore different types of autoencoders, including variational autoencoders (VAEs), and their applications in dimensionality reduction and data reconstruction. We also discuss explainable AI techniques, which enhance the interpretability of deep learning models, ensuring that complex architectures can be understood and trusted in scientific applications. Additionally, we discuss student-teacher learning, a paradigm in which a smaller model (student) learns from a larger, pre-trained model (teacher), improving efficiency and generalization. Next, we examine hypernetworks, which dynamically generate weights for other networks, enabling increased adaptability in deep learning models. We then shift our focus to volume rendering, a crucial technique for visualizing three-dimensional scalar fields, and discuss its significance in scientific visualization. Finally, we provide insights into ensemble visualization, explaining how techniques such as flow estimation and temporal interpolation can be leveraged to analyze complex spatiotemporal datasets. These fundamental concepts set the stage for the methods developed and evaluated in the later chapters.*

## 2.1 NEURAL NETWORKS

Artificial neural networks (ANNs) are computational models inspired by biological neuronal networks. They consist of interconnected artificial neurons, where the strength of each connection is determined by a weight. A neuron processes its inputs by computing a weighted sum and applying an activation function, which can be linear or nonlinear, depending on the problem at hand (Demuth et al., 2014). Neurons are organized into layers, each performing distinct transformations on the input data.

Information flows through an ANN in two main phases. During the forward propagation phase, input data passes through successive layers until an output is produced. If the network is trained using supervised learning, the output is compared to known target values, and an error is computed—typically using loss functions such as mean squared error or cross-entropy (Bishop, 2006). The backward propagation (first introduced by (Rosenblatt, 1962)) phase then follows, where the error is propagated backward through the network to adjust the weights, optimizing performance over successive training iterations (epochs).

By iteratively updating its weights, a neural network learns to perform specific tasks without explicit task-specific programming. This ability to generalize from examples makes neural networks powerful tools for a wide range of applications, including pattern recognition, classification, and predictive modeling. The gradient of the loss function $\zeta$ (which computes the error mentioned before) with respect to the weights looks like:

$$\frac{\partial \zeta}{\partial W_{l,ij}} = \frac{\partial \zeta}{\partial z_{l+1,i}} \frac{\partial z_{l+1,i}}{\partial W_{l,ij}}, \tag{2.1}$$

where $W_{l,ij}$ denotes the weight in layer $l$ for the node with output index $j$ and input $i$, and $z_{l+1,i}$ is the input to all neurons in layer $l + 1$. The update of the weights is based on basic gradient descent optimization (Bishop, 2006).

A deep neural network (DNN) is an ANN with multiple hidden layers, enabling it to learn complex, nonlinear (pattern) relations (Hinton et al., 2012a; Goodfellow et al., 2016). The number of hidden layers in a DNN can range from a few to thousands, depending on the complexity of the task. In a DNN, each layer processes and transforms its input data before passing it to the next layer. During training, each neuron adjusts its weights to minimize the error and improve performance on the given dataset. As mentioned earlier, this process follows the principles of supervised learning, where the model learns from labeled data through forward and backward propagation. Figure 2.1 illustrates an example of a deep neural network with two hidden layers, demonstrating the hierarchical nature of feature extraction in DNNs.

The first algorithm for supervised learning of binary classifiers was the Perceptron, introduced by Rosenblatt in 1958 (Rosenblatt, 1958). This was a simple feedforward neural network with a single hidden layer, trained using the backpropagation algorithm. Later, this architecture was extended to Multilayer Perceptrons (MLP) (Pal and Mitra, 1992), which allowed for deeper networks and improved learning capabilities.

The rise of deep learning was driven by advancements in computational power, particularly the use of GPUs and supercomputers, which significantly accelerated training processes. As a result, deep neural networks (DNNs) became feasible, demonstrating remarkable performance in tasks such as image classification, video analysis, text processing, and natural language processing. In 2006, Geoffrey Hinton *et al.* (Hinton et al., 2006) introduced a method for learning high-level representations

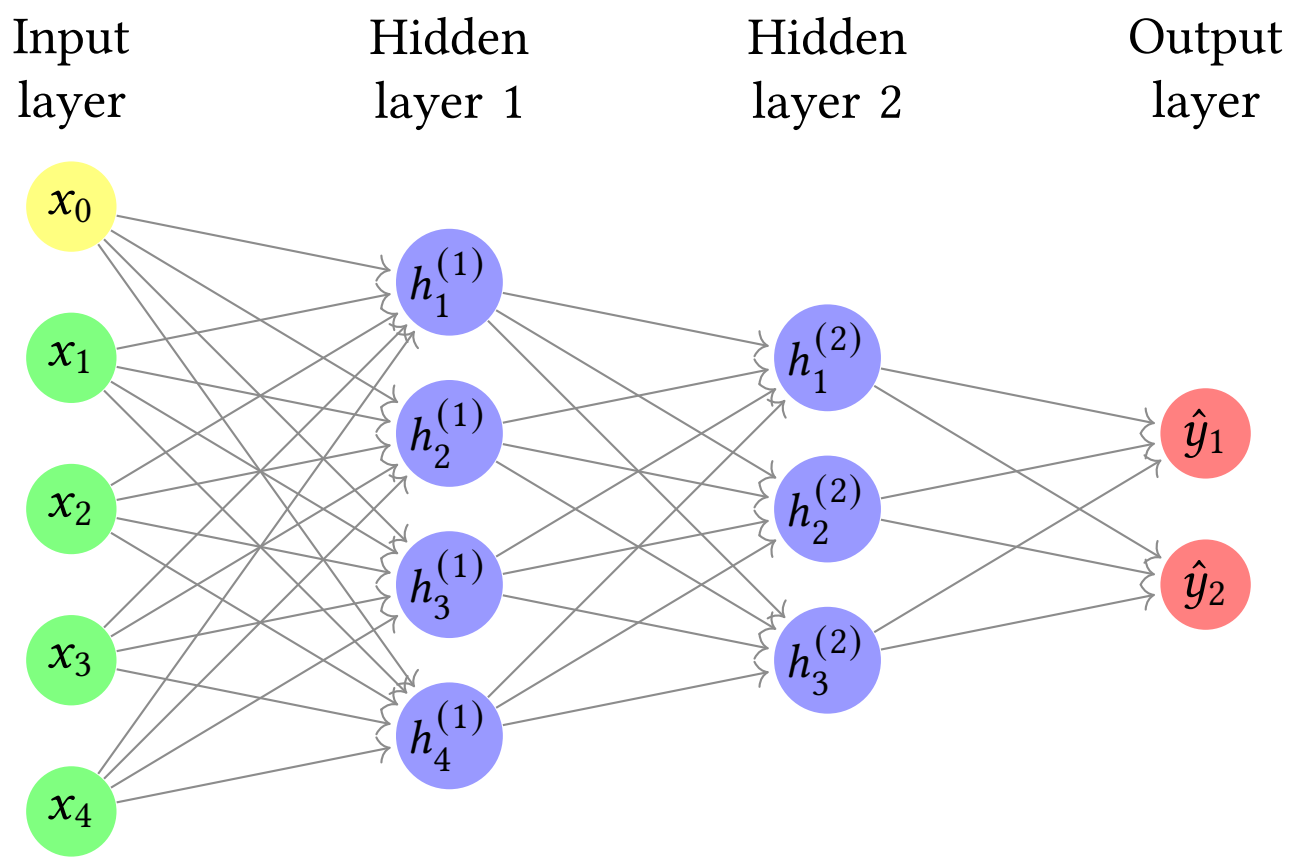

Figure 2.1: An example of a Deep Neural Network.

using restricted Boltzmann machines, further advancing deep learning research.

Neural networks come in various architectures, each designed for specific tasks:

- Feedforward Neural Networks (FNNs) – the simplest architecture where information moves in one direction, from input to output.
- Convolutional Neural Networks (CNNs) – specialized for image processing and feature extraction (Lawrence et al., 1997).
- Recurrent Neural Networks (RNNs) and Long Short-Term Memory (LSTMs) - designed for sequential data, such as time series and speech recognition (Hochreiter and Schmidhuber, 1997).
- Autoencoders (AEs) and Variational Autoencoders (VAEs) – used for dimensionality reduction, representation learning, and generative tasks.

These architectures can be used for both supervised and unsupervised learning (Sathya and Abraham, 2013). In the following sections, we will explore the specific types of neural networks relevant to these learning types in more detail.

## 2.2 CONVOLUTIONAL NEURAL NETWORKS

The structure of Convolutional Neural Networks (CNNs) is inspired by biological vision processes (Fukushima and Miyake, 1982) and closely resembles the organization of the visual cortex. CNNs are composed of multiple layers, primarily designed to automatically extract hierarchical features from input data.

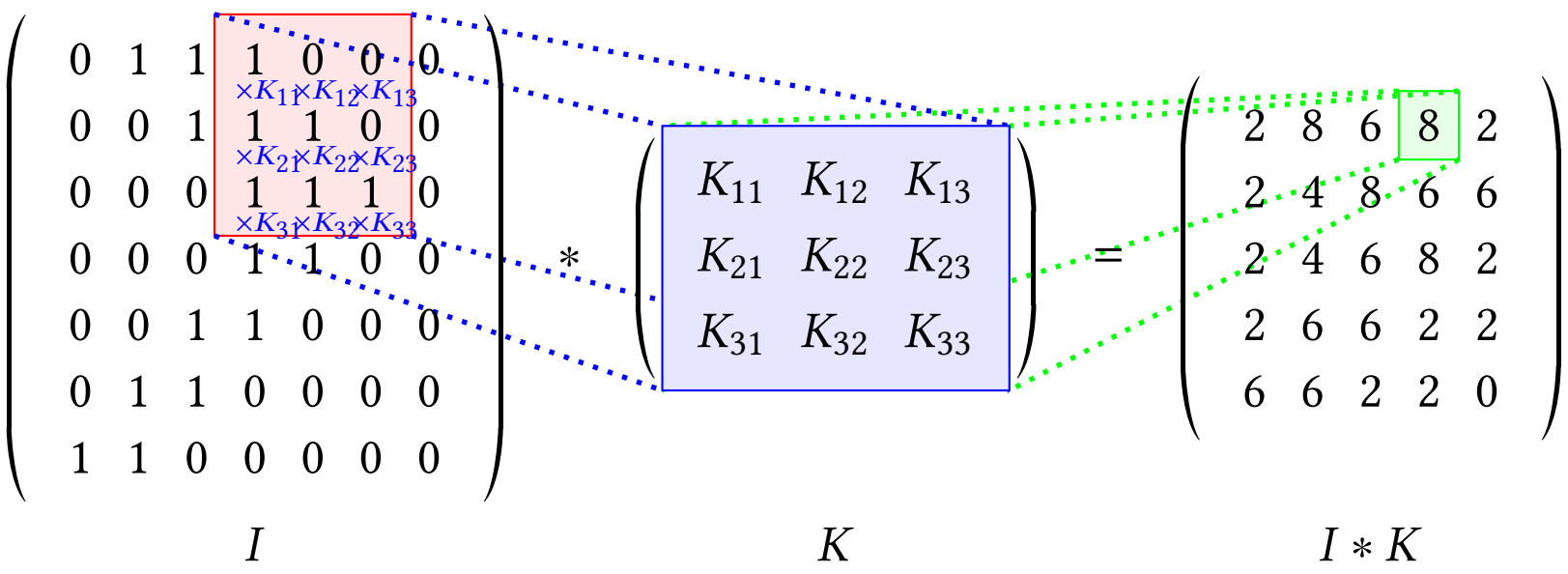

Figure 2.2: Mathematical operation of convolution: an image $I$ was convolved with a kernel $K$. During the convolution, each cell value of a red square is multiplied with a corresponding cell value of the blue square (kernel) and then all results are summed up. The final result is shown in green on the right.

A typical CNN consists of three main types of layers:

- Convolutional Layers – these apply convolution operations to extract local patterns and features, producing multiple feature maps. These layers detect essential characteristics such as edges, textures, and object shapes in the case of visual data.
- Pooling Layers – usually placed after convolutional layers, pooling reduces the spatial size of feature maps, improving computational efficiency and robustness to small spatial variations.
- Fully Connected Layers – after feature extraction, fully connected layers aggregate the learned representations and contribute to reducing sensitivity to spatial positioning, making CNNs effective for classification tasks.

An illustration of the convolution operation is shown in Figure 2.2.

To introduce non-linearity, activation functions such as the Rectified Linear Unit (ReLU) (Hahnloser et al., 2000) are commonly used between layers. While convolutional operations are computationally intensive, modern GPU acceleration allows efficient training of large-scale CNN models (Krizhevsky et al., 2012).

CNNs can be designed for different types of data:

- 2D CNNs – primarily used for processing static images.
- 3D CNNs – extend convolutions to three dimensions, allowing networks to capture spatio-temporal features, making them suitable for video and volumetric data (Tran et al., 2018).

3D convolutions analyze sequences of images, incorporating temporal context that is missing in 2D CNNs. It is also possible to combine 2D and 3D convolutions in hybrid architectures, where temporal convolutions can be placed at different levels of the network.

Additionally, CNNs may include transposed convolutional layers (also known as deconvolutional layers), which perform the inverse of convolutions. Unlike standard convolutional layers that reduce spatial dimensions, transposed convolutions expand feature maps, making them essential for tasks like image generation and upsampling.

## 2.3 TRANSFORMERS

Transformers have revolutionized deep learning, particularly in the fields of natural language processing (NLP) and computer vision. Originally introduced in the paper "Attention Is All You Need" (Vaswani et al., 2017), transformers replace traditional recurrent and convolutional architectures with self-attention mechanisms, enabling more efficient learning of long-range dependencies in data.

### 2.3.1 *Core Architecture of Transformers*

A transformer model primarily consists of encoder and decoder blocks, each containing:

- Multi-Head Self-Attention (MHSA) – captures contextual relationships between all tokens in an input sequence, regardless of their distance.
- Position-Wise Feedforward Networks – fully connected layers applied to each token independently.
- Positional Encoding – since transformers lack recurrence, positional encodings are added to input embeddings to retain order information.
- Layer Normalization and Residual Connections – improve stability and gradient flow during training.

A single self-attention mechanism can be formulated as follows:

$$\text{Attention}(Q, K, V) = \text{softmax}\left(\frac{QK^T}{\sqrt{d_k}}\right)V, \tag{2.2}$$

where:

- $Q$ (queries), $K$ (keys), and $V$ (values) are projections of the input data.
- $d_k$ is the dimensionality of keys (used for scaling).

- The softmax function normalizes the attention scores, determining how much focus should be placed on each input token, by converting a vector of raw scores $z$ into probabilities as $\text{softmax}(z_i) = \frac{e^{z_i}}{\sum_j e^{z_j}}$ (Goodfellow et al., 2016).

The multi-head attention (MHA) mechanism extends this concept by applying multiple self-attention layers in parallel, each with different learned projections:

$$\text{MHA}(Q, K, V) = \text{Concat}(\text{head}_1, \ldots, \text{head}_h)W^O, \tag{2.3}$$

where each attention head computes its own self-attention operation as

$$\text{head}_i = \text{Attention}(QW_i^Q, KW_i^K, VW_i^V),$$

with $W_i^Q$, $W_i^K$, and $W_i^V$ being learned projection matrices specific to the $i$-th head. Each head captures different aspects of the input relationships by focusing on different parts of the input in parallel. The outputs from all heads are then concatenated and passed through a learned linear transformation matrix $W^O$, where the superscript $O$ denotes output. This final projection maps the concatenated outputs back to the original embedding dimension, ensuring consistent representation size across layers.

### 2.3.2 *Transformers in Large Language Models (LLMs)*

Transformers serve as the backbone for modern large language models (LLMs), which are deep neural networks trained on vast amounts of text data to perform a wide range of language-related tasks, such as text generation, translation, and question answering. Examples include:

- BERT (Devlin et al., 2019) – a bidirectional model trained using masked language modeling.
- GPT-series (e.g., GPT-4) (Radford et al., 2019; Achiam et al., 2023) – autoregressive models generating coherent text based on large-scale training data.
- T5 (Raffel et al., 2020) – a unified text-to-text framework for multiple NLP tasks.

These models leverage transformers' ability to process entire sequences in parallel, leading to significant improvements in scalability and efficiency over recurrent architectures such as RNNs and LSTMs, which process data sequentially and suffer from limited parallelism.

### 2.3.3 *Transformers in Computer Vision*

Vision Transformers (ViTs) (Dosovitskiy et al., 2021) apply transformers to image classification and segmentation by treating image patches as tokens. Unlike convolutional networks, ViTs learn global feature representations directly from raw images, allowing them to surpass CNN-based models in some cases.

Key adaptations for vision tasks include:

- Patch Embeddings – images are divided into small patches, which are then flattened (i.e., converted from a 2D array of pixel values into a 1D vector) and linearly projected.
- Positional Encoding – similar to NLP, positional encodings are added to retain spatial structure.
- Self-Attention in Image Tokens – enables long-range dependencies across different image regions.

Transformer-based architectures have also been extended to tasks such as:

- Object detection (DETR (Carion et al., 2020))
- Image synthesis (DALL·E (Ramesh et al., 2021))
- Video understanding (TimeSformer (Bertasius et al., 2021))

Transformers offer several key advantages over traditional deep learning architectures. One of their most notable strengths is their ability to model long-range dependencies, unlike CNNs and RNNs, which rely on local receptive fields or sequential computations. This allows transformers to capture global context effectively, making them well-suited for tasks requiring an understanding of relationships across entire sequences. Another advantage is scalability; transformers can be efficiently trained on massive datasets using modern hardware accelerators such as GPUs and TPUs (Tensor Processing Units), leveraging large-scale parallelization. Unlike recurrent models, transformers process entire sequences simultaneously, removing the need for sequential computations and significantly reducing training time. Additionally, transformers demonstrate remarkable versatility, being successfully applied across various domains, including natural language processing, computer vision, and multimodal learning. Their adaptability has led to the development of unified AI models capable of handling multiple modalities within a single framework, further expanding their applicability in real-world scenarios.

## 2.4 AUTOENCODERS

Autoencoders are a class of neural networks designed for dimensionality reduction and feature learning in an unsupervised manner (Wang et al., 2016; Gadirov, 2020). Their primary objective is to learn a compact representation of input data by reconstructing it as accurately as possible. Autoencoders are particularly useful when working with high-dimensional data, as they can extract meaningful latent features while discarding noise and redundant information. Their applications span various domains, including anomaly detection, image processing, information retrieval, drug discovery, and machine translation.

Several variations of autoencoders exist, each tailored to specific applications and optimization objectives. In the following sections, we describe the fundamental structure of a standard autoencoder and explore its extensions.

### 2.4.1 *Standard Autoencoder*

A standard autoencoder consists of two main components: an encoder and a decoder. The encoder receives the input data and compresses it into a lower-dimensional latent representation (also known as the code). This representation captures the most salient features of the input. The decoder then attempts to reconstruct the original input from this latent space, ensuring minimal information loss. The overall structure of an autoencoder is illustrated in Figure 2.3.

Mathematically, for an input $x \in \mathbb{R}^d$, the encoding process in a single-layer autoencoder can be expressed as:

$$h = \sigma(Wx + b), \tag{2.4}$$

where $h$ is the latent representation, $W \in \mathbb{R}^{m \times d}$ is the weight matrix, $b \in \mathbb{R}^m$ is the bias vector, and $\sigma$ is an activation function such as ReLU or sigmoid.

The decoder reconstructs the input using a similar transformation:

$$\hat{x} = \sigma(W'h + b'), \tag{2.5}$$

where $\hat{x}$ is the reconstructed output, $W' \in \mathbb{R}^{d \times m}$ is the weight matrix of the decoder, and $b'$ is the bias vector.

Autoencoders can be implemented using deep neural networks with multiple hidden layers, often composed of convolutional and fully connected layers. The training process aims to minimize the reconstruction loss, commonly measured using either mean squared error (MSE) or cross-entropy (De Boer et al., 2005):

For mean squared error:

$$\mathcal{L}_{\text{MSE}}(x, \hat{x}) = \|x - \hat{x}\|^2, \tag{2.6}$$

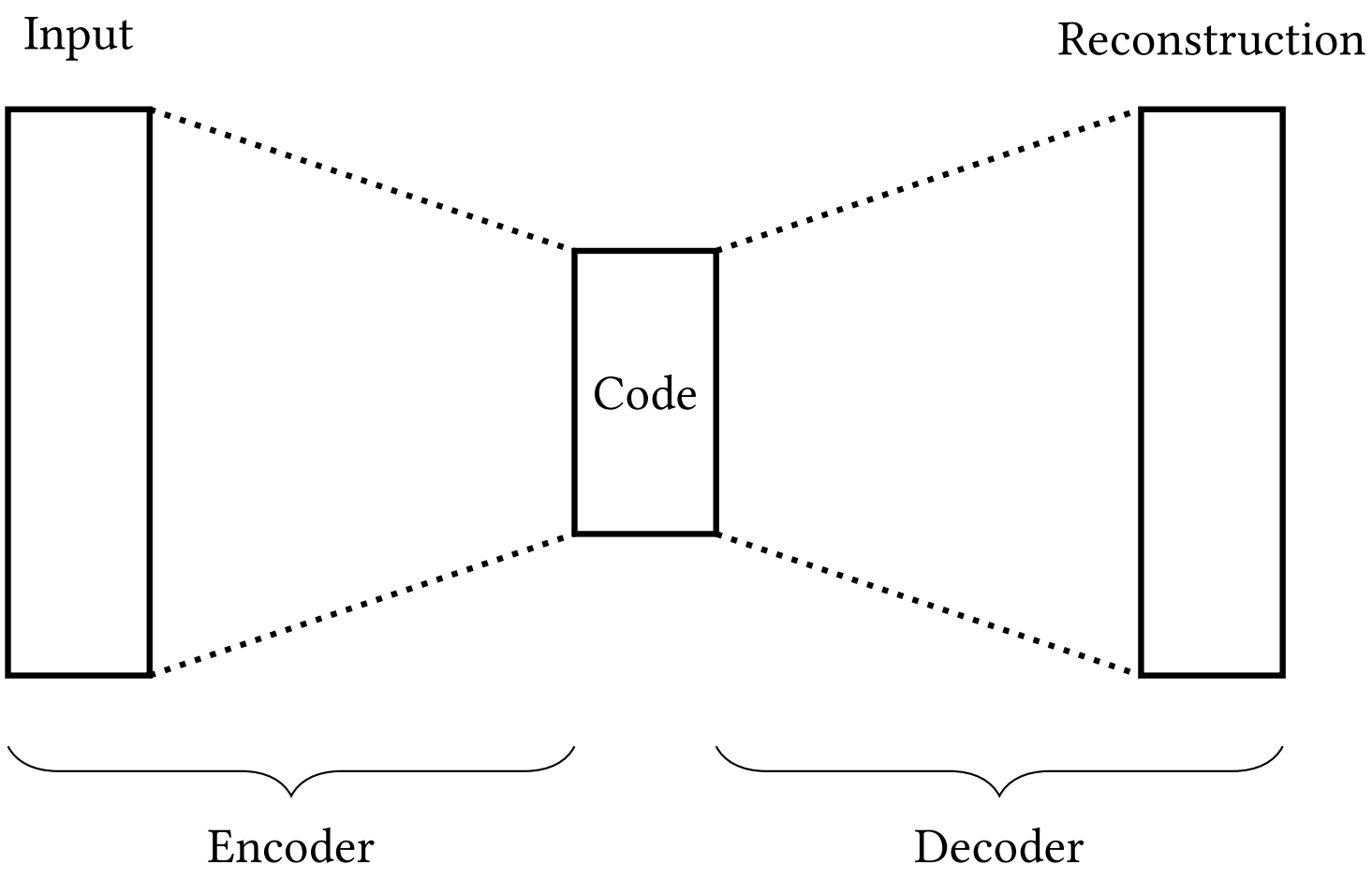

Figure 2.3: Autoencoder structure consisting of two parts: an encoder and a decoder. The middle part corresponds to a compressed representation, here referred to as the "code".

where $x$ is the input and $\hat{x}$ is the reconstructed output.

For binary cross-entropy:

$$\mathcal{L}_{\mathrm{BCE}}(x, \hat{x}) = -\sum_i \left(x_i \log(\hat{x}_i) + (1 - x_i)\log(1 - \hat{x}_i)\right), \qquad (2.7)$$

where the sum is over all input features.

The choice of loss function depends on the nature of the input data: MSE is commonly used for continuous-valued inputs, while cross-entropy is preferred for binary or probabilistic outputs.

Training of an autoencoder is performed using backpropagation (Hecht-Nielsen, 1992), optimizing the network parameters to reduce the reconstruction error. Through this process, autoencoders learn efficient data representations, making them powerful tools for dimensionality reduction, anomaly detection, and generative modeling.

A standard autoencoder does not impose explicit constraints on its latent representation. However, several regularized variants exist to enhance its learning capabilities. These include sparse autoencoders (Section 2.4.2), which encourage sparsity in the latent space; denoising autoencoders, which learn to reconstruct inputs from corrupted versions; and contractive autoencoders, which enforce robustness by penalizing small changes in the input space. These variations improve the generalization and applicability of autoencoders across different tasks.

### 2.4.2 *Sparse Autoencoder*

A sparse autoencoder is a regularized variant of the standard autoencoder that enforces sparsity in the latent representation. Unlike traditional autoencoders, which may allow all hidden units to be active simultaneously, sparse autoencoders constrain only a subset of neurons to be active at a given time. This sparsity constraint improves feature learning and prevents overfitting (Ng et al., 2011).

Sparsity can be achieved using different regularization techniques:

- **L1 and L2 Regularization:** The Least Absolute Shrinkage and Selection Operator (LASSO, or L1) and Ridge Regression (L2) are common regularization methods (Arpit et al., 2016). L1 regularization enforces sparsity by driving some weight values to zero, while L2 regularization penalizes large weight values, encouraging smaller and more distributed activations.

- **Kullback-Leibler (KL) Divergence:** Another method is to use KL divergence (Kullback and Leibler, 1951), which measures the difference between the average activation of a neuron and a desired sparsity level. By minimizing this divergence, the model ensures that most neurons remain inactive while only a few fire strongly.

- **Dropout Regularization:** Dropout is another widely used regularization technique (Hinton et al., 2012b), where a fraction of neurons is randomly deactivated during training. This prevents the model from relying too much on specific neurons and promotes more robust feature learning.

Mathematically, the KL divergence sparsity constraint can be expressed as:

$$\mathcal{L}_{\text{sparse}} = \sum_{j=1}^{n_h} KL(\rho \| \hat{\rho}_j), \tag{2.8}$$

where $\rho$ is the desired activation level, $\hat{\rho}_j$ is the average activation of neuron $j$, and $KL(\cdot)$ is the Kullback-Leibler divergence function:

$$KL(\rho \| \hat{\rho}_j) = \rho \log \frac{\rho}{\hat{\rho}_j} + (1-\rho) \log \frac{1-\rho}{1-\hat{\rho}_j}. \tag{2.9}$$

By incorporating these regularization techniques, sparse autoencoders improve feature selection, enhance generalization, and provide better representations for high-dimensional data.

### 2.4.3 *Denoising Autoencoder*

A **Denoising Autoencoder (DAE)** is a variant of the standard autoencoder designed to learn robust feature representations by reconstructing clean data from corrupted inputs (Vincent et al., 2008). Unlike conventional autoencoders, which learn to reproduce the input exactly, DAEs are trained to remove noise and recover the underlying signal. This forces the network to focus on essential structures rather than memorizing trivial details.

The corruption process can involve different types of noise, such as:

- **Gaussian noise:** Adds small perturbations sampled from an isotropic Gaussian distribution.
- **Salt-and-pepper noise:** Randomly sets pixel values to either the minimum or maximum intensity.
- **White noise:** A uniform random distribution of noise applied across the input (Boyat and Joshi, 2015).

Given an original input $x$, a corrupted version $\tilde{x}$ is obtained by applying a noise function $\mathcal{N}$, such that:

$$\tilde{x} = x + \mathcal{N}(x). \tag{2.10}$$

The DAE is then trained to reconstruct the clean input $x$ from the noisy version $\tilde{x}$ by minimizing a reconstruction loss, typically the Mean Squared Error (MSE):

$$\mathcal{L}_{\text{DAE}} = \|x - f_\theta(\tilde{x})\|^2, \tag{2.11}$$

where $f_\theta$ represents the encoding and decoding functions parameterized by $\theta$.

By training on corrupted data, DAEs learn robust feature representations that generalize better to variations in real-world data. This makes them highly effective for tasks such as image denoising, anomaly detection, and representation learning.

### 2.4.4 *Contractive Autoencoder*

A Contractive Autoencoder (CAE) is a regularized variant of the standard autoencoder designed to learn robust representations by penalizing sensitivity to small variations in the input space (Rifai et al., 2011). Unlike standard autoencoders, which only minimize reconstruction loss, CAEs introduce an additional regularization term that encourages invariance to minor perturbations in the input.

The key idea is to constrain the learned representations such that small changes in the input produce minimal variations in the latent space. This is achieved by adding a contractive penalty to the loss function, defined as the Frobenius norm of the Jacobian of the encoder's activations with respect to the input:

$$\mathcal{L}_{\text{CAE}} = \mathcal{L}_{\text{recon}} + \lambda \sum_i \|\nabla_x h_i\|^2, \tag{2.12}$$

where:

- $\mathcal{L}_{\text{recon}}$ is the reconstruction loss (e.g., Mean Squared Error),
- $h_i$ represents the activation of the $i$-th hidden unit (neuron) in the hidden layer,
- $\nabla_x h_i$ is the Jacobian matrix measuring how changes in input $x$ affect the hidden representation,
- $\lambda$ is a hyperparameter controlling the strength of the contractive penalty.

By enforcing a small Jacobian norm, CAEs encourage learning representations that are robust to small variations in the input. This property makes them particularly useful for tasks such as denoising, anomaly detection, and feature extraction in high-dimensional datasets.

Unlike denoising autoencoders, which introduce explicit noise to the input, CAEs achieve robustness implicitly by discouraging sharp changes in the learned features. This makes them well-suited for applications where stability in representation is crucial.

### 2.4.5 *Variational and Beta-Variational Autoencoders*

Variational Autoencoders (VAEs) build upon traditional autoencoders but incorporate probabilistic inference to model latent variable distributions (Kingma and Welling, 2014). While standard autoencoders map inputs to fixed latent representations, VAEs impose a structured distribution on the latent space, ensuring continuity and meaningful interpolation. This allows for better generalization and enables generative capabilities.

VAEs belong to the family of generative models and are formulated within the framework of variational inference and probabilistic graphical models. One limitation of standard autoencoders is that their latent space lacks regularization, leading to overfitting and poor generalization. VAEs address this by enforcing a prior distribution over the latent space and learning an approximate posterior that closely follows this prior.

#### 2.4.5.1 *Variational Inference and Approximate Posteriors*

Performing efficient probabilistic inference in deep models is challenging, particularly when dealing with intractable posterior distributions and large datasets. Variational Bayesian (VB) methods approximate intractable posteriors by optimizing a variational distribution to be close to the true posterior. However, traditional VB approaches often yield non-differentiable analytical solutions, making gradient-based optimization infeasible. Monte Carlo methods, which involve sampling, can be used but are computationally expensive and inefficient for large-scale learning.

#### 2.4.5.2 *Stochastic Gradient Variational Bayes (SGVB)*

(Kingma and Welling, 2014) introduced a novel approach that overcomes these challenges by developing an unbiased estimator for intractable posterior distributions and applying this estimator to construct VAEs. Their key contributions include:

- A reparameterization strategy that transforms the intractable posterior into a differentiable function, enabling backpropagation-based learning.
- The introduction of Stochastic Gradient Variational Bayes (SGVB), which optimizes the variational lower bound using stochastic gradient descent.

The core idea of SGVB is to approximate the intractable posterior by learning a parameterized variational distribution that is computationally efficient. The posterior is modeled as a Gaussian distribution with learnable mean and variance, enabling the application of standard gradient-based optimization techniques.

#### 2.4.5.3 *Auto-Encoding Variational Bayes (AEVB) Algorithm*

To facilitate scalable inference in large datasets, the Auto-Encoding Variational Bayes (AEVB) algorithm was introduced (Kingma and Welling, 2014). It leverages the SGVB estimator and applies it within an autoencoder architecture. The model consists of two neural networks:

- An *encoder*, which approximates the posterior distribution over latent variables.
- A *decoder*, which reconstructs the input data from the latent representation.

Unlike standard autoencoders, VAEs do not directly encode a deterministic latent vector. Instead, the encoder learns to output a distribution from which latent variables are sampled. This probabilistic formulation ensures smooth and continuous transitions within the latent space.

#### 2.4.5.4 *Empirical Evaluation and Applications*

VAEs have been evaluated on various benchmark datasets, including MNIST (Kingma and Welling, 2014) and Frey Face (Kingma and Welling, 2014; Lawrence et al., 1997), demonstrating superior performance compared to earlier approaches such as Wake-Sleep (Hinton et al., 1995) and Monte Carlo Expectation-Maximization (MCEM) (Wei and Tanner, 1990). Experimental results indicate that the variational lower bound serves as a natural regularizer, reducing the need for additional constraints to prevent overfitting (Kingma and Welling, 2014).

By leveraging latent space regularization, VAEs enable applications in generative modeling, anomaly detection, and structured representation learning. However, selecting an appropriate prior distribution remains a challenge, and further research is required to explore alternative priors for complex data distributions.

#### 2.4.5.5 *Mathematical Formulation of Variational Autoencoders*

To formalize the mathematical model of variational autoencoders (VAEs), we consider a dataset consisting of samples from a distribution $x$. A latent variable $z$ is assumed to be drawn from a prior distribution $p_{\theta^*}(z)$, which governs the data generation process through the conditional distribution $p_{\theta^*}(x|z)$. Since the true parameter $\theta^*$ is unknown, we model the distributions using parameterized functions $p_\theta(z)$ and $p_\theta(x|z)$ with differentiable probability density functions.

#### 2.4.5.6 *Marginal Likelihood and Posterior Distribution*

The marginal likelihood of the data is expressed as:

$$p_\theta(x) = \int p_\theta(z) p_\theta(x|z) dz. \tag{2.13}$$

Applying Bayes' theorem, the posterior distribution is given by:

$$p_\theta(z|x) = \frac{p_\theta(x|z) p_\theta(z)}{p_\theta(x)}. \tag{2.14}$$

However, computing $p_\theta(x)$ requires marginalizing over all possible latent variables, which is generally intractable due to the complexity of $p_\theta(x|z)$, often modeled as a deep neural network.

#### 2.4.5.7 *Variational Approximation and the Recognition Model*

To approximate the intractable posterior $p_\theta(z|x)$, we utilize a recognition model $q_\phi(z|x)$, which is learned as an encoder network in the VAE framework. The key difference from standard autoencoders is that instead of mapping inputs to a deterministic latent vector, VAEs encode data into a probability distribution over the latent space.

To measure the difference between the approximate posterior $q_\phi(z|x)$ and the true posterior $p_\theta(z|x)$, the Kullback-Leibler (KL) divergence (introduced in Equation 2.9 in a binary form) is used:

$$D_{KL}(q_\phi(z|x)||p_\theta(z|x)) = \sum_z q_\phi(z|x) \log \frac{q_\phi(z|x)}{p_\theta(z|x)}. \tag{2.15}$$

Applying Bayes' theorem, we obtain:

$$\begin{aligned} D_{KL}[q_\phi(z|x) \,||\, p_\theta(z|x)] = \mathbb{E}_{q_\phi(z|x)} \big[ &\log q_\phi(z|x) - \log p_\theta(x|z) \\ &- \log p_\theta(z) + \log p_\theta(x) \big]. \end{aligned} \tag{2.16}$$

Since $p_\theta(x)$ is independent of $z$, we can rewrite Equation 2.16 as:

$$\begin{aligned} \log p_\theta(x) - D_{KL}[q_\phi(z|x) \,||\, p_\theta(z|x)] = \mathbb{E}_{q_\phi(z|x)}&[\log p_\theta(x|z)] \\ &- D_{KL}[q_\phi(z|x) \,||\, p_\theta(z)]. \end{aligned} \tag{2.17}$$

#### 2.4.5.8 *Variational Lower Bound (ELBO)*

Rearranging the terms in Equation 2.17, we obtain:

$$\log p_\theta(x) = D_{KL}\big[q_\phi(z|x) \,\|\, p_\theta(z|x)\big] + \mathcal{L}(\theta, \phi; x), \tag{2.18}$$

where $D_{KL}(\cdot\|\cdot)$ is defined in Equation 2.15 and $\mathcal{L}(\theta, \phi; x)$ is the evidence lower bound (ELBO), defined as

$$\mathcal{L}(\theta, \phi; x) = \mathbb{E}_{q_\phi(z|x)}\big[\log p_\theta(x|z)\big] - D_{KL}\big[q_\phi(z|x) \,\|\, p_\theta(z)\big]. \tag{2.19}$$

In Equation 2.18, the two terms on the right-hand side are (1) the KL divergence from Equation 2.15 and (2) the ELBO defined in Equation 2.19, $\mathcal{L}(\theta, \phi; x)$.

Since the KL divergence is always non-negative, maximizing the ELBO is equivalent to maximizing the marginal likelihood $\log p_\theta(x)$. The second term in Equation 2.19 regularizes the latent space by ensuring that the learned distribution is close to the prior, while the first term enforces accurate reconstruction.

#### 2.4.5.9 *Gaussian Assumption and KL Divergence Term*

Typically, the prior $p_\theta(z)$ is chosen as a standard Gaussian $\mathcal{N}(0, I)$, while the approximate posterior is modeled as a Gaussian $\mathcal{N}(\mu, \sigma^2)$ with parameters $\mu$ and $\sigma$ output by the encoder. The KL divergence between these two Gaussians has a closed-form solution:

$$D_{KL}\big(q_\phi(z|x) \,\|\, p_\theta(z)\big) = -\frac{1}{2} \sum_{i=1}^{d} \big(1 + \log \sigma_i^2 - \mu_i^2 - \sigma_i^2\big), \tag{2.20}$$

where $d$ is the dimensionality of the latent variable $z$, and the sum runs over each latent dimension $i$. This term penalizes deviations from the prior, enforcing smoothness in the latent space and preventing overfitting.

#### 2.4.5.10 *Reparameterization Trick*

Backpropagation computes gradients by applying the chain rule to deterministic operations in the computational graph, a directed acyclic graph that represents the sequence of operations and data flow used to compute the output from the input. However, the operation $z \sim q_\phi(z|x)$ is non-differentiable because sampling from a distribution does not yield a gradient with respect to the encoder parameters $\phi$. The reparameterization trick circumvents this by expressing the random variable $z$ as a deterministic function of the parameters and an auxiliary noise variable:

$$z = \mu + \sigma \odot \epsilon, \quad \epsilon \sim \mathcal{N}(0, I), \tag{2.21}$$

where $\odot$ denotes element-wise multiplication. In this context, the *stochastic node* refers to the point in the computational graph where randomness is introduced by the noise $\epsilon$. By isolating the randomness into $\epsilon$ and making $z$ a differentiable function of $\mu$ and $\sigma$, gradients can flow through the sampling step, enabling efficient optimization via backpropagation.

#### 2.4.5.11 *Training the VAE*

During training, the encoder network predicts $\mu$ and $\log \sigma$, which are used to sample $z$. The decoder then reconstructs the input from $z$. The objective function consists of two competing terms:

- A reconstruction loss, typically modeled as the log-likelihood $\log p_\theta(x|z)$.
- A regularization term given by the KL divergence, encouraging a structured latent space.

By optimizing the ELBO using stochastic gradient descent, VAEs learn to generate meaningful representations of high-dimensional data, making them powerful tools for generative modeling, anomaly detection, and representation learning.

#### 2.4.5.12 *β-Variational Autoencoder*

The $\beta$-variational autoencoder ($\beta$-VAE) (Higgins et al., 2017) extends the standard variational autoencoder by introducing a hyperparameter $\beta$ that controls the trade-off between the reconstruction quality and the smoothness of the latent space. The objective function of the $\beta$-VAE differs from that of the standard VAE by incorporating a weighting factor for the Kullback-Leibler (KL) divergence term.

#### 2.4.5.13 *Formulation as a Constrained Optimization Problem*

The $\beta$-VAE objective can be formulated as a constrained optimization problem with a constraint $\delta$ on the KL divergence:

$$\begin{aligned} \max_{\phi,\theta} \quad & \mathbb{E}_x\Big[\mathbb{E}_{q_\phi(z|x)}[\log p_\theta(x|z)]\Big] \\ \text{s.t.} \quad & D_{KL}\big(q_\phi(z|x) \,\|\, p(z)\big) \le \delta, \end{aligned} \tag{2.22}$$

where $\delta \ge 0$ specifies the maximum allowable divergence. By introducing a non-negative Lagrange multiplier $\beta$ for this constraint, we obtain the equivalent unconstrained (Lagrangian) objective:

$$\mathcal{L}_{\beta\text{-VAE}}(\theta, \phi; x) = -\,\mathbb{E}_{q_\phi(z|x)}[\log p_\theta(x|z)] + \beta\, D_{KL}\big(q_\phi(z|x) \,\|\, p(z)\big). \tag{2.23}$$

Here, $\beta$ is the Lagrange multiplier associated with the KL divergence constraint.

#### 2.4.5.14 *Effect of the Hyperparameter $\beta$*

The hyperparameter $\beta$ influences the balance between the reconstruction term and the regularization term:

- When $\beta = 1$, the objective function reduces to the standard VAE formulation.
- When $\beta < 1$, the model prioritizes reconstruction, making it behave more like a standard autoencoder.
- When $\beta > 1$, the KL term is emphasized, enforcing a more disentangled (independent) and structured (Higgins et al., 2017) latent representation at the cost of reconstruction accuracy.

A higher $\beta$ value constrains the amount of information retained in the latent space, leading to improved disentanglement of latent factors. Disentanglement refers to the degree to which each dimension of the latent representation independently corresponds to a single, distinct generative factor of variation, so that varying one latent coordinate produces a change in one interpretable feature of the output while leaving other features unchanged. However, excessive regularization can degrade reconstruction quality, making it crucial to tune $\beta$ based on the dataset and the desired properties of the latent space.

#### 2.4.5.15 *Applications and Trade-offs*

$\beta$-VAE has been widely used in applications that require disentangled representations, such as interpretable latent space learning, generative modeling, and structured data synthesis. However, selecting an appropriate $\beta$ remains an open challenge, as overly restrictive constraints may lead to underfitting, while too little regularization may not enforce the desired structure in the latent space.

### 2.4.6 *Autoencoders for Volume Rendering Time Prediction*

Autoencoders (AEs) are widely used for dimensionality reduction and feature extraction, particularly in applications where high-dimensional data needs to be efficiently compressed while preserving essential structural details. While AEs are commonly applied in tasks such as anomaly detection, denoising, and generative modeling, they can also be leveraged for predictive modeling in scientific visualization. One such application is volume rendering time prediction, where AEs extract meaningful representations from volumetric data to aid in estimating rendering time.

Recent methods have proposed AE-based pipelines that encode volumetric data into compact feature vectors, which are then combined with rendering parameters to predict rendering times. Typically, these approaches employ symmetric 3D convolutional autoencoders to compress input volumes into low-dimensional latent representations that preserve key structural characteristics while reducing computational overhead. The extracted features are then passed to a predictive network that estimates rendering costs under varying camera configurations and time steps. Such AE-driven compression and predictive modeling enable fast and accurate performance estimation, improving scalability and resource allocation in large-scale visualization workflows (Yin et al., 2025).

## 2.5 DIMENSIONALITY REDUCTION

Visualizing and analyzing high-dimensional datasets is a fundamental challenge in many scientific and machine learning applications. Dimensionality reduction (DR) techniques address this by projecting high-dimensional data onto a lower-dimensional space while preserving its essential structure (Liu et al., 2016). This process reduces the number of input features or dimensions while retaining the most informative features. DR is commonly employed for data visualization, storage optimization, noise reduction, and feature extraction, and also serves as a preprocessing step for (other) machine learning models.

A significant advantage of dimensionality reduction is mitigating the curse of dimensionality, a term introduced by Bellman in 1957 (Bellman, 1966). In high-dimensional spaces, data points tend to become increasingly sparse, rendering traditional distance metrics ineffective. Even in large datasets, an excessive number of dimensions can make statistical analysis unreliable. DR methods alleviate this problem by reducing redundancy and improving the discriminability of meaningful structures within the data.

Dimensionality reduction techniques are broadly categorized into feature selection and feature extraction approaches. Feature selection involves identifying a subset of the most relevant variables from the original dataset, whereas feature extraction constructs a new, lower-dimensional space by deriving new features from the input data.

Additionally, DR techniques can be classified into linear and nonlinear methods. Linear methods, such as Principal Component Analysis (PCA) and Linear Discriminant Analysis (LDA), construct transformations where the new feature axes remain interpretable in terms of the original variables. Nonlinear methods, often referred to as manifold learning techniques, such as t-SNE (t-distributed Stochastic Neighbor Embedding) (Maaten and Hinton, 2008) and UMAP (Uniform Manifold Approximation and Projection) (McInnes et al., 2018), are designed to capture complex geometric structures within the data but may lack direct interpretability.

An effective DR technique should not only generate a meaningful low-dimensional representation but also allow for a quantitative evaluation of its performance. Different metrics, such as reconstruction error, trustworthiness, and preservation of neighborhood structures, are commonly used to assess the quality of embeddings.

In this work, we employ a variety of DR methods, including Autoencoders, PCA, t-SNE, and UMAP, to analyze and visualize ensemble datasets. The following sections provide an in-depth discussion of these techniques and their applications.

### 2.5.1 *Principal Component Analysis (PCA)*

Principal Component Analysis (PCA) is a well-established and widely used technique for dimensionality reduction, pattern recognition, and data visualization (Wold et al., 1987). It is a linear, non-parametric method based on linear algebra, implemented through either the eigenvalue decomposition of the covariance matrix or the singular value decomposition (SVD) of the data matrix (Shlens, 2014). PCA seeks to transform high-dimensional data into a lower-dimensional space while preserving as much variance as possible.

The method constructs a new coordinate system, where the principal components (PCs) are defined such that each successive component captures the maximum possible variance while remaining orthogonal to the previous components. This transformation ensures that the data is projected onto a set of uncorrelated axes, effectively reducing redundancy and improving interpretability.

Given a dataset represented as a zero-mean matrix $X \in \mathbb{R}^{m \times n}$ (where $m$ is the number of samples and $n$ is the number of features), the covariance matrix is computed as:

$$C = \frac{1}{m} X^T X. \tag{2.24}$$

The principal components are obtained by solving the eigenvalue decomposition problem:

$$CV = V\Lambda, \tag{2.25}$$

where $V$ contains the eigenvectors (principal components) and $\Lambda$ is a diagonal matrix with corresponding eigenvalues, indicating the variance captured by each component. The data can then be projected onto the principal components:

$$Z = XV_k, \tag{2.26}$$

where $V_k$ contains the top $k$ eigenvectors corresponding to the largest eigenvalues, and $Z \in \mathbb{R}^{m \times k}$ is the lower-dimensional representation.

Alternatively, PCA can be computed via singular value decomposition (SVD):

$$X = U\Sigma W^T, \tag{2.27}$$

where $U$ and $W$ contain the left and right singular vectors, respectively, and $\Sigma$ is a diagonal matrix containing the singular values. The columns of $W$ form a complete orthonormal basis of principal directions. Selecting the first $k$ columns of $W$ corresponds to projecting the data onto the subspace spanned by the top $k$ principal components, thus achieving dimensionality reduction (compression).

**Importance of Normalization.** Before applying PCA, it is crucial to normalize the data to have zero mean and unit variance. This step ensures that all features contribute equally to the principal components and prevents features with larger magnitudes from dominating the decomposition.

$$X_{\text{norm}} = \frac{X - \mu}{\sigma}, \tag{2.28}$$

where $\mu$ is the mean and $\sigma$ is the standard deviation of each feature.

**Nonlinear Extension: Kernel PCA.** While PCA is a linear technique, its nonlinear extension, Kernel PCA, maps data into a higher-dimensional space using a kernel function before applying PCA (Schölkopf et al., 1998). This allows PCA to capture complex structures that are not linearly separable.

PCA is widely used in data visualization, noise reduction, feature extraction, and exploratory data analysis. It serves as a fundamental tool for understanding high-dimensional datasets and is often used as a preprocessing step for machine learning models.

### 2.5.2 *t-Distributed Stochastic Neighbor Embedding (t-SNE)*

t-Distributed Stochastic Neighbor Embedding (t-SNE) is a dimensionality reduction technique that has become widely used for visualizing high-dimensional datasets (Maaten and Hinton, 2008). Unlike PCA, t-SNE is a non-linear, parametric method that focuses on preserving local structures in the data. However, it is computationally expensive and susceptible to getting stuck in local optima.

The goal of t-SNE is to represent high-dimensional data in a lower-dimensional space, typically two or three dimensions, to facilitate human interpretation. The results are commonly visualized using scatter plots.

**Pairwise Similarities in High-Dimensional Space.** The first step in t-SNE is computing pairwise similarities between data points in the high-dimensional space. The similarity between two data points $x_i$ and $x_j$ is expressed as a conditional probability that measures how likely $x_j$ would be chosen as a neighbor of $x_i$:

$$p_{j|i} = \frac{\exp(-\left\|x_i - x_j\right\|^2 / 2\sigma_i^2)}{\sum_{k \neq i} \exp(-\|x_i - x_k\|^2 / 2\sigma_i^2)}, \tag{2.29}$$

where $\sigma_i^2$ is the variance of the Gaussian distribution centered at $x_i$. The variance $\sigma_i$ is determined based on a user-defined perplexity parameter, which typically ranges between 5 and 50. The perplexity controls the effective number of neighbors considered when measuring similarity.

To ensure symmetric relationships, the pairwise similarities are further symmetrized as follows:

$$p_{ij} = \frac{p_{j|i} + p_{i|j}}{2N}, \tag{2.30}$$

where $N$ is the total number of data points.

**Mapping to a Low-Dimensional Space.** Once pairwise similarities are established in the high-dimensional space, the algorithm constructs a corresponding low-dimensional representation. In this space, the similarities between two points $y_i$ and $y_j$ are computed using a Student's t-distribution with one degree of freedom:

$$q_{ij} = \frac{(1 + \left\|y_i - y_j\right\|^2)^{-1}}{\sum_{k \neq l}(1 + \|y_k - y_l\|^2)^{-1}}. \tag{2.31}$$

Unlike traditional Gaussian-based similarity measures, the t-distribution provides better separation of clusters and ensures scale invariance.

**Optimization via KL Divergence.** Finally, the embedding is found by minimizing the divergence between the distributions $p$ and $q$:

$$D_{KL}(p \parallel q) = \sum_{i \neq j} p_{ij} \log \frac{p_{ij}}{q_{ij}}, \tag{2.32}$$

where $D_{KL}(p\|q)$ denotes the KL divergence (introduced in Equation 2.9) between the two full joint distributions with entries $p_{ij}$ and $q_{ij}$. This loss function encourages similar points in the high-dimensional space to remain close in the lower-dimensional space while pushing dissimilar points apart.

**Limitations and Practical Considerations.** Although t-SNE is widely used, it has several drawbacks. The method has a quadratic computational complexity, making it impractical for very large datasets. Additionally, t-SNE is sensitive to initialization and perplexity settings, requiring careful tuning. Because the optimization process is stochastic, multiple runs are often necessary to achieve a stable solution. To improve efficiency, t-SNE is often applied after an initial dimensionality reduction step, such as PCA, which helps remove noise and accelerates convergence. Proper data normalization is also crucial, as t-SNE relies on Euclidean distances. Despite its computational cost, t-SNE remains one of the most effective techniques for visualizing high-dimensional data, particularly for identifying clusters and underlying structures in complex datasets.

### 2.5.3 *Uniform Manifold Approximation and Projection (UMAP)*

Uniform Manifold Approximation and Projection (UMAP) is a non-linear dimensionality reduction technique introduced by McInnes *et al.* (McInnes et al., 2018). Similar to t-SNE, UMAP aims to create a low-dimensional representation of high-dimensional data while preserving both local and global structures. However, it differs significantly in its mathematical foundation, as it is based on concepts from Riemannian geometry and algebraic topology. Unlike t-SNE, which relies on probability distributions for similarity measurements, UMAP constructs a weighted k-nearest neighbors (k-NN) graph using a smoothed distance metric (Peterson, 2009).

**Graph Construction and Weighting.** Given a high-dimensional dataset with $N$ points $\{x_1, x_2, \ldots, x_N\}$, UMAP first builds a weighted directed k-nearest neighbors graph by defining the probability of an edge between points $x_i$ and $x_j$ using:

$$p_{ij} = \exp\left(-\frac{\max(0, d(x_i, x_j) - \rho_i)}{\sigma_i}\right), \tag{2.33}$$

where $d(x_i, x_j)$ is the distance between $x_i$ and $x_j$, $\rho_i$ is the distance to the nearest neighbor (used to ensure local adaptivity), and $\sigma_i$ controls the smoothness of the neighborhood function.

**Low-Dimensional Embedding.** Once the high-dimensional graph is constructed, UMAP optimizes a corresponding low-dimensional representation $\{y_1, y_2, \ldots, y_N\}$ by minimizing the difference between the two graphs. The similarity in the low-dimensional space is modeled using:

$$q_{ij} = \frac{1}{1 + d(y_i, y_j)^2}. \tag{2.34}$$

**Optimization.** The embeddings are found by minimizing the cross-entropy loss between the two distributions:

$$\mathcal{L} = \sum_{i \neq j} p_{ij} \log q_{ij} + (1 - p_{ij}) \log(1 - q_{ij}). \tag{2.35}$$

This cost function encourages that points close in the high-dimensional space remain close in the low-dimensional embedding, while distant points remain separate.

**Key Hyperparameters.** UMAP introduces several hyperparameters that influence its performance:

- **Number of nearest neighbors** ($k$): Defines the size of the local neighborhood considered for manifold approximation. Smaller values emphasize local structure, while larger values capture global relationships.
- **Minimum distance** ($d_{\min}$): Determines how closely points can be packed together in the low-dimensional space. Lower values preserve cluster density, while higher values spread points more evenly.
- **Number of components** ($m$): Specifies the target dimensionality of the projection, commonly set to 2 or 3 for visualization.
- **Distance metric**: Defines how distances are computed in the input space. Common choices include Euclidean, Manhattan, Mahalanobis, and Minkowski distances.

**Advantages and Limitations.** Compared to t-SNE, UMAP is computationally more efficient and better preserves global structures while maintaining local details. However, it is sensitive to hyperparameter choices, and since it constructs graphs based on local distances, the relative spacing between clusters may not always be meaningful. Furthermore, due to its stochastic nature, repeated runs with different initializations may yield slightly different results.

Despite these limitations, UMAP is one of the most powerful and flexible dimensionality reduction techniques, striking a balance between computational efficiency, interpretability, and structure preservation in high-dimensional data analysis.

## 2.6 EXPLAINABLE AI

Explainable Artificial Intelligence (XAI) is a field of research that aims to make machine learning models, particularly deep neural networks, more transparent and interpretable (Gilpin et al., 2018). As modern AI systems grow in complexity, their decision-making processes become increasingly opaque, leading to the so-called "black-box" problem. XAI seeks to bridge this gap by providing insights into how models make predictions, helping to promote trust, reliability, and fairness in AI-driven decision-making.

**Techniques for Explainability.** Several approaches have been developed to enhance the interpretability of AI models:

- **Feature Importance Methods:** Techniques such as SHAP (Shapley Additive Explanations) (Lundberg and Lee, 2017) and LIME (Local Interpretable Model-agnostic Explanations) (Ribeiro et al., 2016) analyze the contribution of individual input features to model predictions.

- **Visualization-based Approaches:** Saliency maps, Grad-CAM (Selvaraju et al., 2017), and attention mechanisms highlight regions in the input data that influence model decisions, improving interpretability in computer vision tasks.

- **Surrogate Models:** Complex models can be approximated by simpler interpretable models, such as decision trees or linear regressions, to provide a human-readable understanding of decision boundaries.

- **Intrinsic Interpretability:** Some models, such as decision trees and generalized additive models (GAMs), are inherently interpretable by design and do not require post-hoc explanation techniques.

**Applications and Challenges.** XAI is particularly crucial in high-stakes domains such as healthcare, finance, and scientific research, where understanding model predictions is essential for accountability and decision-making. In scientific visualization, XAI can help interpret latent representations in autoencoders, analyze the impact of simulation parameters in hypernetworks, and provide insights into how neural networks reconstruct missing data in ensemble datasets.

Despite significant advancements, challenges remain. Many explanation techniques provide approximations rather than exact reasoning, and different methods may yield conflicting interpretations. Fur-thermore, achieving a balance between model performance and interpretability remains an open research question.

Explainability in AI continues to be an active area of study, with ongoing efforts to develop more robust and faithful interpretation methods that enhance trust and usability in deep learning applications.

In this work, we aim to develop methods that go beyond making predictions (e.g., temporal interpolations) by modeling the underlying dynamics of the data, particularly through flow estimation. By incorporating domain knowledge into deep learning models, such as enforcing physically meaningful constraints and leveraging parameter-aware adaptations, we move toward more interpretable and trustworthy AI models. This approach not only improves prediction accuracy, but also enhances the transparency of how models derive their outputs by aligning model behavior with known physical assumptions, reducing their reliance on purely black-box learning.

## 2.7 STUDENT-TEACHER LEARNING

Student-teacher learning is a widely used paradigm in deep learning where a **teacher model** supervises a **student model** to enhance the latter's learning process. This framework is particularly beneficial for knowledge transfer, improving generalization, and enabling smaller models to learn from larger, more complex ones.

### 2.7.1 *Types of Student-Teacher Learning*

There are two primary approaches within this framework:

- **Knowledge Distillation** (Hinton et al., 2015): The student network learns to approximate the teacher's output. Instead of training solely on ground-truth labels, the student minimizes the difference between its predictions and those of the teacher. This process often employs **softened probability distributions** from the teacher's logits to transfer knowledge more effectively.
- **Learning with Privileged Information** (Vapnik and Izmailov, 2015): The teacher provides additional auxiliary supervision (e.g., latent features, additional modalities) that the student does not have access to during inference. This enriches the learning process and enhances student performance.

These approaches have been unified under the **Teacher-Student Optimization Framework** (Lopez-Paz et al., 2016). Based on the learning process, student-teacher models can be categorized as:

- **Offline Student-Teacher Learning**: The teacher model is fully trained before the student model is trained to replicate its knowledge.

- **Online Student-Teacher Learning**: The teacher and student networks are trained jointly in an end-to-end fashion (Hu et al., 2022a).

Online student-teacher learning is particularly effective in dynamically adapting supervision signals, ensuring continuous refinement of the student model. Instead of using a separate pre-trained teacher network, lightweight teacher components can be embedded within the main network to guide training, as seen in prior works (Huang et al., 2022).

### 2.7.2 *Mathematical Formulation*

In knowledge distillation, a student model $S(x; \theta_S)$ learns from both ground-truth labels $y$ and the soft targets provided by a teacher model $T(x; \theta_T)$, where $x$ denotes the input data (e.g., images or text), $\theta_S$ are the learnable parameters of the student model, and $\theta_T$ are the fixed parameters of the pre-trained teacher model. The overall loss function combines standard classification loss with a knowledge distillation loss:

$$\mathcal{L} = \alpha \mathcal{L}_{CE}(S(x), y) + (1 - \alpha)\mathcal{L}_{KD}(S(x), T(x)), \tag{2.36}$$

where:

- $\mathcal{L}_{CE}$ is the cross-entropy loss with ground-truth labels,
- $\mathcal{L}_{KD}$ is the **knowledge distillation loss**, minimizing the divergence between the student and teacher outputs,
- $\alpha$ is a balancing factor controlling the trade-off between direct supervision and distillation.

The distillation loss typically employs **temperature scaling** $\tau$ to smooth the teacher's predictions, improving training stability:

$$\mathcal{L}_{KD} = D_{KL}\left(\sigma\left(\frac{T(x)}{\tau}\right) \parallel \sigma\left(\frac{S(x)}{\tau}\right)\right), \tag{2.37}$$

where $D_{KL}$ is the Kullback-Leibler divergence (introduced in Equation 2.9 in a binary form), and $\sigma(\cdot)$ represents the softmax function.

### 2.7.3 *Application in FLINT*

In the context of FLINT (Chapter 4), student-teacher learning is employed to enhance both flow estimation and scalar field interpolation. By incorporating a teacher module within the network, FLINT benefits from intermediate supervision, allowing the student model to progressively refine its predictions. Instead of relying on a separate, pre-trained teacher model, FLINT integrates a lightweight teacher block that refines its intermediate predictions. This online student-teacher architecture ensures that both scalar field interpolation and flow estimation are optimized jointly, leading to higher accuracy—i.e., lower reconstruction error for scalar and vector fields—and temporal coherence in scientific ensemble data.

## 2.8 HYPERNETWORKS

Hypernetworks (Ha et al., 2016) have emerged as a powerful paradigm in deep learning, enabling one neural network (the hypernetwork) to generate the weights of another target network dynamically. This approach offers greater flexibility in capturing complex relationships between data and model parameters, particularly in scenarios where the target model needs to adapt to different conditions without retraining. Hypernetworks are especially useful in tasks involving parameter space exploration, as they allow the target network to adjust its behavior based on external inputs such as simulation parameters.

A standard neural network learns a set of weights $\theta$ that are optimized to minimize a loss function. In contrast, a hypernetwork learns a mapping function that generates these weights dynamically:

$$\theta = H(\mathbf{p}; \phi),$$

where:

- $H$ is the hypernetwork with its own parameters $\phi$,
- $\mathbf{p}$ is an input vector (e.g., simulation parameters),
- $\theta$ represents the weights of the target network.

Instead of directly training a fixed set of weights, the hypernetwork learns to generate the weights based on $\mathbf{p}$, allowing the target network to adapt on-the-fly to different input conditions. This dynamic adaptation enables efficient generalization across various configurations without requiring costly retraining.

Hypernetworks have been successfully applied in fields such as:

- **Neural Architecture Search (NAS):** They help in optimizing architectures by dynamically adjusting network structures (Brock et al., 2017).
- **Generative Modeling:** They improve sample quality by parameterizing generative models (Skorokhodov et al., 2021).
- **Meta-Learning:** They enhance learning efficiency by adapting network parameters based on task-specific inputs (Przewieźlikowski et al., 2023).

### 2.8.1 *Applications and Relevance to HyperFLINT*

In HyperFLINT (Chapter 5), we leverage hypernetworks to extend traditional neural networks for flow estimation and temporal interpolation in scientific ensembles. Instead of using a fixed set of weights, the hypernetwork dynamically generates convolutional filters in the target network based on simulation parameters. This allows HyperFLINT to adapt seamlessly to different simulation conditions, improving both accuracy and efficiency while enabling parameter space exploration.

## 2.9 VOLUME RENDERING

Volume rendering is a fundamental technique for visualizing three-dimensional scalar fields, commonly used in scientific visualization, medical imaging, and computer graphics. Unlike surface-based rendering, which only represents object boundaries, volume rendering enables the visualization of internal structures within a dataset.

### 2.9.1 *Direct and Indirect Volume Rendering*

Volume rendering techniques are broadly categorized into:

- **Direct Volume Rendering (DVR)**: This approach visualizes the entire volumetric data without explicitly extracting surfaces. Methods such as **Ray Marching** and **Ray Casting** (Levoy, 1988) compute how light interacts with the volume to produce images with realistic depth and shading.
- **Indirect Volume Rendering**: In contrast, methods such as **Marching Cubes** (Lorensen and Cline, 1998) first extract explicit geometric representations (e.g., isosurfaces) before applying traditional rendering techniques.

Direct volume rendering is widely used for scientific visualization as it retains all volumetric details, allowing users to analyze inner structures more effectively (see Figure 2.4 for an example).

### 2.9.2 *Mathematical Formulation*

In direct volume rendering, the propagation of radiance (or intensity) along a viewing ray through a participating medium is governed by the radiative transfer equation. For an emission-absorption model, the differential change in radiance $I(t)$ at distance $t$ from the viewpoint is given by:

$$\frac{\mathrm{d}I(t)}{\mathrm{d}t} = g(t) - \kappa(t)I(t), \tag{2.38}$$

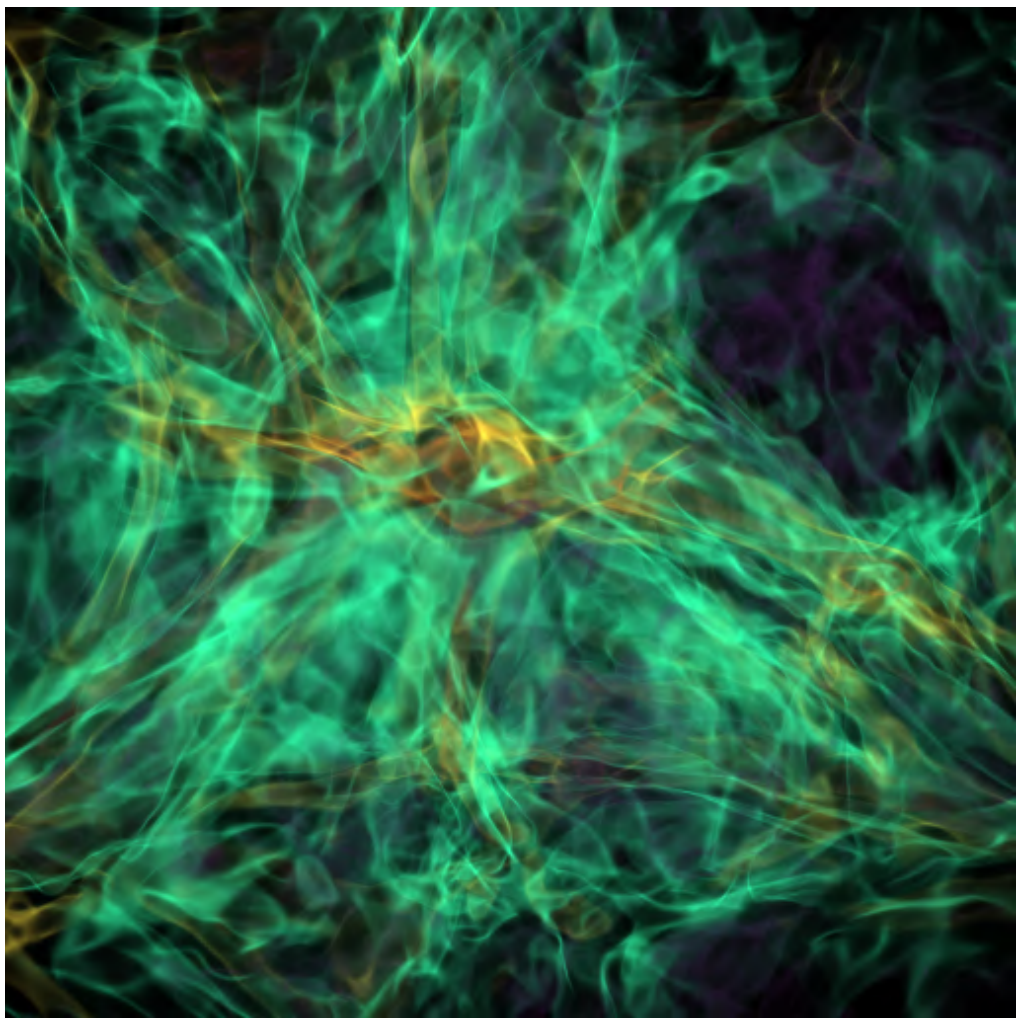

(a) DVR of volumetric data using the yt toolkit.

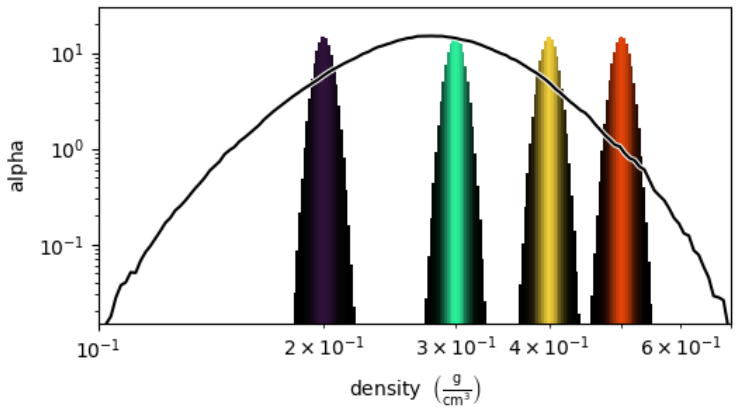

(b) Transfer function used for mapping scalar values to color and opacity.

Figure 2.4: Example of DVR produced with the yt toolkit (Turk et al., 2011). The DVR image (a) visualizes density values of dark matter in an expanding universe, generated by applying the transfer function (b) to map scalar values to color and opacity. This enables exploration of internal structures without explicit surface extraction.

where $g(t)$ denotes the emitted radiance (source term), and $\kappa(t)$ is the extinction coefficient representing attenuation due to absorption and out-scattering.

Integrating this equation from $t = 0$ (the ray entry point, possibly carrying background radiance $I(0)$) to $t = D$ (the viewpoint), we obtain the volume rendering integral (Max, 2002):

$$I(D) = \underbrace{I(0)\, e^{-\int_0^D \kappa(t')\,dt'}}_{\text{absorption term}} + \underbrace{\int_0^D g(t)\, e^{-\int_t^D \kappa(t')\,dt'}\,dt}_{\text{emission term}}\,. \tag{2.39}$$

Here, $I(D)$ is the radiance observed at the viewpoint, and $I(0)$ is the incoming background radiance, both attenuated by the medium. This equation is typically approximated numerically via ray casting or ray marching, where the viewing ray is sampled at discrete intervals to accumulate color and opacity contributions.

### 2.9.3 *Transfer Functions*

A crucial aspect of volume rendering is the design of **transfer functions**, which map scalar values to color and opacity:

$$T : v \mapsto (r, g, b, \alpha), \tag{2.40}$$

where $v$ is the scalar value at a given voxel, and $(r, g, b, \alpha)$ represents the assigned color $(r, g, b)$ and transparency $(\alpha)$. Transfer functions enhance the visibility of specific structures, making them essential for scientific applications.

### 2.9.4 *Neural Networks for Volume Rendering*

Deep learning-based volume rendering methods have emerged to accelerate and enhance visualization. These methods use neural networks to model complex volumetric structures, improving rendering speed and quality. Notable approaches include:

- **Neural Radiance Fields (NeRF)** (Mildenhall et al., 2021): Neural Radiance Fields (NeRF) represent a 3D scene as a continuous function, parameterized by a neural network that maps 3D coordinates $\mathbf{x} = (x, y, z)$ and viewing direction $\mathbf{d} = (\theta, \phi)$ to color $\mathbf{c} = (r, g, b)$ and volume density $\tau$:

  $$F_\theta : (\mathbf{x}, \mathbf{d}) \longrightarrow (\mathbf{c}, \tau). \tag{2.41}$$

  NeRF learns a volumetric radiance field using a multilayer perceptron (MLP), trained via a volume rendering loss that minimizes the discrepancy between rendered pixel values and ground-truth images. Rendering is performed using differentiable ray marching, where the color of a pixel is computed by integrating emitted radiance and accumulated transmittance along a ray $\mathbf{r}(s)$ through the volume. The ray is defined as:

  $$\mathbf{r}(s) = \mathbf{o} + s\mathbf{d},$$

  where $\mathbf{o} \in \mathbb{R}^3$ is the ray origin, $\mathbf{d} \in \mathbb{R}^3$ is the unit direction vector, and $s \in [s_0, s_1]$ is the distance along the ray. The rendered color $I(\mathbf{r})$ is then computed as:

  $$I(\mathbf{r}) = I(s_0)\, e^{-\int_{s_0}^{s_1} \tau(t)\,\mathrm{d}t} + \int_{s_0}^{s_1} \mathbf{c}(\mathbf{r}(s), \mathbf{d})\, \tau(\mathbf{r}(s))\, e^{-\int_{s}^{s_1} \tau(t)\,\mathrm{d}t}\, \mathrm{d}s,$$

where $\tau(\mathbf{r}(s))$ is the volume density at point $\mathbf{r}(s)$, and $\mathbf{c}(\mathbf{r}(s), \mathbf{d})$ is the emitted color at that point in direction $\mathbf{d}$.

This formulation models how light is emitted and absorbed along the ray, capturing the volumetric appearance of the scene. NeRF's implicit representation enables high-quality novel view synthesis but requires extensive sampling and high computational costs.

- **3D Gaussian Splatting (3DGS)** (Kerbl et al., 2023): Unlike NeRF, 3DGS represents a scene using a set of learnable 3D Gaussians, which can be efficiently rasterized and composited. Each Gaussian is defined by a mean $\mu \in \mathbb{R}^3$, covariance $\Sigma \in \mathbb{R}^{3\times 3}$, opacity $\alpha$, and color $\mathbf{c}$. The projected image is obtained by accumulating Gaussian contributions along the viewing direction:

$$I(\mathbf{p}) = \sum_i w_i G(\mathbf{p} - \pi(\mu_i), \Sigma_i), \tag{2.42}$$

  where $G(\cdot, \Sigma_i)$ is a 2D Gaussian kernel, $\pi(\mu_i)$ is the projection (2D coordinates) of the 3D point $\mu_i$ onto the image plane at discrete pixel coordinate $p$, and $w_i$ is a learned weight. 3DGS enables real-time rendering by leveraging GPU rasterization pipelines, making it highly efficient for interactive applications.

  More recent state-of-the-art methods focus on enhancing scalability and generalization. For instance, GS-Cache (Bauer et al., 2025) introduces a novel path-space radiance caching strategy for volumetric rendering by adapting 3D Gaussian splatting as a multi-level, trainable cache, significantly improving rendering quality while maintaining real-time performance and supporting dynamic scene changes.

- **Hybrid Neural Volume Rendering**: This approach combines neural network predictions with traditional ray marching techniques (Zhang et al., 2023). Instead of representing an entire scene as an MLP (as in NeRF), hybrid methods use deep learning models to refine classical volume rendering processes, improving rendering speed and visual fidelity while maintaining interpretability. For example, FoVolNet leverages foveated deep neural networks to accelerate volume rendering by focusing computational resources on perceptually important regions, achieving real-time performance while preserving visual quality (Bauer et al., 2022).

## 2.10 ENSEMBLE VISUALIZATION

Ensemble data arise in various scientific domains where multiple simulations or experiments are conducted under different parameter settings. Unlike traditional single-instance datasets, ensemble data include an additional *member* dimension, representing different realizations of the same physical phenomenon (Wang et al., 2018).

A typical ensemble dataset is defined in terms of:

- **Ensemble instances:** Multiple ensembles (e.g., from different methods or conditions), each capturing a distinct setting.
- **Members:** Individual simulation or experiment runs within an ensemble.
- **Time steps:** Temporal snapshots capturing system evolution over time.
- **Spatial domain:** The grid or mesh defining the physical or simulated space.
- **Variables:** Scalar or vector fields representing physical quantities (e.g., density, temperature, velocity).

This results in a high-dimensional data structure: *ensemble instance* × *member* × *time step* × *location* × *variable*. Such data structures are common in fields such as astrophysics, computational fluid dynamics (CFD), climate science, and high-energy physics, where ensemble simulations help capture variability and uncertainty in complex systems.

### 2.10.1 *Challenges in Ensemble Visualization*

Visualizing ensemble data is challenging due to:

- The **high dimensionality** and large volume of data.
- The need for **efficient comparison** between ensemble members.
- The difficulty of conveying **uncertainty and variability**.
- The necessity of **interactive exploration** to extract meaningful insights.

### 2.10.2 *Techniques for Ensemble Visualization*

Traditional visualization approaches in ensemble visualization often rely on aggregation, statistical summarization, or compression techniques (Wang et al., 2018). These methods reduce the complexity of ensemble data while preserving essential patterns and trends.

Several visualization techniques have been developed to handle ensemble datasets:

- **Side-by-side visualization:** Displays different members in parallel, allowing direct visual comparison.

- **Linked brushing:** Interactively highlights corresponding data points across multiple views.
- **Spaghetti plots:** Used for tracking multiple trajectories or paths within an ensemble.
- **Glyph-based visualization:** Encodes ensemble properties using graphical symbols.
- **Probability density plots:** Represent distributions of scalar values across ensemble members.
- **Attribute blocks:** Use color-coded representations for statistical summaries.

**Machine Learning for Ensemble Visualization.** Recent advancements in machine learning have further improved ensemble visualization by enabling:

- **Dimensionality reduction techniques** (e.g., t-SNE, UMAP, autoencoders) to project high-dimensional ensemble data into interpretable low-dimensional spaces.
- **Neural networks for feature extraction** to identify important patterns and similarities among ensemble members.
- **Uncertainty quantification** using probabilistic deep learning models to enhance ensemble representation.
- **Flow estimation and prediction** to reconstruct missing temporal or spatial data within ensemble simulations.

By integrating these techniques, modern ensemble visualization frameworks enable more effective exploration, interpretation, and decision-making in scientific, engineering, and other data-driven applications.

## 2.11 SUMMARY AND OUTLOOK

Building on the foundations laid in this chapter, the next Chapter 3 delves into the specific techniques developed in this work to improve ensemble visualization. We focus on designing and evaluating autoencoder-based methods tailored for expressive and structure-preserving dimensionality reduction of spatial ensembles. By enhancing model architectures, training strategies, and evaluation protocols, we aim to advance the interpretability and effectiveness of machine learning in scientific visualization tasks.

# 3 EVALUATION AND SELECTION OF AUTOENCODERS FOR EXPRESSIVE DIMENSIONALITY REDUCTION OF SPATIAL ENSEMBLES

*In this chapter, we focus on evaluating and selecting autoencoder configurations for expressive dimensionality reduction of spatial ensembles. We first propose several improvements to existing autoencoder-based approaches, originally designed for feature extraction, to optimize their ability to capture spatial structures while achieving an effective trade-off between accuracy and dimensionality reduction. These enhancements include architecture selection heuristics, guided parameter tuning based on projection quality metrics, and a data-efficient approach leveraging partially labeled ensemble members. We then establish a benchmarking framework to assess the impact of different autoencoder variants across diverse ensemble datasets. Using this benchmark, we identify optimal configurations for feature extraction in spatial ensembles. A comprehensive evaluation across different datasets demonstrates that autoencoder-guided projections can better preserve spatial and spatio-temporal relationships, outperforming traditional dimensionality reduction techniques in specific cases.*

## 3.1 INTRODUCTION

Driven by technological advances, scientific ensembles of increasing size are obtained from simulations and experiments. They offer significant potential for new insights in various domains across engineering and natural sciences, but their analysis induces many challenges (Wang et al., 2018). Dimensionality reduction (DR) techniques have been successfully applied for analyzing large sample collections (e.g., Kehrer and Hauser (2013)), and especially 2D projections widely used to provide a visual impression of the data distribution (Liu et al., 2016). However, when applied directly to spatial data, the expressiveness of projection techniques like, e.g., Uniform Manifold Approximation and Projection (UMAP) (McInnes et al., 2018) is generally negatively impacted by the high dimensionality and sparsity of the data.

This work explores the usage of unsupervised feature learning techniques to produce more suitable data representations for DR of spatial data, and specifically for 2D projections. We investigate standard and

This chapter is based on the published paper "Evaluation and Selection of Autoencoders for Expressive Dimensionality Reduction of Spatial Ensembles" (Gadirov et al., 2021) sparse *autoencoders* (AE), as well as more advanced versions such as *Sliced-Wasserstein* and *$\beta$-Variational* autoencoders (SWAE and $\beta$-VAE). Our study with two different scientific ensembles demonstrates that they improve expressiveness in comparison to directly projecting the spatial data via UMAP. However, it also shows that the performance of autoencoder variants varies depending on the data characteristics, i.e., it is not clear *a priori* which one to choose to adequately capture what is of interest in the data. To address this, we propose to employ complementary metrics quantifying the quality of a projection and selecting a specific autoencoder variant based on Pareto efficiency. These metrics assess the quality of the projection based on labels (generally provided by an expert). While labels can be prohibitively expensive to obtain for the full ensemble, we demonstrate that a small subset of labeled members is already sufficient to yield expressive results. In the context of this chapter, we define expressiveness as the ability of a 2D projection to preserve meaningful structures present in the original high-dimensional data. This includes the accurate separation of clusters, the maintenance of spatial or spatio-temporal similarity relationships, and the avoidance of artifacts due to sparsity or dimensional collapse. An expressive projection enables better visual interpretability and supports downstream tasks such as clustering or classification.

We consider (1) the study of several autoencoder variants for dimensionality reduction with diverse scientific ensembles, (2) the evaluation of projection metric stability for small partial labelings, and (3) the Pareto-efficient selection of a variant on this basis to be the main contributions of this work.

## 3.2 RELATED WORK

**Ensemble visualization.** The analysis of ensemble data generally is a challenging visualization task (Obermaier and Joy, 2014). Potter *et al.* (Potter et al., 2009) as well as Sanyal *et al.* (Sanyal et al., 2010) proposed early approaches to study climate ensembles, while Waser *et al.* (Waser et al., 2010) described a system for the interactive steering of simulation ensembles. Kehrer *et al.* (Kehrer and Hauser, 2013), Sedlmair *et al.* (Sedlmair et al., 2014), and Wang *et al.* (Wang et al., 2018) provided detailed surveys of techniques in the area. Bruckner and Möller (Bruckner and Möller, 2010) employ squared differences to explore the visual effects simulation space, Hummel*et al.* (Hummel et al., 2013) compute region similarity via joint variance, and Kumpf*et al.* (Kumpf et al., 2018) track statistically-coherent regions using optical flow. Hao *et al.* (Hao et al., 2016) calculate shape similarities for particle data using an octree structure, while He *et al.* (He et al., 2018) employ surface density estimates for distances between surfaces. Fofonov *et al.* (Fofonov et al., 2015) propose fast isocontour calculation for visual representation of ensembles. For projection of high-dimensional data, Vernier *et al.* (Vernier et al., 2020) and Espadoto *et al.* (Espadoto et al., 2019) propose spatial and temporal stability metrics to evaluate the quality of PCA, t-SNE, UMAP, and Autoencoders. Related approaches include Neural Network Projection (NNP) (Espadoto et al., 2020), which learns projections from data and user-defined 2D layouts, and HyperNP (Appleby et al., 2022), which extends this idea to model ensembles of projections under varying hyperparameter combinations. Bertini *et al.* (Bertini et al., 2011) presented a systematic analysis of quality metrics supporting exploration. We, however, study how autoencoders impact the 2D projection quality when combined with traditional DR techniques. In contrast to prior work, we focus on the role of autoencoder-based feature extraction—used either as a preprocessing step or as a replacement for DR—in enhancing the expressiveness and structure of low-dimensional embeddings.

**Autoencoder-based Feature Extraction.** Several works confirm the ability of autoencoders to extract expressive features for ensemble data. Hinton and Salakhutdinov (2006) first demonstrated that autoencoders can be utilized for DR and can be applied to large datasets. Plaut (2018) performed principal component analysis using a linear autoencoder. Han *et al.* (Han et al., 2018) proposed *FlowNet*, an autoencoder-based framework for extracting flow features such as streamlines and stream surfaces. Their experiments on various volumetric flow datasets demonstrate an interactive interface that supports exploration of flow lines and surfaces through clustering, filtering, and selection. Jain *et al.* (Jain et al., 2017) utilized a deep convolutional autoencoder to obtain a compact representation of multivariate time-varying volumes by learning high-level features. Lekschas *et al.* (Lekschas et al., 2020) developed a convolutional autoencoder-based technique *PEAX* for interactive visual pattern search. Guo *et al.* (Guo et al., 2017) developed a deep convolutional autoencoder minimizing the reconstruction and clustering losses for end-to-end learning of embedded features for clustering. Ge *et al.* (Ge et al., 2019) achieved state-of-the-art clustering performance on MNIST via dual adversarial autoencoders. He *et al.* (He et al., 2020) proposed a deep learning approach for comparison of multiple ensembles. Guo *et al.* (Guo et al., 2020) developed a visual analytics system based on autoencoders for medical records. Way and Greene (2017) utilize variational autoencoders to extract biologically relevant features from gene expression data. We also employ deep convolutional autoencoders to enable visual exploration of scientific data, but focus on a study of autoencoder variants and model selection in a partially labeled scenario.

## 3.3 STUDY SETUP, METRICS AND SELECTION

Standard DR techniques, such as PCA, t-SNE, and UMAP (Section 2.5), become less effective when applied directly to high-dimensional ensemble data (e.g., Hinton and Salakhutdinov (2006), Way and Greene (2017)).

0 1 2 3 4 unlabeled

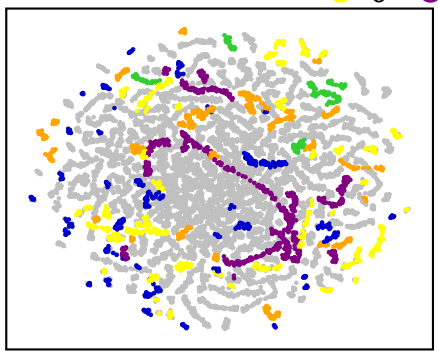

(a) UMAP only (Baseline), n.h. = 0.975, s. = -0.052

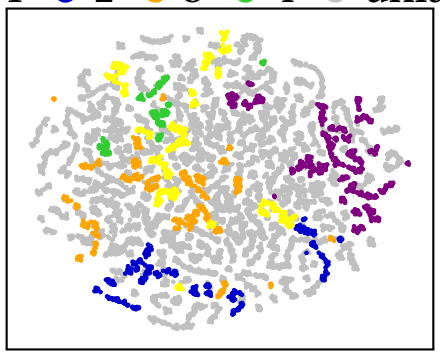

(b) 2D Sparse AE 64, n.h. = 0.978, s. = -0.036

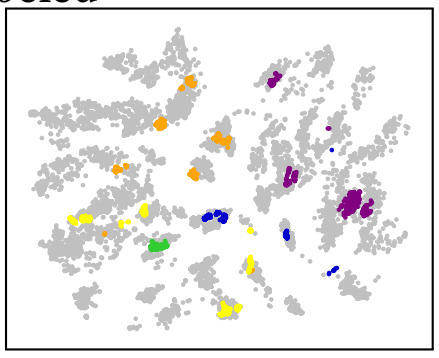

(c) 2D SWAE 2 *, n.h. = 0.994, s. = 0.373

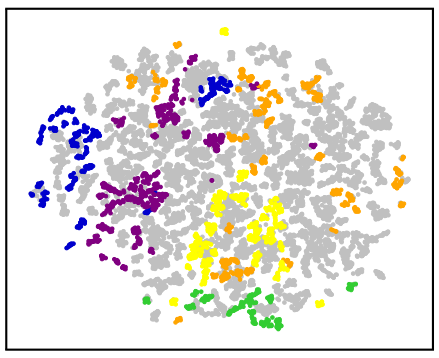

(d) 2D SWAE 32, n.h. = 0.952, s. = -0.025

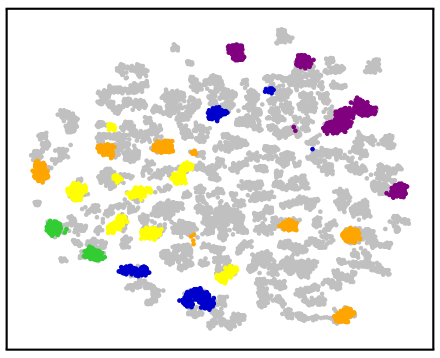

(e) 2D VAE 128, n.h. = 0.992, s. = 0.033

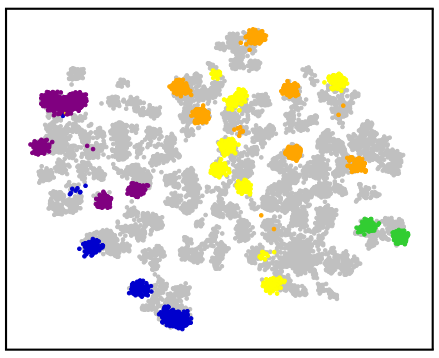

(f) 2D $\beta(2)$-VAE 128, n.h. = 0.993, s. = 0.034

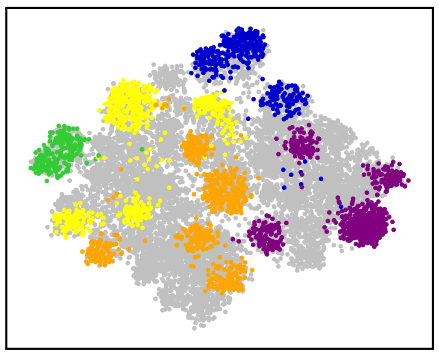

(g) 2D $\beta(4)$-VAE 256, n.h. = 0.975, s. = 0.190

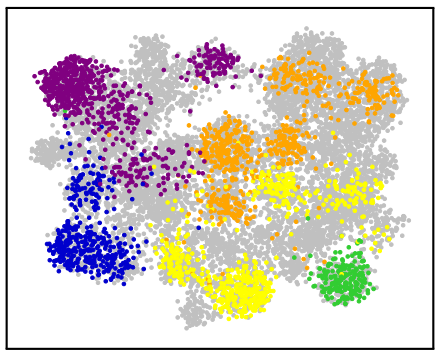

(h) 2D $\beta(8)$-VAE 32 *, n.h. = 0.938, s. = 0.348

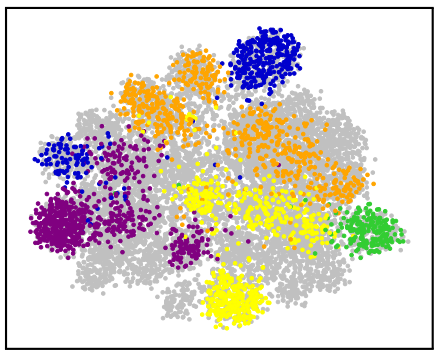

(i) 2D $\beta(10)$-VAE 128, n.h. = 0.916, s. = 0.313

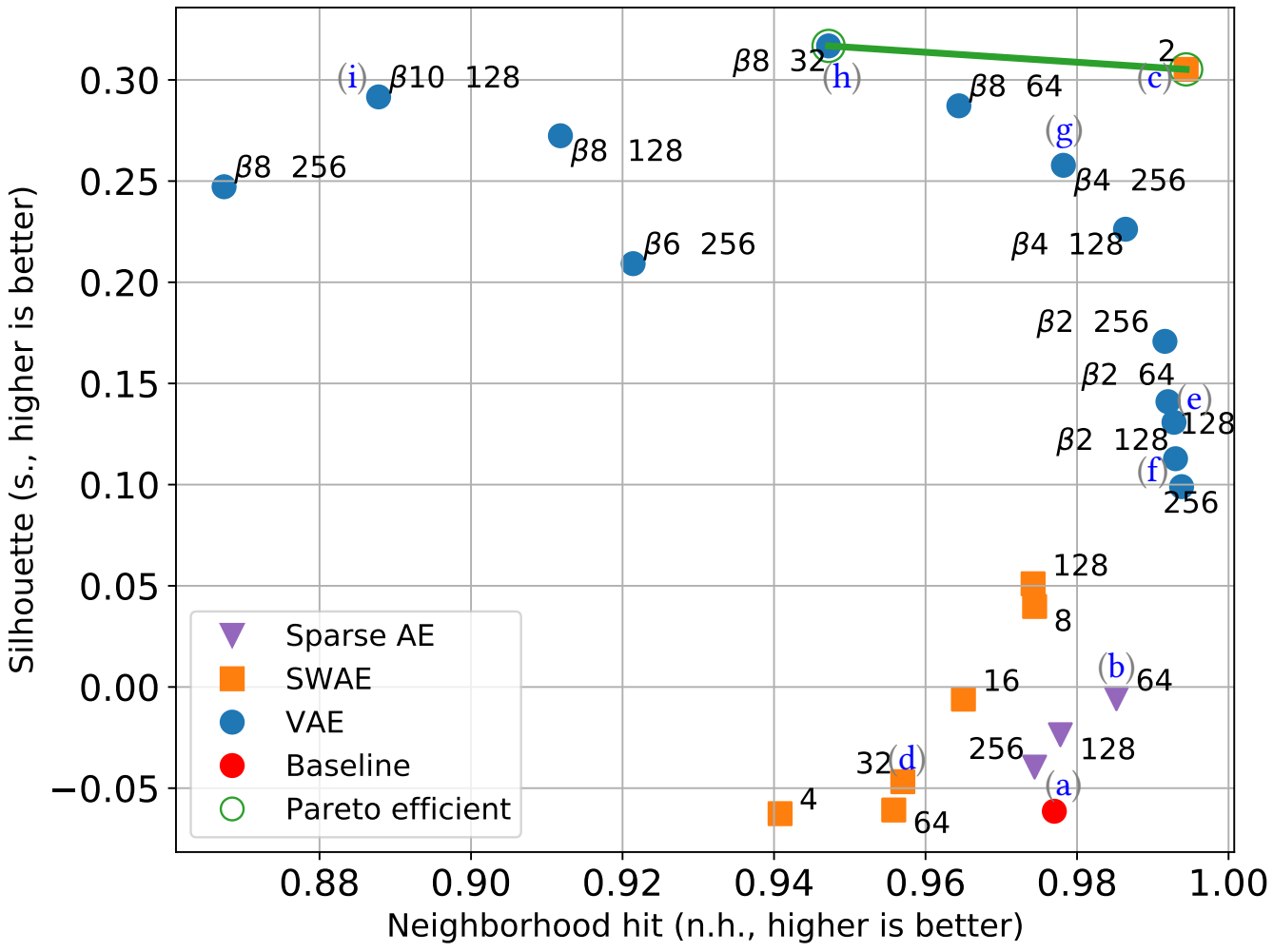

(j) Pareto frontier of autoencoder variants (projection view links in gray)

Figure 3.1: (a–i) Applying autoencoders for feature learning prior to 2D projection enhances result for spatial ensembles (here: channel structures in soil from MCMC, Figures 3.3a and 3.3c), but the outcome highly depends on architectures, parameters, and underlying data. A partial labeling of 1% suffices to yield expressive quality metrics, and (j) allows exploration of suitable variants (marked with *) on the Pareto frontier.

To overcome this limitation, we first apply autoencoders to compress the ensemble data into a lower-dimensional latent space, and then construct a 2D projection from this representation. An autoencoder is a neural network for unsupervised learning of compact data encodings (Section 2.4), consisting of an encoder that compresses the input and a decoder that reconstructs it as accurately as possible. We explore multiple autoencoder architectures and evaluate their performance both visually and quantitatively. The quantitative metrics further serve as the basis for Pareto-efficient model selection.

**Scientific Ensemble Datasets.** We consider two ensemble datasets in our study. The first dataset depicts channel structures in soil from Markov Chain Monte Carlo (MCMC), consisting of independent members generated during simulation (Reuschen et al., 2020). The images are monochrome with a spatial resolution of $50 \times 50$, meaning each image is represented as a 2D array of 2,500 scalar values (i.e., one scalar per pixel), which serves as the input dimensionality for our proposed method. In total, there are 95K images. The second dataset Drop Dynamics stems from a physical experiment to study the impact of a droplet with a film (Geppert et al., 2016). The captured experiment images, like those in the previous dataset, are monochrome with a resolution of $160 \times 224$, resulting in 35,840 scalar values per image. In total, there are 135K images from 1K members. Subsets of the members of both ensembles are exemplified in the grid views in Figure 3.3. Examples of different timesteps from various members of Drop Dynamics are in Figure 3.2.

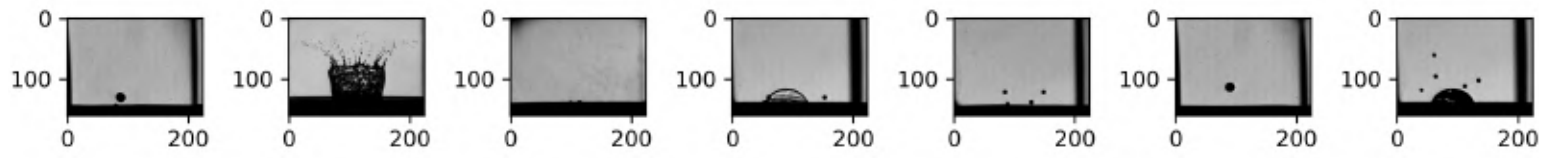

Figure 3.2: Drop Dynamics samples from different ensemble members at different timesteps.

A subset from both datasets was manually labeled, as required by the evaluation metrics we employ, which rely on class-based separation performance. Labeling is based on distinct behavior types observed within the ensembles, and was performed by grouping images with qualitatively similar patterns into discrete class labels. For the MCMC dataset, we defined five categorical classes (● ● ● ● ●), which reflect qualitatively different channel structures discovered through visual analysis of the ensemble and were cross-validated with co-authors for labeling consistency. For the Drop Dynamics dataset, we adopted eight pre-defined classes from the original work by Geppert *et al.* (Geppert et al., 2016): "bubble", "bubble-splash", "column", "crown", "crown-splash", "splash", "drop", and "none" (for images not fitting any other category). In total, we use 2.5K labeled images for MCMC and 7.2K for Drop Dynamics. On both projection views (Figures 3.1 and 3.5), gray points (●) indicate unlabeled samples.

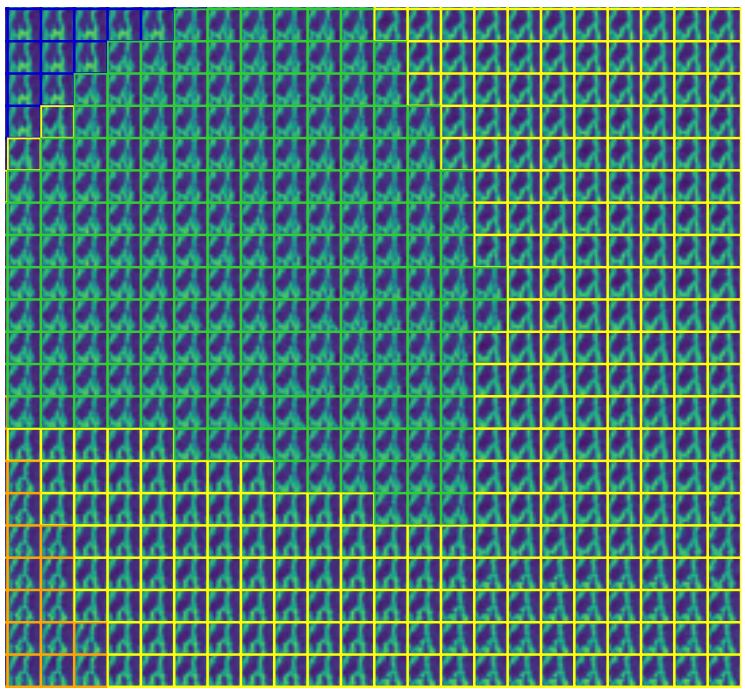

(a) MCMC (corresponding to Fig. 3.1c)

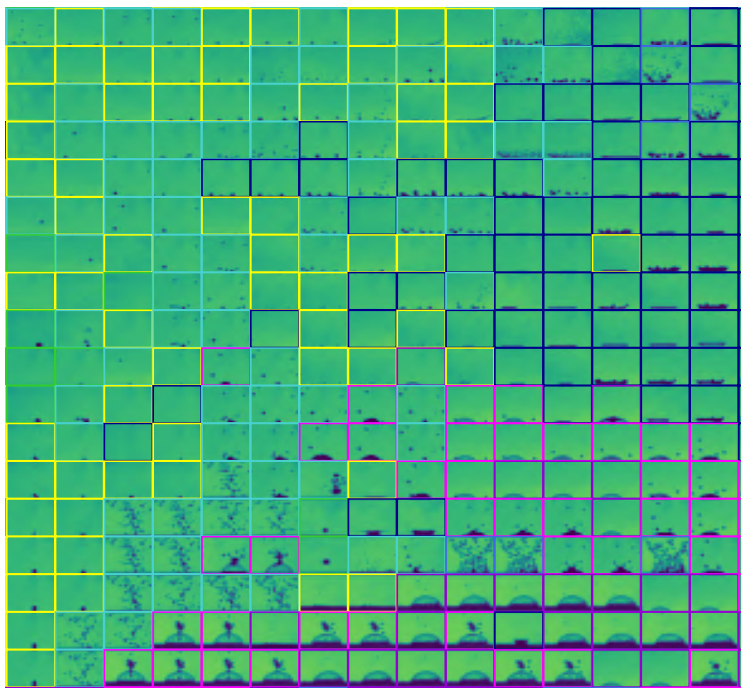

(b) Drop Dynamics (from Fig. 3.5e)

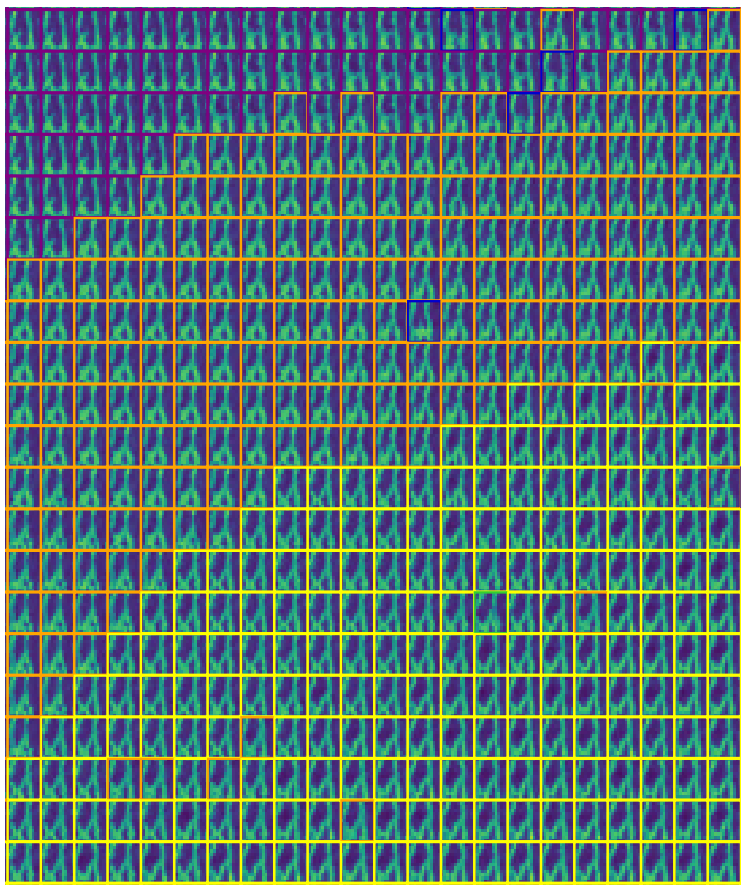

(c) MCMC (corresponding to Fig. 3.1h)

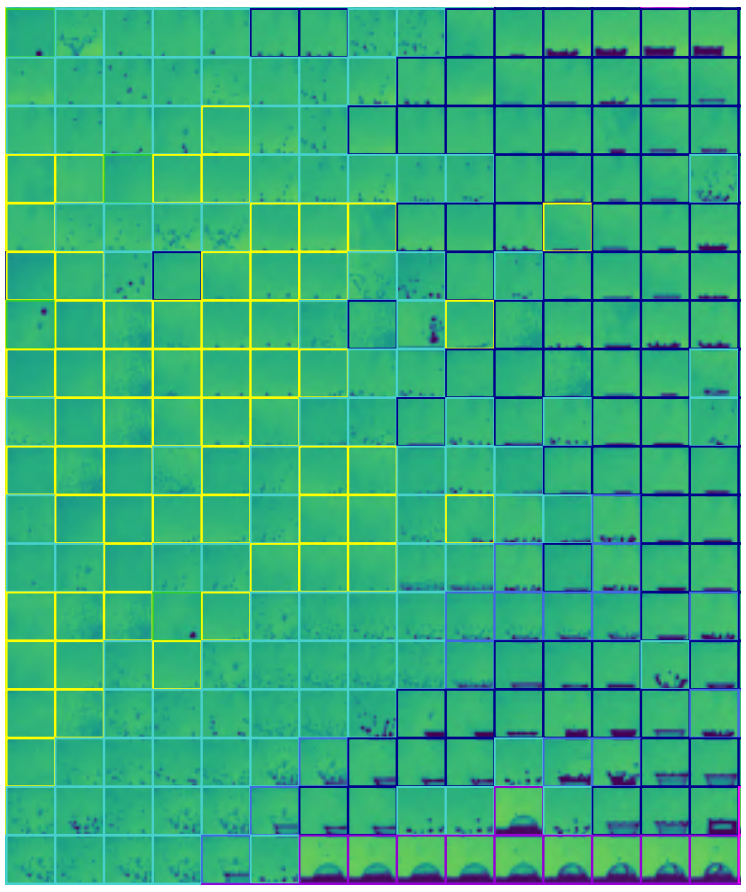

(d) Drop Dynamics (from Fig. 3.5g)

Figure 3.3: Grid views with colored frames around each image representing labels. The new position in the grid was found via linear assignment from a 2D projection to the grid (Quadrianto et al., 2009).

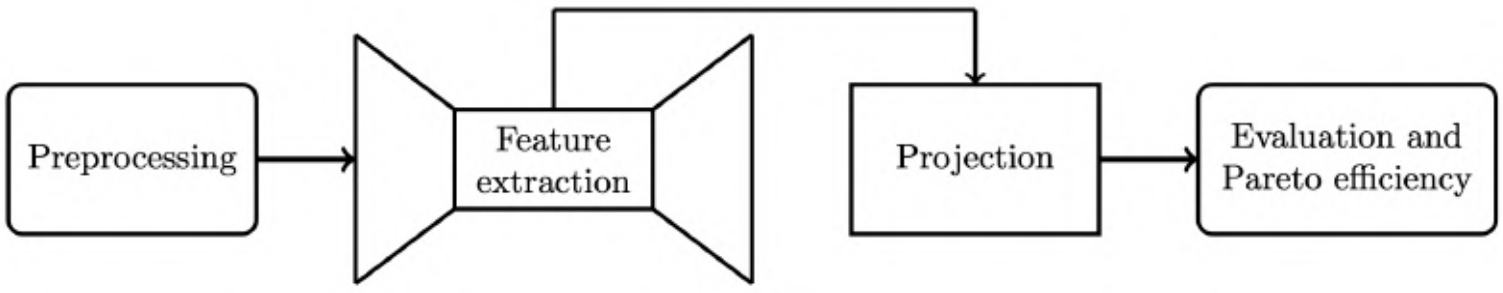

Figure 3.4: Overview of our approach. We first preprocess the data and feed it into an autoencoder (left trapezoid) to learn latent features. Although the autoencoder also reconstructs the input (right trapezoid), our focus is the extracted latent representation (center), which is then passed to projection (dimensionality-reduction) techniques to obtain a visualization. The resulting visualizations are evaluated with our metrics to trace a Pareto frontier of solutions. The pipeline supports multiple autoencoder variants.

### 3.3.1 *Autoencoder-based Feature Extraction*

In this study, we explore multiple variants of autoencoders (AEs) for feature extraction, including standard autoencoders, sparse autoencoders, variational autoencoders (VAE) (Kingma and Welling, 2014), and their constrained version, $\beta$-VAE (Higgins et al., 2017). Additionally, we consider the sliced-Wasserstein autoencoder (SWAE) (Kolouri et al., 2018), which is based on Wasserstein autoencoders (Tolstikhin et al., 2018) and achieves improved reconstruction quality over standard VAEs while maintaining similar latent space properties.

All considered autoencoder architectures (AE, SWAE, VAE, and $\beta$-VAE) follow a symmetric design, where the decoder mirrors the encoder. Both the encoder and decoder are implemented using deep neural networks comprising multiple hidden convolutional and fully connected layers. An overview of this pipeline is illustrated in Figure 3.4.

#### 3.3.1.1 *Architecture and Training Details*

For autoencoders (AEs) and variational autoencoders (VAEs), input images are progressively downsampled using convolutional layers. A stride of 2 is applied in each spatial dimension for 2D inputs, while a stride of 3 is used for 3D inputs. Each convolutional layer employs a kernel size of 3 across all dimensions and contains 64 filters with zero padding. The Adam optimizer (Kingma and Ba, 2014) is used for training, with a learning rate of 0.0005. ReLU activation functions are applied throughout, with random uniform weight initialization. For sparse autoencoders, L1 and L2 regularization is incorporated to enforce sparsity in the latent representation.

For the MCMC dataset, which lacks temporal information, only 2D convolutional models were used. These models were trained and validated on a dataset of 20K unlabeled images. In contrast, for the spatio-temporal Drop Dynamics dataset, both 2D and 3D autoencoders were employed. These models were trained and validated on a dataset of 15K unlabeled images.

Since this study utilizes unsupervised machine learning and dimensionality reduction techniques, labels are not used for training. Instead, they are employed solely for evaluation purposes, as discussed in the metrics section below.

EXAMPLE ARCHITECTURE. Table 3.1 presents an example of the 3D autoencoder (AE) and variational autoencoder (VAE) architecture used for the Drop Dynamics ensemble dataset. The network consists of a series of convolutional layers that progressively downsample the input, followed by a bottleneck layer, where the latent representation is learned. The key difference between AE and VAE is highlighted in the bottleneck, where VAE employs two separate dense layers for the mean ($\mu$) and logarithm of variance ($\log \sigma$) to facilitate the reparameterization trick. The decoder then reconstructs the input using transposed convolutional layers, gradually upsampling the representation back to its original resolution. This architecture enables efficient feature extraction while maintaining the ability to reconstruct high-dimensional data.

Table 3.1: 3D AE/VAE architecture used for Drop Dynamics ensemble. Differences in the latent space between AE and VAE are explicitly marked in the middle rows.

| **Layer type** | **Output Shape** | **Details** |
|---|---|---|
| Input | (batch size, 3, h, w, 1) | height = h, width = w |
| Conv3D | (batch size, 1, h/2, w/2, 64) | kernel size = (3, 3, 3), stride = (3, 2, 2) |
| Conv3D | (batch size, 1, h/4, w/4, 64) | kernel size = (1, 3, 3), stride = (1, 2, 2) |
| Conv3D | (batch size, 1, h/8, w/8, 64) | kernel size = (1, 3, 3), stride = (1, 2, 2) |
| Conv3D | (batch size, 1, h/16, w/16, 64) | kernel size = (1, 3, 3), stride = (1, 2, 2) |
| Flatten | (batch size, 1, (h/16) · (w/16) · 64) | reshape before dense layer |
| Dense | (batch size, num. of units) | first dense layer of encoder |
| *AE*: Dense<br>*VAE*: Dense ($\mu, \log \sigma$)<br>*VAE*: Sample $z$ | (batch size, latent dimension)<br>(batch size, latent dimension)<br>(batch size, latent dimension) | second dense layer<br>two parallel dense layers for VAE<br>reparameterization trick for VAE |
| Dense | (batch size, 1, (h/16) · (w/16) · 64) | first dense layer of decoder |
| Reshape | (batch size, 1, (h/16) · (w/16) · 64) | reshape before deconvolutions |
| Conv3DTranspose | (batch size, 1, h/8, w/8, 64) | kernel size = (1, 3, 3), stride = (1, 2, 2) |
| Conv3DTranspose | (batch size, 1, h/4, w/4, 64) | kernel size = (1, 3, 3), stride = (1, 2, 2) |
| Conv3DTranspose | (batch size, 1, h/2, w/2, 64) | kernel size = (1, 3, 3), stride = (1, 2, 2) |
| Conv3DTranspose | (batch size, 1, h, w, 64) | kernel size = (3, 3, 3), stride = (3, 2, 2) |
| Conv3DTranspose | (batch size, 1, h, w, 1) | kernel size = (3, 3, 3), stride = (1, 1, 1) |

#### 3.3.1.2 *Objective Functions for Autoencoder Variants*

STANDARD AUTOENCODER A standard AE is trained using a reconstruction loss, typically measured by mean squared error (MSE) between the input $x$ and the reconstructed output $x'$:

$$\mathcal{L}(x, x') = \|x - x'\|^2 . \tag{3.1}$$

The network is trained using standard backpropagation techniques (Hecht-Nielsen, 1992).

SPARSE AUTOENCODER To enforce sparsity in the latent representation, we apply L1 regularization to the weight matrix of the final dense layer in the encoder and L2 regularization to prevent overfitting (Ng et al., 2011).

SLICED-WASSERSTEIN AUTOENCODER (SWAE) SWAE is a generative model that does not require adversarial training (Kolouri et al., 2018). The SWAE objective function minimizes the Wasserstein distance between the input distribution $p_X$ and the reconstructed distribution $p_{X'}$, while also regularizing the latent space distribution $p_Z$ to follow a uniform prior $q_Z$ using the sliced-Wasserstein distance:

$$\arg\min_{\phi,\theta} \quad W_c(p_X, p_{X'}) + \lambda SW_c(p_Z, q_Z), \tag{3.2}$$

where $\phi$ and $\theta$ are the parameters of the encoder and decoder, respectively.

VARIATIONAL AUTOENCODER (VAE) VAEs introduce a probabilistic latent space by assuming a Gaussian prior on the latent variable distribution (Kingma and Welling, 2014). The objective function consists of a reconstruction term and a KL-divergence regularization term, forming the variational lower bound:

$$\mathcal{L}(\theta, \phi; x) = -D_{KL}(q_\phi(z|x) \| p_\theta(z)) + \mathbb{E}_{q_\phi(z|x)}[\log p_\theta(x|z)]. \tag{3.3}$$

Here, the prior $p_\theta(z)$ is assumed to follow a unit Gaussian distribution $\mathcal{N}(0, I)$, while the approximate posterior $q_\phi(z|x)$ is parameterized as a Gaussian $\mathcal{N}(\mu, \sigma^2)$, with $\mu$ and $\sigma$ learned by the encoder.

$\beta$-VARIATIONAL AUTOENCODER ($\beta$-VAE) $\beta$-VAE extends the standard VAE by introducing a weighting factor $\beta$ to control the balance between reconstruction quality and latent space regularization (Higgins et al., 2017), as described in Section 2.4.5.14:

$$\mathcal{L}(\theta, \phi; x, z, \beta) = -\mathbb{E}_{q_\phi(z|x)}[\log p_\theta(x|z)] + \beta D_{KL}(q_\phi(z|x) \| p(z)). \tag{3.4}$$

The choice of $\beta$ determines the behavior of the model:

- $\beta < 1$: Behaves more like a standard AE, prioritizing reconstruction quality.
- $\beta = 1$: Reduces to the standard VAE.
- $\beta > 1$: Encourages a more structured and disentangled latent space at the cost of reconstruction accuracy.

In our experiments, we tested $\beta$ values ranging from 0.1 to 100 to explore its effect on the latent space structure.

### 3.3.2 *Projection to 2D Space*

After transforming the ensemble data from physical to feature space, we obtain a latent vector for each data sample. These feature vectors are subsequently projected to a 2D space using DR techniques. In general, we find that directly reducing the dimensionality of ensembles to 2D using autoencoders is inefficient for most of the models since the autoencoder cannot reasonably reconstruct the input which is an indication of poorly learned features (although there are exceptions, see discussion in Section 3.4. In the following, we focus on UMAP projections, which consistently outperformed other dimensionality reduction techniques such as t-SNE and PCA in our initial comparative analysis using the same labeled subsets. This outperformance of UMAP is quantified using the projection evaluation metrics discussed next in Section 3.3.3. UMAP is a non-linear technique and uses a smoothed version of k-nearest neighbors distance. We utilize UMAP with *min. distance = 1.0*, deviating from the default to emphasize global structure preservation over local clustering. This value controls how close points are located on the newly created 2D map and produces visually less overlapping projections.

### 3.3.3 *Metrics*

We use four commonly used and complementary metrics to evaluate the quality of the 2D projections: *neighborhood hit*, *silhouette*, *separability*, and *spread*. While neighborhood hit and silhouette capture the local and global clustering structure, separability and spread assess the geometric distribution and spacing of clusters, providing a more complete assessment of projection quality.

The *neighborhood hit* metric (Vernier et al., 2020) is based on k-nearest neighbors (Peterson, 2009) ($k = 17$ in our setting; values ranging from 7 to 27 yield similar results), and since it computes the fraction of neighbors belonging to the same class for each labeled data point, it provides an accurate measure of cluster purity and local separation. The output is in the range of [0, 1], where *higher values* indicate better preservation of local neighborhood structure in the projection.

The *silhouette* metric (Rousseeuw, 1987) measures how similar a point is to its own cluster compared to other clusters, using Euclidean distance. It computes the average distance between a point and all other points in the same cluster (intra-cluster distance), and compares it to the smallest average distance between that point and all points in any other cluster (nearest inter-cluster distance). The resulting silhouette score lies in the range [-1, 1], where higher values indicate more compact and well-separated clusters. This metric is especially useful for identifying well-defined cluster boundaries and assessing how confidently a point belongs to its assigned cluster.

The *separability* metric quantifies the geometric distance between clusters by computing the pairwise distances between cluster centroids. The final score is derived from the average inter-cluster distance, normalized by the intra-cluster spread. Higher values indicate that clusters are not only well separated but also compact, making them easier to distinguish in 2D space. This metric is especially useful for datasets where visual separation is more important than local neighborhood consistency.

The *spread* metric assesses the overall spatial compactness of the projected data. It computes the standard deviation of the projected points along each axis and aggregates them to determine how tightly the data is clustered in the 2D embedding space. Lower spread values indicate more compact and concentrated projections, helping reduce excessive dispersion and improve visual clarity. While some level of dispersion is necessary to avoid overplotting, overly spread-out projections may dilute structure and hinder interpretability.

### 3.3.4 *Pareto Efficient Selection*

We identify the top-performing models for extracting meaningful low-dimensional representations based on multi-objective optimality criteria (Pareto frontier). To this end, we use two evaluation metrics—neighborhood hit and silhouette score (higher values indicate better performance)—and plot them in a 2D space to assess clustering quality in the final projections. The Pareto frontier highlights models that achieve an optimal trade-off between these metrics, from which a suitable candidate can be selected for further inspection, visualization, and analysis of the ensemble data projection.

## 3.4 EVALUATION

We now discuss the 2D projection results of our ensembles, illustrate Pareto-efficient model selection, and assess the robustness of evaluation metrics when computed from labeled subsets (see Table 3.2 and Table 3.3 for a comparison of all model variants).

**MCMC (Figure 3.1).** In the baseline method that uses UMAP directly (a), we can observe well separated but dispersed clusters of (reflected by high neighborhood hit but low silhouette). The projection results improved for all classes with the 2D sparse autoencoder (b). Note that the number after the autoencoder type indicates the dimensionality of the latent space. However, classes partitioned into different clusters and yielding highly irregular shapes can still be observed. This is reflected by low values in the silhouette which position AE results between the baseline and $\beta$-VAEs in Figure 3.1j. For SWAE results (c and d), we observe an improvement with all classes being more tightly clustered. Both metrics are improved because of the influence of the SWAE objective function. Interestingly, SWAE with direct projection to 2D (not using UMAP) also yields a good projection (c). 2D VAE (e) and 2D $\beta(2)$-VAE (f) further improve the results. Here, the number in $\beta()$ denotes the value of the $\beta$ parameter used in the $\beta$-VAE configuration, which controls the strength of the regularization term in the loss function. We see well-separated clusters and even higher neighborhood hit and silhouette values. Models like 2D $\beta(4)$-, $\beta(8)$-, and $\beta(10)$-VAE (g, h, and i) create clusters in the shape of Gaussian distributions, further improving metric scores. These are also on the Pareto frontier ( Figure 3.1j). Interestingly, $\beta(8)$-VAE with a comparably low latent space dimensionality of 32 is able to extract features properly. Overall, we observe a significant benefit in performing feature extraction on the MCMC ensemble. Visualizations show that AE, SWAE, and ($\beta$-)VAE with different $\beta$ values all outperform the baseline. In the final projection of almost all models, we see data points belonging to the same class located close to each other, forming clusters of the same classes. In the case of $\beta$-VAE, properly selected values of $\beta$ (not too high) can improve the results and lead to visually pleasing projections.

**Drop Dynamics ( Figure 3.5).** The baseline result (a) successfully produces clusters of similar samples and thus achieves a relatively high neighborhood hit. This is clearly noticeable in the case of classes "bubble", "bubble-splash", "splash", and samples without any behavior category ("none"). However, it scatters the samples of the same type across many small clusters, which is highlighted by the poor silhouette values. In the case of simple 2D Sparse AE with latent dimensionality 64 (b), the model also cannot group all data points belonging to the same class into one cluster. We can see the "bubble", "bubble-splash", and "splash" type samples present in different parts of the final projection, which is based on the latent space learned by the autoencoder. In (c), clusters start to appear in the form of Gaussian distributions ("bubble"), due to the KL divergence regularization term in the VAE loss calculation, improving the silhouette score. In (d), corresponding to the 3D AE, we observe more cohesive and well-separated clusters for the "crown" and "drop" classes, which is reflected in the higher neighborhood hit score. Elongated clusters still persist, since there is no KL regularization applied in this case.

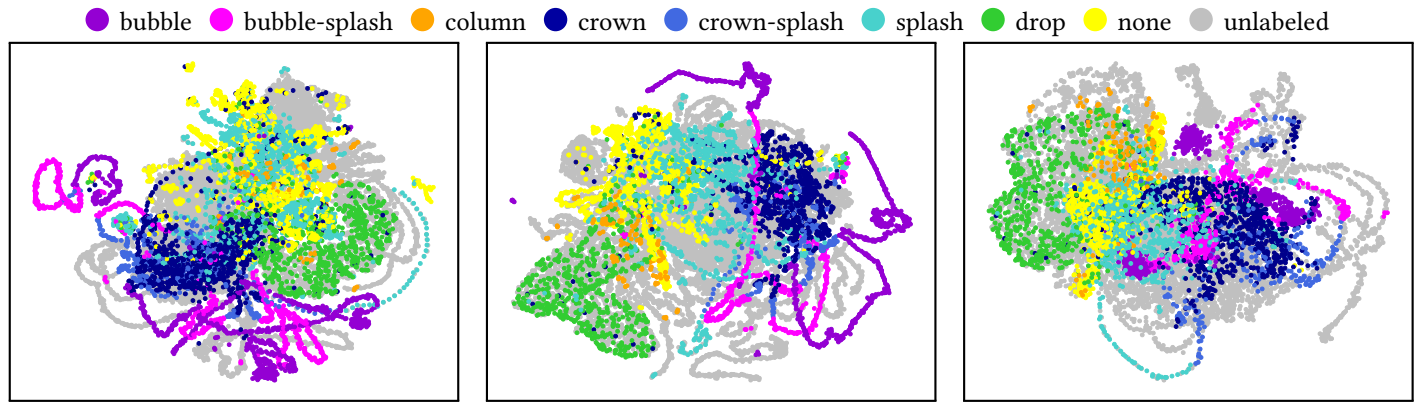

(a) UMAP only (Baseline), n.h. = 0.638, s. = -0.096

(b) 2D Sparse AE 64, n.h. = 0.686, s. = -0.076

(c) 2D VAE 256, n.h. = 0.6, s. = -0.023

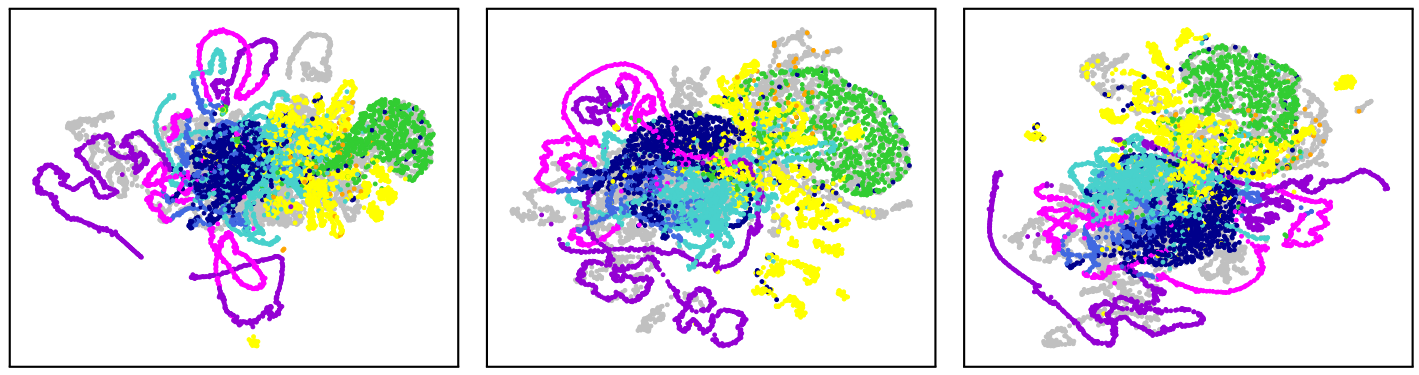

(d) 3D AE 256 *, n.h. = 0.775, s. = -0.075

(e) 3D SWAE 32 *, n.h. = 0.761, s. = -0.069

(f) 3D SWAE 128, n.h. = 0.785, s. = -0.074

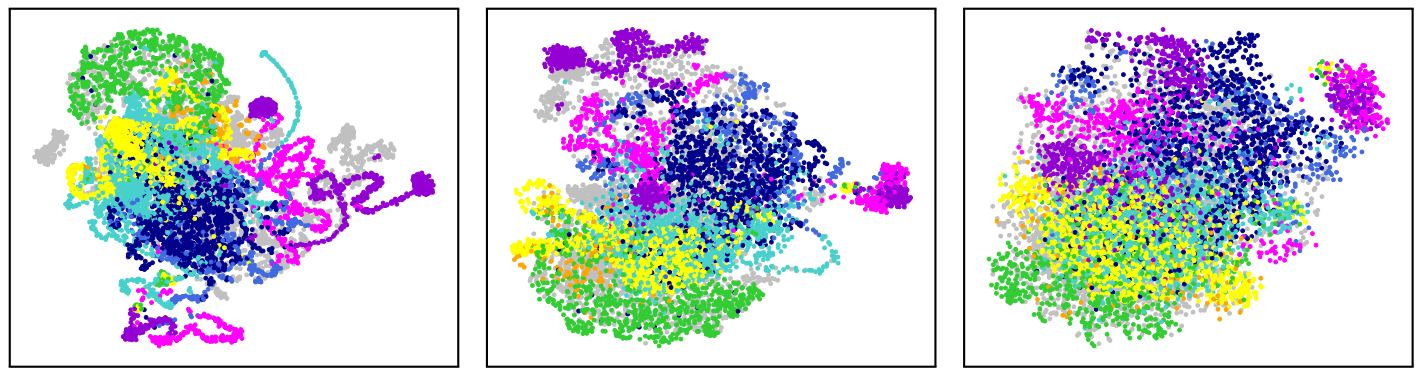

(g) 3D $\beta(0.1)$-VAE 256 *, n.h. = 0.720, s. = 0.012

(h) 3D VAE 256, n.h. = 0.613, s. = -0.002

(i) 3D $\beta(4)$-VAE 256, n.h. = 0.424, s. = -0.096

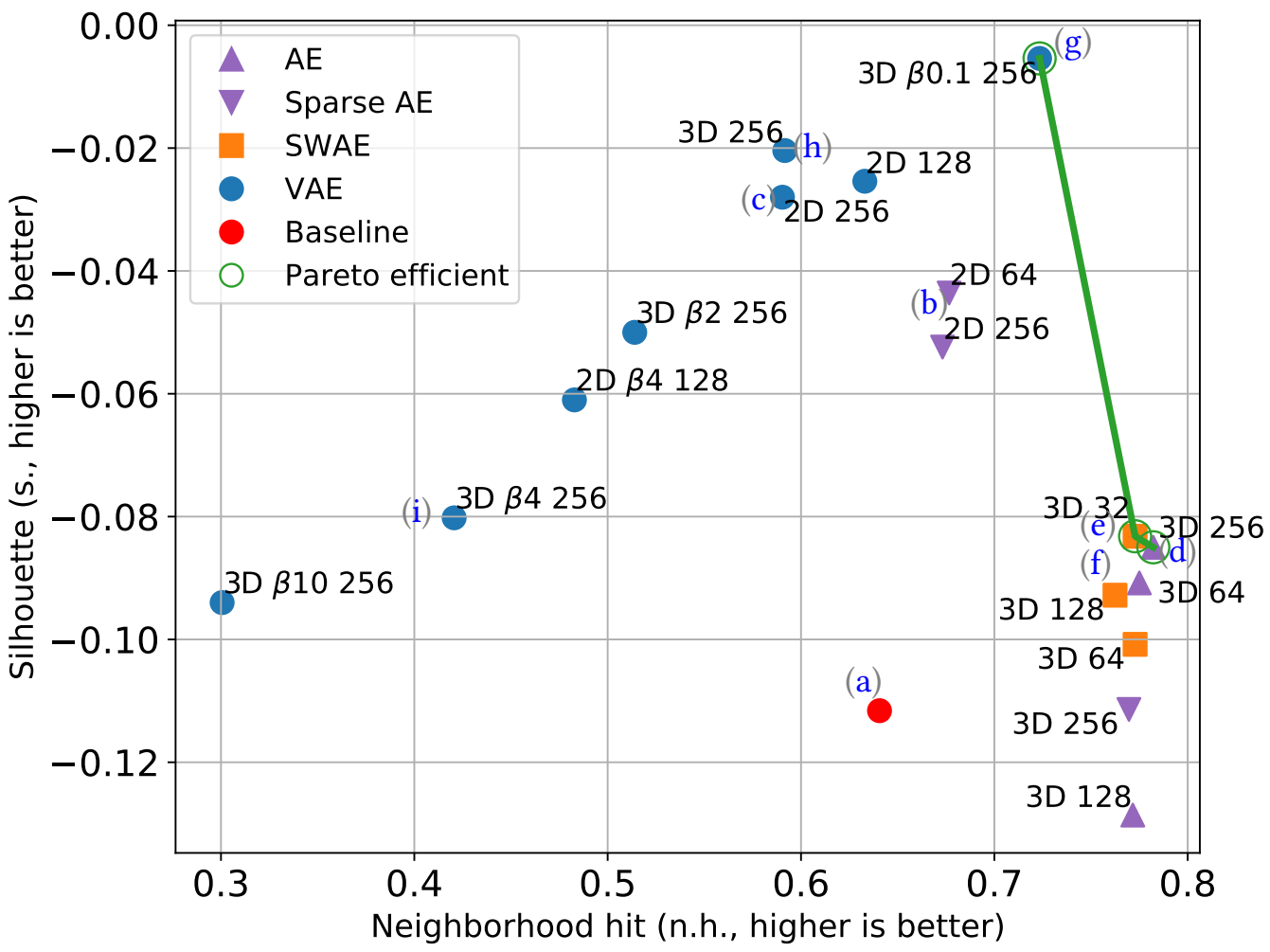

(j) Pareto frontier of autoencoder variants (projection view links in gray)

Figure 3.5: (a–i) Projection views of the Drop Dynamics ensemble (* indicates the projection belonging to a set of Pareto efficient models in (j), see discussion in Section 3.4). (j) Metric values of projection and respective Pareto frontier (marked in green, links to projections in gray).

In (e) and (f) corresponding to the SWAEs, most clusters are separated from each other, but still have multiple subclusters. This is related to WAE loss calculation: different latent codes remain far away from each other. In (g) corresponding to the $\beta(0.1)$-VAE with latent dimensionality of 256, we can notice better connected clusters in the case of "crown". In the case of "bubble", clusters are starting to appear in the form of a Gaussian distribution. This is reflected in the higher neighborhood hit and silhouette scores. In (h) corresponding to the VAE ($\beta = 1$), likewise decent projections can also be observed. Due to the KL divergence term in the VAE loss calculation, the 3D VAE model created clusters in the form of Gaussian distributions. It can be noticed e.g. in the case of "bubble" type samples, no elongated clusters can be observed. In models with higher $\beta$ values a greater effect of KL loss can be observed. The influence of $\beta$ trades off neighborhood hit for silhouette metric, helping to bring points of the same class closer, but also mixing some clusters. Such trade-offs are why we use Pareto optimality for our model selection ( Figure 3.5j). However, too high values of $\beta$ (e.g. $\beta >= 4$) mix the data points in the final visualization because the input has only a minor impact on the latent vector, which leads to poorly learned features ( Figure 3.5i).

Overall, we see that the autoencoder-based feature extraction can improve the baseline results regarding the metrics, which capture important characteristics under the presence of a large chunk of unlabeled (previously unseen) data in particular. We also note that most 3D convolutional models outperform models with 2D convolutions. 3D models were able to learn better features by using three time steps in the input instead of one, without the need to increase the dimension of the latent space.

The performance of all models used for feature extraction on the Drop Dynamics and MCMC ensembles is summarized in Tables 3.2 and 3.3, respectively. The evaluation considers three metrics: *Separability*, *Neighborhood hit*, and *Spread*. For the Pareto-frontier analysis in Figures 3.1 and 3.5, only *Neighborhood hit* and *Silhouette* scores were used, as these provided the most reliable trade-off for model selection. Nevertheless, we report results for three metrics in the tables to offer a more comprehensive assessment of model performance. The best results for each metric are highlighted in bold.

### 3.4.1 *Pareto-Efficient Model Selection*

In multi-objective evaluation, a model is considered Pareto-efficient if no other model achieves better performance in one metric without sacrificing performance in another (Section 3.3.4). In our case, the two objectives are the *Neighborhood hit* and *Silhouette* scores, both of which should be as high as possible. The Pareto frontier thus represents a set of models that provide the best possible trade-offs between these two metrics, allowing users to select a solution according to their priorities without settling for an objectively inferior choice.

Table 3.2: Metrics scores of various models performing feature extraction and projection for the Drop Dynamics ensemble.

| Method type | Convolution | Latent dim. | Separability | Neighborhood hit | Spread |
|---|---|---|---|---|---|
| Sparse AE | 2D | 64 | 0.775±0.009 | **0.691±0.012** | 0.156±0.008 |
| Sparse AE | 2D | 128 | 0.747±0.012 | 0.676±0.017 | 0.172±0.024 |
| Sparse AE | 2D | 256 | 0.745±0.013 | 0.655±0.018 | 0.168±0.009 |
| VAE | 2D | 128 | 0.735±0.008 | 0.635±0.014 | 0.153±0.009 |
| VAE | 2D | 256 | 0.704±0.005 | 0.601±0.006 | 0.143±0.015 |
| $\beta$(4)-VAE | 2D | 128 | 0.579±0.018 | 0.471±0.018 | 0.156±0.008 |
| AE | 3D | 64 | 0.723±0.007 | 0.622±0.007 | 0.129±0.01 |
| AE | 3D | 128 | 0.728±0.021 | 0.631±0.02 | 0.132±0.011 |
| AE | 3D | 256 | 0.74±0.008 | 0.642±0.014 | 0.131±0.009 |
| Sparse AE | 3D | 256 | 0.72±0.009 | 0.616±0.01 | 0.127±0.011 |
| SWAE | 3D | 32 | 0.774±0.009 | 0.678±0.007 | 0.138±0.004 |
| SWAE | 3D | 64 | 0.764±0.016 | 0.668±0.016 | 0.142±0.015 |
| SWAE | 3D | 128 | **0.779±0.017** | 0.669±0.015 | 0.135±0.008 |
| SWAE | 3D | 256 | 0.761±0.013 | 0.67±0.011 | 0.139±0.008 |
| $\beta$(0.1)-VAE | 3D | 256 | 0.726±0.007 | 0.619±0.003 | 0.136±0.021 |
| VAE | 3D | 256 | 0.647±0.024 | 0.541±0.018 | 0.111±0.011 |
| $\beta$(2)-VAE | 3D | 256 | 0.57±0.01 | 0.457±0.011 | 0.127±0.024 |
| $\beta$(4)-VAE | 3D | 256 | 0.477±0.017 | 0.37±0.011 | **0.106±0.034** |
| $\beta$(10)-VAE | 3D | 256 | 0.35±0.02 | 0.266±0.011 | 0.113±0.027 |
| Baseline | – | – | 0.745±0.005 | 0.651±0.006 | 0.174±0.023 |

As shown in Figure 3.1j for the MCMC ensemble, significant improvements over the baseline (●) were achieved with ($\beta$-)VAEs and SWAE. The most efficient models in this case are $\beta$-VAE variants with $\beta$ values ranging from 2 to 8, and the SWAE with direct 2D projection. These models consistently balance cluster separability with local neighborhood preservation.

For the Drop Dynamics ensemble (Figure 3.5j), Pareto efficiency was achieved by several architectures, including 2D/3D AEs, 3D SWAEs, and 3D VAEs. The Pareto frontier models, connected by a green line, are the 3D AE, the 3D SWAE, and the 3D $\beta$(0.1)-VAE. These configurations represent the most balanced trade-offs between the two metrics for this dataset, and their positions on the frontier suggest that no single model dominates across all objectives.

### 3.4.2 *Stability of Metrics using Labeled Subsets*

Since the labeled data available for both datasets is limited, we use all available labels for our evaluation. However, to demonstrate that meaningful metric comparisons can be achieved even with very small labeled subsets, we conduct a stability analysis by testing on random subsets of varying sizes. Specifically, we sample different proportions of the available labels—ranging from 0.1% to 2.5% for MCMC and from 0.04% to 5.33% for Drop Dynamics—and evaluate the projection quality. These small percentages correspond to absolute numbers of 2–2.5K labels, which is the entirety of our manually labeled data. As shown in Figure 3.6, as the number of labeled points increases, the metric scores converge with low variance, confirming the stability and robustness of our evaluation. In contrast, very small subsets lead to high variance and occasional metric degradation. For example, when the label subset becomes extremely sparse (c), the neighborhood hit metric breaks down, producing misleadingly low values with low variance due to the neighborhood containing points from all classes. In such cases, the score trends toward the theoretical lower bound of *1/number-of-classes*. Importantly, this analysis demonstrates that even a small fraction of labeled data—representing all the labels we have access to—is sufficient for consistent and reliable evaluation, without requiring additional manual labeling efforts.

Table 3.3: Metrics scores of various models performing feature extraction and projection for the MCMC ensemble.

| Method type | Convolution | Latent dim. | Separability | Neighborhood hit | Spread |
|---|---|---|---|---|---|
| Sparse AE | 2D | 64 | 0.984±0.003 | 0.977±0.005 | 0.173±0.028 |
| Sparse AE | 2D | 128 | 0.982±0.005 | 0.966±0.01 | 0.221±0.026 |
| Sparse AE | 2D | 256 | 0.982±0.004 | 0.962±0.007 | 0.217±0.046 |
| SWAE | 2D | 2 | **0.994±0.002** | **0.995±0.001** | 0.15±0.013 |
| SWAE | 2D | 4 | 0.951±0.006 | 0.908±0.012 | 0.243±0.011 |
| SWAE | 2D | 8 | 0.978±0.005 | 0.964±0.012 | 0.194±0.011 |
| SWAE | 2D | 16 | 0.979±0.007 | 0.962±0.012 | 0.187±0.014 |
| SWAE | 2D | 32 | 0.977±0.008 | 0.959±0.008 | 0.184±0.018 |
| SWAE | 2D | 64 | 0.973±0.006 | 0.948±0.01 | 0.192±0.039 |
| SWAE | 2D | 128 | 0.982±0.006 | 0.969±0.006 | 0.196±0.016 |
| VAE | 2D | 128 | **0.994±0.002** | **0.995±0.001** | 0.164±0.025 |
| VAE | 2D | 256 | **0.994±0.002** | **0.995±0.001** | 0.169±0.015 |
| $\beta$(2)-VAE | 2D | 32 | 0.844±0.333 | 0.839±0.345 | 0.182±0.05 |
| $\beta$(2)-VAE | 2D | 64 | 0.992±0.001 | 0.994±0.001 | 0.153±0.008 |
| $\beta$(2)-VAE | 2D | 128 | 0.992±0.001 | 0.994±0.001 | 0.14±0.009 |
| $\beta$(2)-VAE | 2D | 256 | 0.992±0.001 | 0.992±0.001 | 0.136±0.009 |
| $\beta$(4)-VAE | 2D | 128 | 0.989±0.002 | 0.987±0.003 | 0.14±0.04 |
| $\beta$(4)-VAE | 2D | 256 | 0.985±0.004 | 0.979±0.007 | 0.126±0.013 |
| $\beta$(6)-VAE | 2D | 128 | 0.983±0.003 | 0.975±0.003 | 0.135±0.027 |
| $\beta$(6)-VAE | 2D | 256 | 0.962±0.007 | 0.941±0.011 | 0.141±0.012 |
| $\beta$(8)-VAE | 2D | 32 | 0.964±0.005 | 0.945±0.009 | 0.132±0.018 |
| $\beta$(8)-VAE | 2D | 64 | 0.979±0.005 | 0.967±0.006 | 0.149±0.028 |
| $\beta$(8)-VAE | 2D | 128 | 0.959±0.011 | 0.935±0.018 | **0.122±0.014** |
| $\beta$(8)-VAE | 2D | 256 | 0.936±0.024 | 0.901±0.033 | 0.138±0.026 |
| $\beta$(10)-VAE | 2D | 128 | 0.943±0.016 | 0.912±0.022 | 0.14±0.015 |
| $\beta$(10)-VAE | 2D | 256 | 0.865±0.021 | 0.807±0.024 | 0.134±0.018 |
| $\beta$(20)-VAE | 2D | 32 | 0.376±0.283 | 0.346±0.268 | 0.231±0.057 |
| $\beta$(20)-VAE | 2D | 64 | 0.883±0.012 | 0.827±0.018 | 0.136±0.006 |
| $\beta$(20)-VAE | 2D | 128 | 0.809±0.043 | 0.733±0.053 | 0.13±0.018 |
| $\beta$(20)-VAE | 2D | 256 | 0.667±0.055 | 0.571±0.06 | 0.157±0.013 |
| $\beta$(100)-VAE | 2D | 128 | 0.249±0.008 | 0.225±0.003 | 0.234±0.024 |
| $\beta$(100)-VAE | 2D | 256 | 0.255±0.007 | 0.226±0.002 | 0.228±0.02 |
| Baseline | – | – | 0.934±0.007 | 0.897±0.008 | 0.278±0.011 |

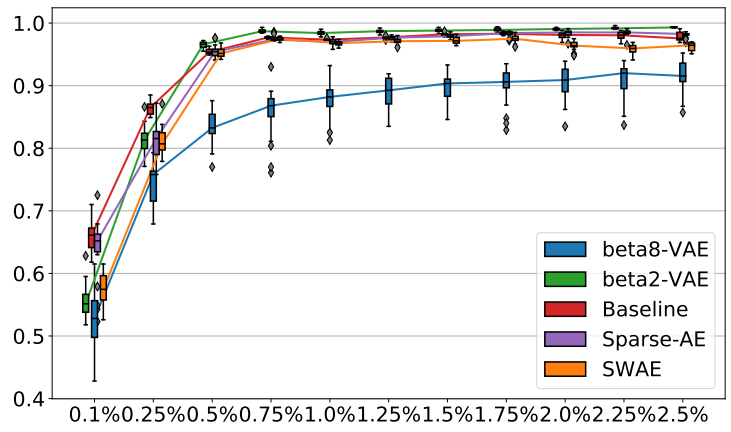

(a) Neighborhood hit, MCMC

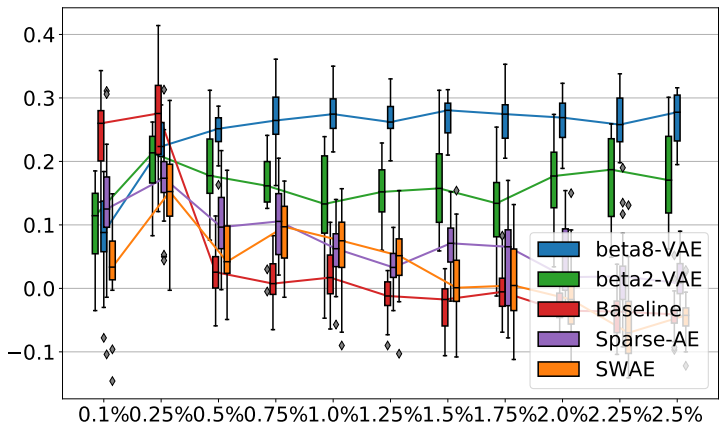

(b) Silhouette, MCMC

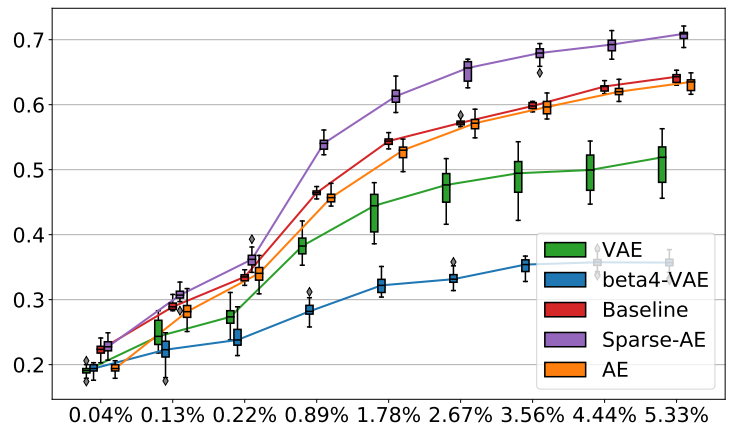

(c) Neighborhood hit, Drop Dynamics

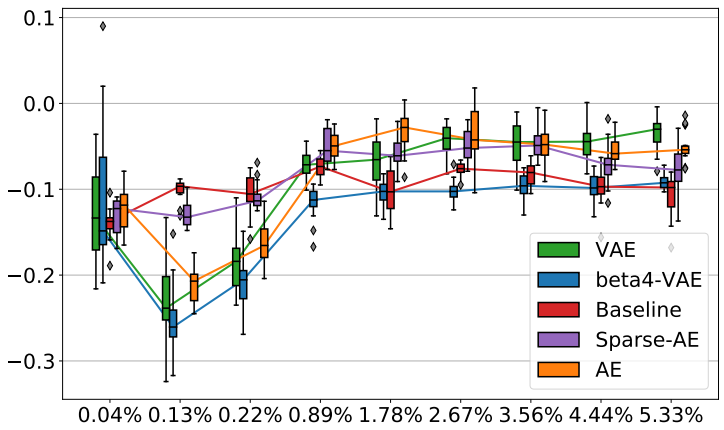

(d) Silhouette, Drop Dynamics

Figure 3.6: Stability of metrics using labeled subsets, replicated on 20 runs.

## 3.5 CONCLUSION AND FUTURE WORK

The premise of this work is that unsupervised feature learning prior to dimensionality reduction with techniques like UMAP improves the separability, structural preservation, and interpretability of low-dimensional projections for spatial and spatiotemporal ensembles. The rationale is that higher-level features extracted via autoencoders from the field data yield a representation that better captures meaningful variations and distinctive patterns within the ensemble, which are crucial for effective analysis. We demonstrate this by showing that all tested autoencoder variants yield improved results when validated using the manually assigned labels that correspond to the characteristic cases.

The properties of the autoencoders are directly reflected in the projection. Most prominently, the Kullback-Leibler divergence term of ($\beta$)-VAE causes the clusters to have a Gaussian distribution, with the influence depending on the Lagrangian multiplier $\beta$. We initially aimed to provide general suggestions on which autoencoder architecture is the most suitable. However, our experiments highlight that the optimal choice of autoencoder architecture can vary depending on the characteristics of the specific dataset. While not evaluated in this chapter, we also found that simple modifications of the datasets, such as normalization and cropping can have a significant influence. This is due to the fact that especially methods with an MSE reconstruction objective encode noisy, randomly varying, and unimportant features presented in the input. To distinguish which differences are meaningful, one requires human input, which we took in the form of labels. In this work, we propose using appropriate projection metrics combined with Pareto optimality to guide model selection. Crucially, since scientific datasets are typically unlabeled and manual annotation is time-intensive, our results show that evaluating a small, representative subset can already support informed choices. While direct labeling may not be available for novel ensembles, the analysis presented here provides a valuable initial framework for selecting suitable models in practice. According to our observations, if the images do not contain a high proportion of useful information (i.e., a large portion of the pixels relevant for feature extraction), models like AEs, SWAEs, or VAEs with small values of $\beta$ (<1) are beneficial. Otherwise, VAEs with higher values of $\beta$ (>1) can achieve better clustering results.

For MCMC with a high proportion of relevant elements in the data, autoencoders achieve a more significant improvement over the baseline in comparison to the Drop Dynamics ensemble (where the majority of pixels just represent background). To further address such cases, future research could explore alternative unsupervised learning approaches beyond autoencoders, as well as generative models employing semi-supervised setups. The robustness and general performance of the proposed pipeline could further benefit from adequate prior preparation of the data, e.g., via noise reduction or segmentation. For $\beta$-VAEs it was challenging to find the best value of the Lagrangian multiplier $\beta$, rather than to employ a fixed value, and so it might be better to gradually increase the $\beta$-weighted KL term during training in order to achieve both disentangled representation and high reconstruction quality. Finally, 2D projection is just one prominent example where prior feature learning is beneficial for visual analysis, and we aim to explore further scenarios in future work like clustering and search.

### 3.5.1 *Future Direction: Semi-Supervised Feature Learning*

A promising direction for future research involves incorporating semi-supervised learning techniques to enhance feature extraction in scientific ensemble datasets. Given the challenges posed by partially labeled data, this approach aims to leverage both labeled and unlabeled samples effectively, combining representation learning with classification objectives.

The first step in this process is obtaining partially labeled subsets from various scientific ensembles. These datasets, including Kármán vortex street (KVS), Markov chain Monte Carlo (MCMC), and Drop Dynamics, contain large amounts of data, but only a fraction is annotated. To generate pseudo-labels for the unlabeled subset, a pre-trained EfficientNet classifier (Tan and Le, 2021) can be fine-tuned on the available labeled data. This pseudo-labeling step enables the utilization of the entire dataset during training, mitigating the limitations imposed by incomplete annotations.

Following this, various autoencoder architectures can be explored for feature extraction and reconstruction, including Vanilla, Sparse, Variational, and Wasserstein autoencoders, building upon previous approaches (Tolstikhin et al., 2018). A key aspect of this study is investigating the integration of metric-based loss functions, such as neighborhood hit and silhouette scores, alongside conventional autoencoder reconstruction losses. These additional loss terms can improve the learned representations by encouraging structure preservation in the latent space.

Further refinement can involve incorporating class information directly into the loss function. This can start by reproducing results from prior work (Hinton and Salakhutdinov, 2006) and subsequently extending autoencoder training with classifier-based loss terms. Specifically, classification loss can be added to each autoencoder variant to encourage more discriminative latent representations. Additionally, incorporating supervised dissimilarity measures (Hajderanj et al., 2019) and unsupervised dissimilarity objectives (Lara and González, 2020) can further enhance the clustering quality of the learned representations. The final stage of this study can integrate all loss components, including reconstruction, regularization, classification, and dissimilarity terms, optimizing each autoencoder variant for semi-supervised feature learning.

The training process can begin with benchmark datasets such as MNIST to validate the proposed methodology before applying it to scientific ensembles. This stepwise approach helps the models generalize well and adapt effectively to more complex datasets. Once trained, the learned representations can be analyzed using projection and clustering techniques. Dimensionality reduction methods, such as t-SNE, UMAP, or PCA, can be employed to visualize the feature space in two dimensions, facilitating comparison with existing approaches.

Finally, the results can be visualized in both 2D and 3D, following established methods for scientific data exploration (Gadirov, 2020). As an additional extension, an interactive visualization tool can be developed, enabling real-time exploration of the learned representations in web browsers. This tool can support both 2D and 3D embeddings, allowing researchers to intuitively investigate feature distributions and cluster structures in high-dimensional data. By integrating semi-supervised learning with autoencoder-based feature extraction, this research aims to improve the representation quality of scientific ensemble datasets. The proposed approach has the potential to bridge the gap between supervised and unsupervised learning, leveraging the strengths of both paradigms to enhance data-driven scientific discovery.

# 4 FLINT: LEARNING-BASED FLOW ESTIMATION AND TEMPORAL INTERPOLATION FOR SCIENTIFIC ENSEMBLE VISUALIZATION

*In the previous chapter, we explored how autoencoders can serve as powerful tools for expressive dimensionality reduction in scientific ensemble visualization, capturing meaningful structural features from high-dimensional image data. Building upon this foundation, we now investigate how convolutional networks—incorporating autoencoder architectures—can be leveraged for flow estimation and temporal interpolation in 2D and 3D ensemble datasets. Our method extends the benefits of learned feature representations to spatio-temporal scientific data, enabling accurate reconstruction of missing information across both spatial and temporal dimensions. Through this approach, we demonstrate how deep learning can enhance scientific visualization tasks, providing robust solutions for flow estimation and interpolation while adapting to different ensemble parameters.*

## 4.1 INTRODUCTION

Technological advancements allow to capture and simulate time-dependent processes at high resolution, both spatially and temporally. This facilitates the acquisition of results from for a large number of runs, forming spatio-temporal ensembles. Such ensembles enable scientists to search a parameter space or estimate the impact of stochastic factors. However, the large volume of generated data often cannot entirely be preserved due to storage or bandwidth limitations (Childs et al., 2019), and, e.g., timesteps and/or variables need to be omitted. Data acquired via experiments are also typically restricted regarding available modalities (e.g., only camera images or scans are available). Reconstructing missing information post-hoc can greatly support visual analysis in such scenarios. Even if all variables of interest are captured and the full data can be preserved, assessing how well data can be reconstructed has been demonstrated to be useful for diverse purposes in visualization, including the choice of techniques in flow visualization (Jänicke et al., 2011), timestep selection (Frey and Ertl, 2017), and ensemble member comparison (Tkachev et al., 2021).

In this chapter, we propose FLINT (learning-based FLow estimation and temporal INTerpolation), a new deep learning-based approach to

This chapter is based on the paper "FLINT: Learning-based Flow Estimation and Temporal Interpolation for Scientific Ensemble Visualization" (Gadirov et al., 2025a) supplement scientific ensembles with an otherwise missing flow field, in scenarios when it had to be omitted or could not be captured, as is often the case for experiments. FLINT not only estimates a corresponding flow field for each timestep but can further reconstruct high-quality temporal interpolants between scalar fields.

Technically speaking, we consider two types of flow: *physical flow* and *optical flow*. In the general case of spatiotemporal *nD* scalar fields (like density of a fluid flow or luminance in 3D light microscopy image sequences) representing an underlying physical phenomenon, we define optical flow as the observed change of the *nD* scalar field patterns. By contrast, physical flow directly denotes the velocity of the actual material objects (moving particles in a fluid, motion of cells in a tissue volume, etc.). Physical flow is usually directly given as output of flow solvers, but can also be determined experimentally where feasible (e.g., via particle displacement velocimetry).

Optical flow can be estimated from arbitrary scalar fields, also ones that are only indirectly affected by physical flow (like a temperature field). Of course, like in the case of optical flow in computer vision we need to assume that 1) there exist observable temporal changes of the field under investigation; and 2) these changes are correlated with physical changes (in order to be meaningful). If no physical flow information is given at all—e.g., in the case of an experiment without dedicated flow measurements or when such information could not be stored due to space constraints—FLINT can estimate high-quality optical flow from scientific data as an additional modality for analysis (e.g., using flow glyphs or integral lines allows to effectively present temporal changes). If physical flow is at least partially given for some members of the ensemble (at the time of model training), FLINT is capable of making use of this and adds estimated physical flow to members during inference.

FLINT is versatile and can operate with any scalar field data, whether 2D or 3D, provided it is defined on a regular grid. In cases where only scalar data is available, FLINT can estimate optical flow. When vector data are available for some members, it can estimate the missing physical flow. Even if the scalar field is correlated but not directly governed by the vector field, FLINT can learn the vector field, although the optical and physical flow may differ in this case. Figure 4.1 shows an overview of the FLINT pipeline.

FLINT draws inspiration from recent state-of-the-art advancements in computer vision research (Teed and Deng, 2020; Dosovitskiy et al., 2015; Luo et al., 2021; Huang et al., 2022). FLINT especially builds upon RIFE (Real-time Intermediate Flow Estimation) (Huang et al., 2022)—a method for video frame interpolation based on optical flow—and extends it in several ways to enable the accurate learning of flow information, and yield high-quality temporal interpolants for scientific data.

FLINT substantially improves over RIFE for better temporal interpolation with scientific ensembles while being significantly faster, and en-ables the estimation of meaningful flow fields from spatiotemporal data, which would otherwise be missing, to open up new opportunities for analysis. Its flexible design allows FLINT to be applied to different scenarios without the need for architectural modifications—regardless of whether the flow field is (partially) available or not—simply by adapting the loss functions used in training the neural network. This makes FLINT suitable for both simulation and experimental ensembles, where the latter may typically lack available flow field data. We refer to scenarios with and without ground-truth flow as "flow-supervised" and "flow-unsupervised", respectively. FLINT performs flow field estimation and temporal interpolation in an integrated way.

A first application of flow estimation by FLINT is its facilitation of applying flow visualization techniques to datasets initially consisting solely of scalar data, which often occurs in practice for experimental data (e.g., (Geppert et al., 2016; Frey et al., 2022)). This broadens the scope of visualization possibilities and supports the presentation of dynamic processes. Another application arises in large-scale data analysis. For example, simulations on supercomputers produce massive amounts of data, only a small fraction of which can practically be stored (Childs et al., 2020). This often necessitates omitting (i) certain variables (like the flow field) and/or (ii) timesteps, either uniformly or via adaptive selection (Frey and Ertl, 2017; Yamaoka et al., 2019). With FLINT we can reconstruct data reduced both by (i) and (ii), thus enabling comprehensive analysis and understanding of complex phenomena.

In summary, we propose FLINT, a new flexible method for the estimation of flow for 2D+time and 3D+time scientific ensembles. The FLINT code is available at: `https://github.com/HamidGadirov/FLINT`. We consider our main contributions to be as follows:

- To the best of our knowledge, FLINT is the first approach to achieve high-quality flow estimation in scenarios (1) where flow information is partially available (*flow-supervised*, e.g., simulations where flow is omitted due to storage constraints) or (2) entirely unavailable (*flow-unsupervised*, e.g., experimental image sequences captured via cameras).
- FLINT produces high-quality temporal interpolants between scalar fields, such as density or luminance.
- FLINT can handle both 2D+time and 3D+time ensembles without requiring domain-specific assumptions, complex pre-training, or intermediate fine-tuning on simplified datasets.

## 4.2 RELATED WORK

**Flow field estimation in scientific visualization.** Kappe *et al.* (Kappe et al., 2015) reconstructed and visualized 3D local flow from 3D+time microscopy data using both image processing and optical flow meth-ods. Kumpf *et al.* (Kumpf et al., 2018) utilized ensemble sensitivity analysis (ESA) to obtain the evolution of sensitivity features in 2D or 3D geo-spatial data. They used forward and backward optical flow-based feature assignment for tracing the evolution of sensitivities through time. Manandhar *et al.* (Manandhar et al., 2018) developed a 3D optical flow estimation method for light microscopy image volumes. A dense 3D flow field was obtained from the 3D displacement vectors in a pair of such volumes. Gu *et al.* introduced Scalar2Vec (Gu et al., 2022), a deep learning-based framework that translates scalar fields into velocity vector fields using a k-complete bipartite translation network (kCBT-Net). We conceptually compare our findings to Scalar2Vec in Section 4.6. However, these approaches involve case-specific assumptions (Kappe et al., 2015; Kumpf et al., 2018; Manandhar et al., 2018) or rely on ground truth (GT) (Gu et al., 2022) from synthetic data, limiting their generalizability to scientific ensembles derived from physical experiments. In contrast, FLINT Does not require either data-specific assumptions or an available GT flow field.

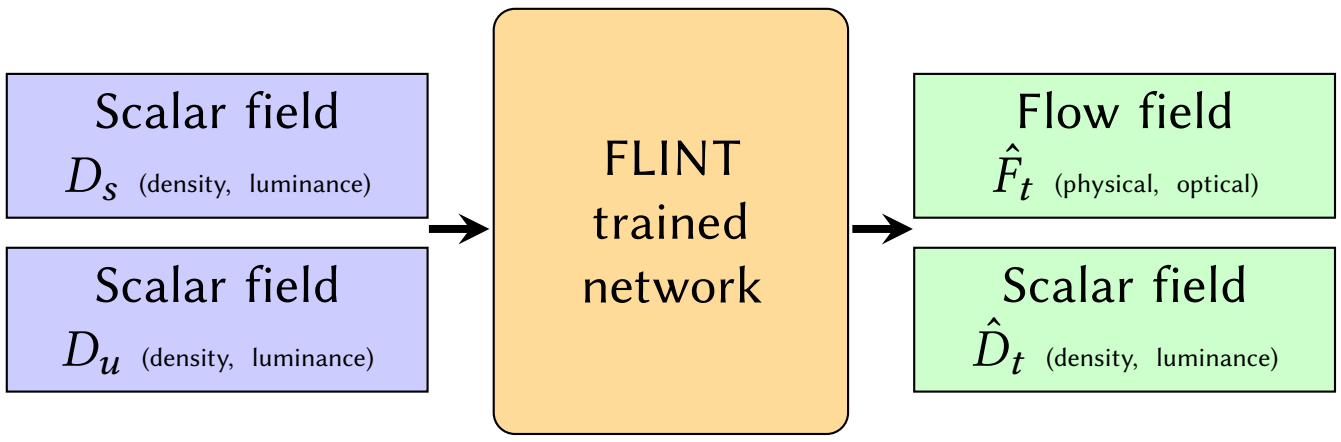

Figure 4.1: Overview of the FLINT pipeline during inference. The trained deep neural network performs flow field estimation $\hat{F}_t$ and temporal (scalar) field interpolation $\hat{D}_t$, where $s < t < u$, by utilizing the available densities $D_s$ and $D_u$ from the previous and following timesteps.

**Optical flow learning in computer vision.** In computer vision, optical flow is defined as the observed change of the pixel brightness pattern in two (or more) images of a moving object. The goal is to compute an approximation to the 2D motion field—a projection of the 3D velocities of surface points of the object onto the image plane—from spatiotemporal patterns of image intensity. Ideally, the optical flow is the same as the motion field, but this need not always be the case. Optical flow methods estimate physical velocity components by relying on constraints such as brightness constancy and imposing local or global spatial smoothness of the estimated velocity field, while aiming to preserve motion discontinuities and detect occlusions. Optical flow estimation has become a crucial component in computer vision applications, including tasks such as video frame interpolation. Recent deep learning-based methods such as RAFT and RAFT-3D (Teed and Deng, 2020, 2021), FlowNet (2.0) (Dosovitskiy et al., 2015; Ilg et al., 2017), PWC-Net (Sun et al., 2018), and ProFlow (Maurer and Bruhn, 2018) achieved remark-able results, yet require complex multi-stage optimization (involving pre-training on simpler datasets and fine-tuning on target datasets), and/or need GT flow vectors for training. Estimating optical flow in scientific ensembles, characterized by unbounded structures and density-only representations, is difficult, and manual labeling is impractical. While effective on benchmark datasets, generalization to scientific ensembles poses a substantial challenge.

FLINT is based on RIFE (Real-Time Intermediate Flow Estimation for Video Frame Interpolation) (Huang et al., 2022) which estimates optical flow using a student-teacher architecture. However, RIFE targets 2D real-world RGB datasets, such as Vimeo90K (Xue et al., 2019), Middlebury (Baker et al., 2011), UCF101 (Soomro et al., 2012), and HD (Bao et al., 2019), while FLINT can also handle 3D ensembles. For learning physical flow, we employ a loss function inspired by the RAFT loss (Teed and Deng, 2020). In scenarios where the flow field is unavailable during model training, we add additional loss components, such as photometric loss (Yu et al., 2016), to facilitate flow-unsupervised learning of the model.

**Machine learning-based upscaling & super-resolution.** ML-based upscaling and super-resolution approaches have gained significant attention across various domains, including image processing, computer vision, and scientific visualization (Ledig et al., 2017; Shi et al., 2016). They can be categorized into three main groups: spatial super-resolution (SSR), temporal super-resolution (TSR), and spatio-temporal super-resolution (STSR). SRCNN (Dong et al., 2015), SRFBN (Li et al., 2019), and SwinIR (Liang et al., 2021) are SSR techniques that aim to improve the spatial resolution of images by generating realistic and fine-grained details. TSR techniques focus on interpolating intermediate timesteps from subsampled time sequences while maintaining the spatial resolution. Methods such as phase-based interpolation (Meyer et al., 2015), SepConv (Niklaus et al., 2017), and SloMo (Jiang et al., 2018) tackle the challenge of synthesizing high-quality intermediate frames to enhance the temporal resolution of videos. While previous approaches, like STNet (Han et al., 2021), can upscale data with a fixed spatial or temporal scale factor, they do not simultaneously address both spatial and temporal super-resolution. Han *et al.* proposed a recurrent generative network, TSR-TVD (Han and Wang, 2019), designed to be trained on one variable and then applied to generate a temporally higher-resolution version of another variable from its lower-resolution counterpart. Recent methods like Filling the Void (Mishra et al., 2024), SSR-TVD (Han and Wang, 2020), FFEINR (Jiao et al., 2024), HyperINR (Wu et al., 2023), CoordNet (Han and Wang, 2022), and STSR-INR (Tang and Wang, 2024) have made significant strides in enabling TSR or SSR of data fields at arbitrary resolution. While FLINT achieves state-of-the-art accuracy in temporal interpolation between two given fields, its primary objective and notable strength lies in flow estimation. This task involves predict-ing flow fields directly from given scalar fields, assuming no flow information is available during the inference stage. Methods like CoordNet or STSR-INR, which are powerful in TSR or SSR tasks, are not designed for predicting vector fields from scalar ones. Our FLINT method not only excels in reconstructing temporal data with high precision but also produces consistent results in supplementing the data with accurate flow information.

**Student-Teacher Learning.** In machine learning, techniques have been developed under the name of teacher-student architecture. One approach is *knowledge distillation* (Hinton et al., 2015) to transfer the knowledge (parameters) learnt by a larger model (*teacher model*) and transfer it to a smaller model (*student model*). A separate strand of research uses the concept of *privileged information* (Vapnik and Izmailov, 2015), where the teacher provides the student during training with additional "privileged" information for knowledge transfer. Both approaches have been unified (Lopez-Paz et al., 2016). The learning can be *offline*, where student networks learn the knowledge from pre-trained teacher networks, or *online*, where student and teacher networks are simultaneously trained, so that the whole knowledge learning process can be end-to-end trainable (Hu et al., 2022a). We have implemented an online student-teacher model and instead of a fully separate teacher network, we employ only one additional block that corresponds to the teacher network, akin to Huang et al. (Huang et al., 2022).

## 4.3 METHOD

FLINT is, to the best of our knowledge, the first method capable of performing flow estimation from available scalar fields in scientific ensembles, while simultaneously achieving temporal super-resolution. Leveraging recent deep learning advancements in optical flow estimation, FLINT can perform interpolation between scalar fields of different timesteps without strictly relying on GT flow vectors during training. This enhances its applicability to scientific ensembles, especially where GT flow data is unavailable.

FLINT implements a student-teacher architecture (Hu et al., 2022a) which has several key advantages. It enables more accurate estimations as it is trained on GT temporal scalar fields (as demonstrated in Section 4.5). Guidance through the teacher model further enhances the student model's learning process, resulting in stable training, faster convergence, and improved model robustness. FLINT does not require pre-training and intermediate fine-tuning on simplified datasets; it already converges on the target datasets, in contrast to previous comparable works in computer vision (Teed and Deng, 2020; Dosovitskiy et al., 2015; Luo et al., 2021). FLINT further yields significantly faster convergence than existing state-of-the-art models, achieving an accuracy of more than 90% within the initial 30% of the total training time. Below, we describe the neural network architecture employed in FLINT ( Section 4.3.1), and present the temporal interpolation and flow estimation pipeline, both for the training and inference phases (Section 4.3.2). Our loss function for training can flexibly adapt to different scenarios as described in Section 4.3.3. FLINT builds upon RIFE (Huang et al., 2022) and modifies as well as extends it in several ways; a detailed comparison is presented in Section 4.3.4.

### 4.3.1 *FLINT Network Architecture*

The network architecture is a feed-forward CNN, see Figure 4.2. The student network consists of $N$ stacked convolutional blocks (*Conv Block*), each incorporating convolutional (*Conv*) and deconvolutional (*Deconv*) layers (the middle column in the orange box of Figure 4.2 shows an expanded view of a convolution block). There are 256 feature channels in all convolutional layers of the first block, 192 of the second and third, and 128 of the last block. The number of channels in the layers of the teacher block *Conv Block*$^{teach}$ is set to 128, similarly to the last block of the student model. We utilize a PReLU activation function (He et al., 2015) in all layers except for the last one. The teacher model crucially features a dedicated Conv Block$^{teach}$ which enables it to directly consider a GT scalar field by receiving $D_t^{GT}$ through an additional channel. Note that traditionally such information is not available directly to the network architecture but is only considered in the loss function. The loss components that drive the training process are shown on the right side in Figure 4.2. Student and teacher models are *jointly* optimized during training (i.e., following an online scheme (Hu et al., 2022a)), with the teacher model refining the student model's results. According to our experiments, this streamlined one-stage online training approach substantially shortens training time compared to two-stage optimization, where the teacher model is trained first and a student network is subsequently trained to align with the teacher's outputs. As demonstrated in Section 4.6.3, our one-stage approach maintains performance while offering this efficiency gain. The proposed FLINT model architecture serves as the foundation for tasks with and without available GT flow fields, differing only in the applied loss functions (Section 4.3.3).

### 4.3.2 *Flow Estimation and Scalar Field Interpolation*

As input, FLINT receives two scalar fields $D_s$ and $D_u$ of the same ensemble member at timesteps $s < u$ and an intermediate timestep $t$, where $s < t < u$. As output, FLINT (1) provides interpolants $\hat{D}_t$ at time $t \in [s, u]$ and (2) predicts the corresponding optical or physical flow field $\hat{F}_t$. First, intermediate flow fields $\hat{F}_{t\to s}$ and $\hat{F}_{t\to u}$ are computed. The *time-backward* flow $\hat{F}_{t\to s}$ refers to the intermediate flow field vec-tors from a frame at time $t$ to an earlier frame at time $s$, while the *time-forward* flow $\hat{F}_{t\to u}$ is from the frame at time $t$ to a later frame at time $u$. In the process, intermediate warped scalar fields are computed and fused by a fusion mask $M$, as follows.

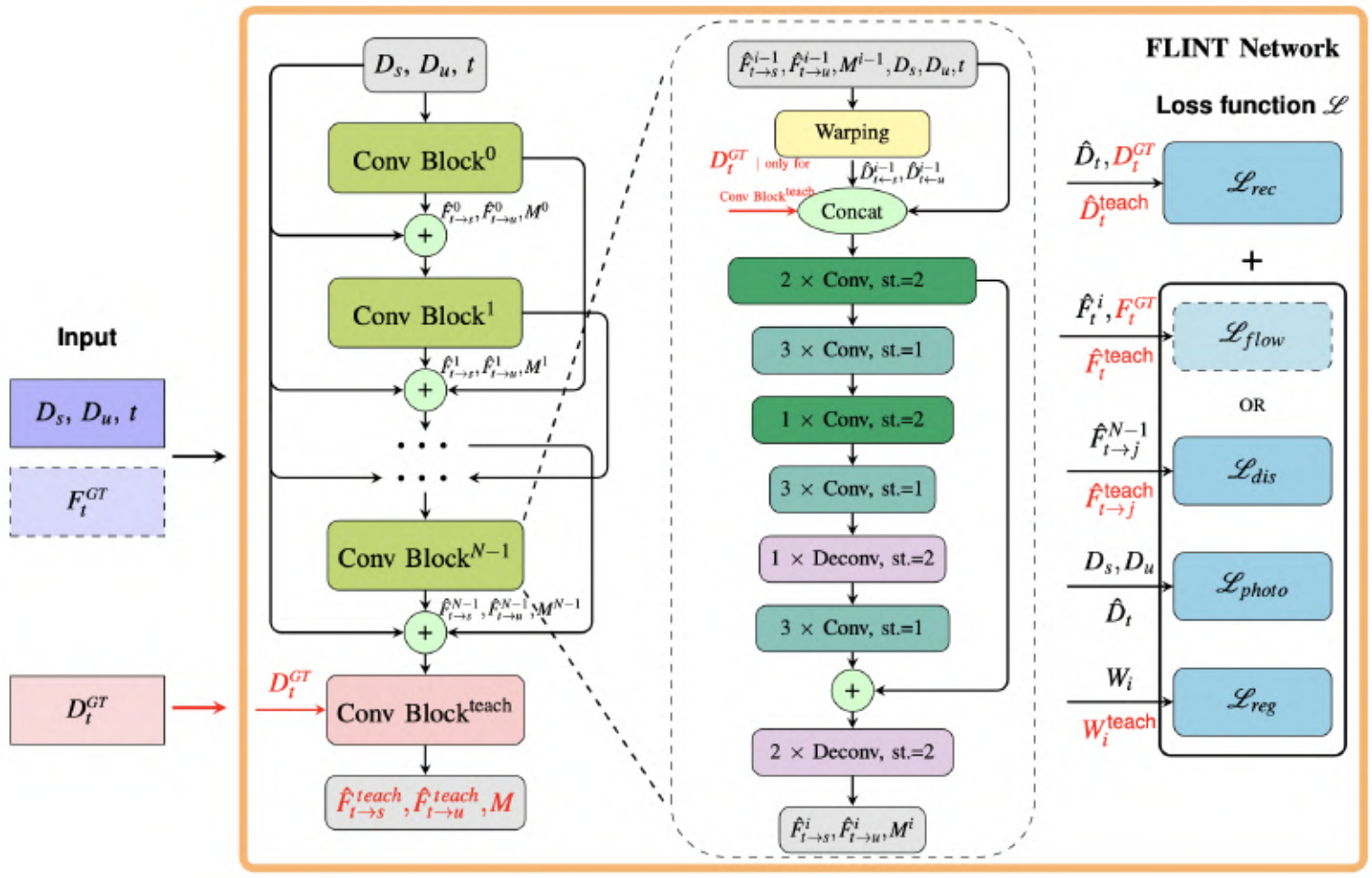

Figure 4.2: FLINT network architecture and pipeline during training (see corresponding Figure 4.1 for inference): given the input fields $D_s$, $D_u$, $D_t^{GT}$ (GT scalar field at time $t$) and $F_t^{GT}$ (GT flow at time $t$) in scenarios in which the latter is available, FLINT predicts the $\hat{D}_t$ scalar field and $\hat{F}_t^i$ flow fields used in the loss function for optimizing network parameters. The FLINT model architecture and loss function are shown in the orange box. The model consists of several stacked blocks of the convolutional network, which takes $D_s$, $D_u$, and $t$ as input and in the $i^{th}$ Conv Block computes estimated flows $\hat{F}^i_{t\to s}$, $\hat{F}^i_{t\to u}$, and fusion mask $M^i$ used for interpolation. We obtain the best results with four ($N = 4$) blocks. The teacher block Conv Block$^{teach}$, which receives $D_t^{GT}$ (the GT field at time $t$, see red arrows) as additional input, is only used during training. We employ an online student–teacher model, where instead of a fully separate teacher network, a single teacher block is integrated into the architecture. The zoomed-in view highlights the structure of a generic Conv Block consisting of backward warping, concatenation, as well as convolutional and deconvolutional layers with specified strides ($st.$). The $D_t^{GT}$ input at the concatenation stage is only used for the teacher Conv Block. The loss function, which can be adjusted depending on the scenario, is shown on the right within the orange box. $W_i$ and $W_i^{teach}$ represent the $i^{th}$ weight matrix of the last convolutional block in the student and teacher network, respectively. FLINT uses the same model architecture for ensembles with and without available GT flow fields. The GT flow $F_t^{GT}$ is only used in the loss function $\mathcal{L}_{flow}$.

**Warping**. In computer graphics warping means changing a source image into a target image. In *forward* warping a mapping is used to specify where each pixel from the source image ends up in the target image; however, holes may occur in the target image. This can be resolved by using *backward* warping. We utilize this technique in FLINT through a reverse mapping that finds, for each pixel $p_t$ in the target image, the point $p_s$ in the source image where it originated. Then resampling around this point $p_s$ is applied by bilinear interpolation of source pixel values to determine the value of the target pixel $p_t$. The warping operator $\overleftarrow{W}$ denotes the combined effect of reverse mapping and bilinear interpolation. In our case the intermediate flow fields define the mappings, see Figure 4.3. Here $\hat{D}_{t\leftarrow s}$ and $\hat{D}_{t\leftarrow u}$ are target images at time $t$ with source images $D_s$ and $D_u$, respectively. This results in two warped scalar fields $\hat{D}_{t\leftarrow s}$ and $\hat{D}_{t\leftarrow u}$:

$$\hat{D}_{t\leftarrow s} = \overleftarrow{W}(D_s, \hat{F}_{t\rightarrow s}), \quad \hat{D}_{t\leftarrow u} = \overleftarrow{W}(D_u, \hat{F}_{t\rightarrow u}). \tag{4.1}$$

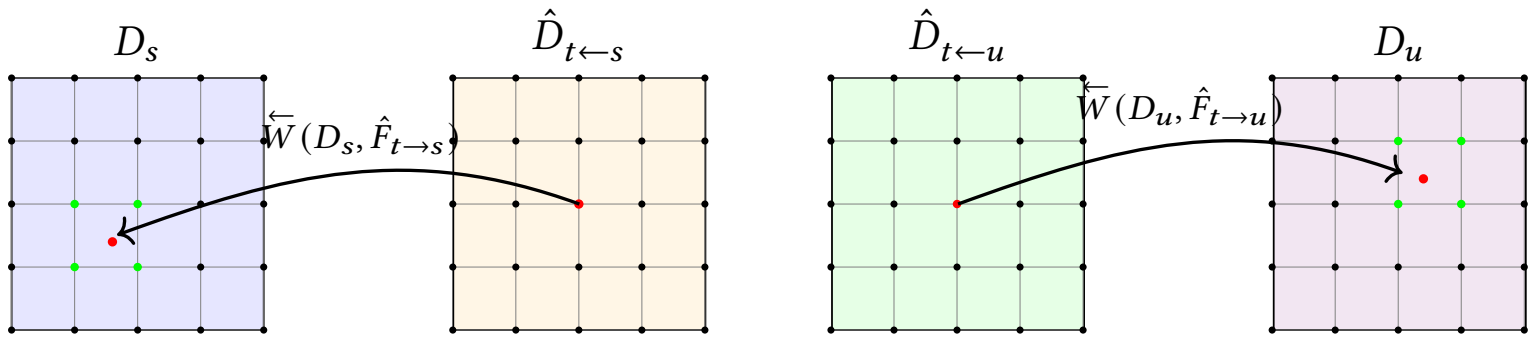

Figure 4.3: Illustration of backward warping $\overleftarrow{W}$: (scalar) fields $D_s$ and $D_u$ are reversely mapped according to the flow fields $\hat{F}_{t\rightarrow s}$ and $\hat{F}_{t\rightarrow u}$. The fields $\hat{D}_{t\leftarrow s}$ and $\hat{D}_{t\leftarrow u}$ are then reconstructed using bilinear interpolation considering the values at the coordinates shown in green.

**Fusion mask**. The fusion mask $M$, where $M(i, j) \in [0, 1], \forall i, j$, combines two intermediate warped scalar fields $\hat{D}_{t\leftarrow s}$ and $\hat{D}_{t\leftarrow u}$ at successive timesteps into an interpolated scalar field $\hat{D}_t$. The values in $M$ are learned by FLINT to ensure a smooth transition between $\hat{D}_{t\leftarrow s}$ and $\hat{D}_{t\leftarrow u}$, minimizing artifacts and preserving the structural integrity of the original fields across space.

**Refining the intermediate flows**. The intermediate flow fields are computed via $N$ convolutional blocks by iterative refinement, see Figure 4.2. This coarse-to-fine process iteratively and jointly refines both the flow fields and the fusion mask at each Conv Block, ensuring progressively consistent and high-quality updates throughout the network. Superscripts $i$ in Figure 4.2 denote the various quantities at iteration $i$. Given two input fields $D_s$ and $D_u$ and a timepoint $t$, with $s < t < u$, Conv Block$^0$ computes a rough estimation of intermediate flow fields $\hat{F}^0_{t\rightarrow s}, \hat{F}^0_{t\rightarrow u}$ and fusion mask $M^0$ to capture large motions. Then in Conv Block$^i$, $i > 0$, $D_s$ and $D_u$ are first backward-warped using Equation 4.1, based on the intermediate flows $\hat{F}^{i-1}_{t\rightarrow s}, \hat{F}^{i-1}_{t\rightarrow u}$ and mask $M^{i-1}$ of the previous iteration (see the yellow Warping block in Figure 4.2, middle column). Next, $D_s$ and $D_u$, warped frames $\hat{D}^{i-1}_{t\leftarrow s}$ and $\hat{D}^{i-1}_{t\leftarrow u}$, intermediate flows $\hat{F}^{i-1}_{t\to s}$ and $\hat{F}^{i-1}_{t\to u}$, mask $M^{i-1}$, and timestep $t$ are concatenated and processed by the stack of convolution and deconvolution layers in Conv Block$^i$. This results in updated $\hat{F}^i_{t\to s}$, $\hat{F}^i_{t\to u}$, and $M^i$, which then enter the next Conv Block. The process continues until the last Conv Block$^{N-1}$ has finished computation, producing final estimates $\hat{F}^{N-1}_{t\to s}$, $\hat{F}^{N-1}_{t\to u}$, and $M^{N-1}$.

**Interpolation and flow estimation**. Interpolated scalar field $\hat{D}_t$, intermediate flow fields $\hat{F}^i_t$, and estimated flow field $\hat{F}_t$ are obtained via:

$$\hat{D}_t = \hat{D}^{N-1}_{t\leftarrow s} \odot M + \hat{D}^{N-1}_{t\leftarrow u} \odot (\mathbf{I} - M) \tag{4.2a}$$

$$\hat{F}^i_t = \hat{F}^i_{t\to u}, \quad \hat{F}_t = \hat{F}^{N-1}_t \quad (N = 4) \tag{4.2b}$$

where $\odot$ denotes element-wise multiplication, and $\mathbf{I}$ is the identity matrix. We determined $N = 4$ as the optimal value, see Section 4.6.3.

**Refinement by teacher module**. During training, a dedicated *Conv Block*$^{teach}$ implements a "privileged distillation scheme" (Lopez-Paz et al., 2016) in the form of a teacher module that has access to a GT scalar field by receiving $D_{GT}$ through an additional channel during concatenation; see Figure 4.2. This produces flow fields $\hat{F}^{teach}_{t\to s}$, $\hat{F}^{teach}_{t\to u}$ and mask $M^{teach}$. By inserting these in Equation 4.2a and Equation 4.2b, an interpolated field $\hat{D}^{teach}_t$ and estimated flow field $\hat{F}^{teach}_t$ are obtained for the teacher. The outputs for student $(\hat{D}_t, \hat{F}_t)$ and teacher $(\hat{D}^{teach}_t, \hat{F}^{teach}_t)$ are used in the loss functions to compute the prediction error which will then be backpropagated to update the network parameters. The process is repeated for many inputs during training (for more details, see Section 4.4.1).

**Inference**. In the inference phase, the scalar fields $D_s$ and $D_u$ are processed by the *trained* network, meaning that all GT information, the teacher block, and the loss functions are absent. The interpolated scalar field $\hat{D}_t$ and reconstructed flow field $\hat{F}_t$ as computed by Equations 4.2a and 4.2b constitute the final output, see Figure 4.1.

### 4.3.3 *Loss Function*

FLINT uses several loss components that can be combined depending on the scenario: (1) flow-supervised—with available GT flow field and (2) flow-unsupervised—without (the components are listed in Figure 4.2 and described further below).

**(1)** The total loss for ensembles with available GT flow field is a linear combination of reconstruction loss $\mathcal{L}_{rec}$ and flow loss $\mathcal{L}_{flow}$:

$$\mathcal{L} = \mathcal{L}_{rec} + \lambda_{flow}\,\mathcal{L}_{flow}, \tag{4.3}$$

where $\lambda_{flow} = 0.2$ for balancing total loss scale w.r.t. the reconstruction component (determined via hyperparameter search, see Section 4.6.3).

**(2)** The total loss of FLINT for ensembles without available flow fields is a linear combination of the reconstruction $\mathcal{L}_{rec}$, distillation $\mathcal{L}_{dis}$, photometric $\mathcal{L}_{photo}$, and regularization $\mathcal{L}_{reg}$ losses:

$$\mathcal{L} = \mathcal{L}_{rec} + \lambda_{dis}\,\mathcal{L}_{dis} + \lambda_{photo}\,\mathcal{L}_{photo} + \lambda_{reg}\,\mathcal{L}_{reg}, \tag{4.4}$$

where $\lambda_{dis} = 10^{-4}, \lambda_{photo} = 10^{-6}, \lambda_{reg} = 10^{-8}$ for balancing total loss scale w.r.t. other loss components (see Section 4.6.3).

**Scalar field interpolation (flow-unsupervised).** To temporally interpolate between fields, we utilize student and teacher blocks (Section 4.3.1), incorporating loss components aimed at improving the accuracy of the interpolated density from Equation 4.2a. The reconstruction loss $\mathcal{L}_{rec}$ measures the $L_1$ distance between the GT $D_t^{GT}$ and the reconstructed field representation from both the student and teacher:

$$\mathcal{L}_{rec} = \|D_t^{GT} - \hat{D}_t\|_1 + \|D_t^{GT} - \hat{D}_t^{teach}\|_1. \tag{4.5}$$

The first term in Equation 4.5 is evaluated using $D_t^{GT}$ but only for the loss calculation; $D_t^{GT}$ is never used as input to the student network, only to the teacher module. We opt for $L_1$ instead of the more common $L_2$ distance, as it is less sensitive to outliers and better preserves sharp gradients and fine structures in the scalar field.

**Physical flow estimation (flow-supervised).** When we have access to GT flow information during training, we incorporate a flow loss component to enhance the quality of the learned flow field (physical flow). This loss function comprises reconstruction and teacher components, as well as the flow loss component. The supervised flow loss measures the $L_1$ distance between the estimated flow from each block of the neural network and the GT flow. We choose $L_1$ over the more common $L_2$ distance as it is less sensitive to outliers and better preserves sharp flow boundaries, which are important in physical simulations. In our experiments, we found that accumulating this measure based on all blocks rather than just the last one yields better results. We also adopt the concept of exponentially increasing weights from RAFT (Teed and Deng, 2020). This loss equation for physical flow estimation can be expressed as follows:

$$\mathcal{L}_{\text{flow}} = \sum_{i=1}^{N} \gamma^{N-i} \|F_t^{GT} - \hat{F}_t^i\|_1 + \|F_t^{GT} - \hat{F}_t^{teach}\|_1, \tag{4.6}$$

where $F_t^{GT}$ is the GT flow at time $t$, $\hat{F}_t^i$ is the flow output from the corresponding $i^{th}$ block of the student network ( Equation 4.2b), $N = 4$ is the number of blocks in the model, and $\hat{F}_t^{teach}$ is the flow output from the teacher block. We experimentally established the value of $\gamma$ as 0.8, aligning with the RAFT loss and validating this choice through our hyperparameter search. Again, $F_t^{GT}$ is only used for calculation of the error; it is never used as input to the student-teacher network.

**Optical flow estimation (flow-unsupervised).** When the flow field is not available for model training, we add distillation (Huang et al., 2022), photometric (Yu et al., 2016), and regularization loss components as conceptual replacements of the supervised physical flow loss. These components help the model to reconstruct the desired timesteps based on the optical flow field, which in this case is learned in a flow-unsupervised mode.

The distillation loss is based on the fact that the model outputs more accurate flow when it receives the GT timesteps of the different field which is available (e.g., density). It is computed as $L_2$ distance between the intermediate flows of the teacher and student networks:

$$\mathcal{L}_{dis} = \sum_{j \in \{s,u\}} \|\hat{F}^{N-1}_{t \to j} - \hat{F}^{teach}_{t \to j}\|_2. \tag{4.7}$$

We choose $L_2$ here as it penalizes larger deviations more strongly, which is beneficial for aligning the student's flow estimates closely with the teacher's outputs.

We also conducted experiments incorporating the smoothness loss component, commonly used in optical flow estimation tasks (Yu et al., 2016). However, in our case, where there are generally no clearly distinguishable objects within our ensembles, the smoothness loss did not improve the results. The metric results obtained by incorporating the smoothness loss component are reported in Table 4.6 under the "FLINT *smooth*" row. The Charbonnier penalty function is used in both photometric and smoothness loss components to provide robustness against outliers (Sun et al., 2014). The photometric loss is computed as the difference between the fields $D_s$, $D_u$, and the reconstructed field $\hat{D}_t$:

$$\mathcal{L}_{photo}(\vec{v}; D_j, \hat{D}_t) = \frac{1}{2} \sum_{j \in \{s,u\}} \sum_{\vec{p} \in P} \rho(D_j(\vec{p}) - \hat{D}_t(\vec{p} + \vec{v}(\vec{p})), \tag{4.8}$$

where $\vec{v}$ are the components of the estimated optical flow field and $\rho(x) = \sqrt{x^2 + \epsilon^2}$ is the Charbonnier penalty function with $\epsilon = 10^{-9}$. The summation $\sum_{\vec{p} \in P}$ indicates that the loss is computed over all pixel coordinates $\vec{p} \in P \subset \mathbb{N}^d$ of the domain in a $d$−dimensional space, and the expression $\vec{p} + \vec{v}(\vec{p})$ represents the updated pixel coordinates after applying the estimated flow vectors $\vec{v}$. We apply the photometric loss component to the last block of FLINT, i.e., $ConvBlock^{N-1}$ in Figure 4.2.

Additionally, we apply $L_1$ regularization (Bishop, 2006) to the weight matrix of the last convolutional block of the student and teacher networks to prevent overfitting of the flow field learning:

$$\mathcal{L}_{reg} = \sum_{i=1}^{L} \left( \|W_i\|_1 + \|W_i^{teach}\|_1 \right), \tag{4.9}$$

where $W_i$ and $W_i^{teach}$ represent the $i^{th}$ weight matrix of the last convolutional block in the student and teacher network, respectively, with $L$ denoting the total number of weights.

### 4.3.4 *Comparison between FLINT and RIFE methods*

We now compare the architectures of FLINT and RIFE in detail (their respective performances are discussed in Section 4.5 and Section 4.6).

**Similarities.** RIFE and FLINT both receive two scalar input fields and employ several convolution blocks to produce approximate intermediate flows and a fusion mask. Warped fields are computed from the input frames with the help of the intermediate flows. Merging the two warped frames with the help of the fusion mask via Equation 4.2a yields the interpolated field. RIFE and FLINT feature an online end-to-end trainable student-teacher architecture using a special teacher convolution block. To calculate the model residual error, both use an optimization that minimizes a loss function, consisting of various terms.

**Improvements and Extensions.** In RIFE the number of student convolution blocks is fixed to three, while in FLINT we can flexibly adapt the number and configuration of blocks for the most optimal performance. Furthermore, RIFE's use of different scales for different blocks with bilinear resize is replaced by convolution and deconvolutional layers, resulting in more learnable parameters. Additionally, FLINT comprises convolutional and deconvolutional layers with varying strides in each block, unlike RIFE's eight convolutional layers with a stride of one per block. This autoencoder-like design effectively captures key input features (Bengio et al., 2013), as demonstrated by the performance gains shown in Table 4.6.

Crucially, FLINT features new kinds of loss functions for handling different scenarios. When no flow fields are available (case **(2)** in Section 4.3.3), FLINT uses additional loss functions, such as photometric loss (Yu et al., 2016) (RIFE has only the first two terms in Equation 4.4). In contrast to RIFE, FLINT can utilize available GT flow fields during training (case **(1)**) to improve estimated flow fields quality during inference.

In RIFE, the intermediate flow fields are only used to obtain the interpolated video frames which appear blurred and discontinuous (see Huang et al. 2022, Fig. 5), especially when the teacher module is omitted. In FLINT we achieve flow fields of high quality so they can serve as meaningful supplements for spatiotemporal data analysis. FLINT further exhibits significantly lower computational cost: eliminating the need for RIFE's encoder-decoder CNN (RefineNet (Jiang et al., 2018; Niklaus and Liu, 2020)) allowed FLINT to halve its training time without impacting result quality according to our experiments. Last but not least, while RIFE is limited to 2D image sequences, FLINT can handle arbitrary 2D scalar fields as well as 3D+time data to open up new opportunities for analysis in scientific visualization. To enable this, we implemented a new 3D warping technique, integrated 3D convolutional and deconvolutional layers, and utilized 3D loss functions.

## 4.4 STUDY SETUP

In this section, we describe the training setup, provide an overview of the datasets used in our experiments and discuss the evaluation methods employed to assess the FLINT results.

### 4.4.1 *Training*

Prior to training, scalar fields are normalized to the [0, 1] range, while vector fields are normalized component-wise to the [-1, 1] range, when required. FLINT is optimized using AdamW (Loshchilov et al., 2017) — an adaptive gradient descent method with weight decay used in backpropagation algorithms for training feed-forward neural networks that combines the benefits from both the Adam optimizer (Kingma and Ba, 2014) and $L_2$-regularization. We employ early stopping with a patience parameter of 30, which is equivalent to regularization (Goodfellow et al., 2016) and helps to prevent overfitting on the training data. We use an experimentally determined learning rate of $6 \times 10^{-4}$ for the 2D case and $1 \times 10^{-4}$ for the 3D case with a cosine annealing scheduler that gradually decreases the learning rate to $6 \times 10^{-6}$ and $1 \times 10^{-6}$, respectively, by the end of the training. We train FLINT with mini-batches of size 32 for both 2D ensemble datasets and mini-batches of size 2 for 3D datasets used in our study (see Section 4.4.2). We split the set of all available data into training, validation (for monitoring training progress), and test subsets.

Note that training and test data are obtained by subsampling the original dataset, which is assumed to provide the GT scalar fields. Then FLINT performs interpolation at missing timesteps.

To support arbitrary interpolation, we chose $t \in [s, u]$ randomly at the training stage. Throughout this work, we use a maximum time window of size 12 to sample the triplets $D_s, D_t, D_u$ for the training set, which was determined via hyperparameter search, to support arbitrary interpolation and flow estimation during the training stage. This window determines the maximum time gap between the timesteps that are used for interpolation (i.e., timesteps $s$ and $u$). Our proposed FLINT model is trained for 120 epochs, resulting in a trained model size of 79MB. The training process takes no more than 12 hours on a single Nvidia Titan V GPU with 12GB of VRAM to converge for all datasets.

### 4.4.2 *Datasets*

We consider four scientific datasets in our study.

**LBS (flow-supervised).** The first datasets is generated by a Lattice Boltzmann Simulation (Mocz, Accessed: 2023). This ensemble is similar to a classic Kármán vortex street simulation and yields density as well as flow fields with a spatial resolution of $100 \times 400$ (see examples in Figure 4.4). We consider an ensemble comprising 21 members, each consisting of 3K timesteps, with varying cylinder size, position, radius, collision timescale, and the dynamic viscosity of the fluid within the simulation. We utilize different members of the ensemble for validation and testing of FLINT. We randomly sample a training subset of 40K timesteps from a 12-sized window in the case of the LBS ensemble as discussed above.

**Droplets (flow-unsupervised).** The second datasets is a Drop Dynamics (Droplets) ensemble derived from a physical experiment investigating the impact of a falling droplet on a film (Geppert et al., 2016). The experiment employs shadowgraphy imaging to study the splash crown and secondary droplets formed after the primary droplet's impact (see examples in Figure 4.5). Shadowgraphs reflect changes in the second derivative of density, which lead to variations in the refractive index of the medium and can therefore be detected optically. The experimental dataset consists of monochrome videos with a spatial resolution of $160 \times 224$, totaling 135K timesteps from 1K ensemble members. This dataset was collected to analyze various droplet impact regimes in relation to parameters like fluid viscosity, droplet velocity, film thickness, and Weber number. We interpret changes in luminosity in the video images as variations in the density field. Since no flow information is available, this constitutes a flow-unsupervised setup. We use different sets of 3K timesteps for validation and testing of FLINT and a randomly sampled subset of 40K timesteps for training.

Examples of fields at different timesteps from various members of both the LBS and Droplets ensembles can be found in Figures 4.4 and 4.5. These figures offer a visual representation of the data from simulations and physical experiments, which are used in our research. One can observe that Figure 4.4 illustrates the considerable structural diversity that exists between the density and flow fields of the LBS ensemble, highlighting the intricate variations within these fields.

**5Jets (flow-supervised).** The third datasets is a 3D+time 5Jets dataset generated by a Navier Stoke flow solver originally obtained from UC Davis. This simulation models five jets entering a cubical region and consists of 2000 timesteps with a spatial resolution of $128 \times 128 \times 128$. It contains density and velocity. We utilize every $10^{th}$ timestep of the simulation and randomly sample a subset of 500 timesteps for training, validation, and testing.

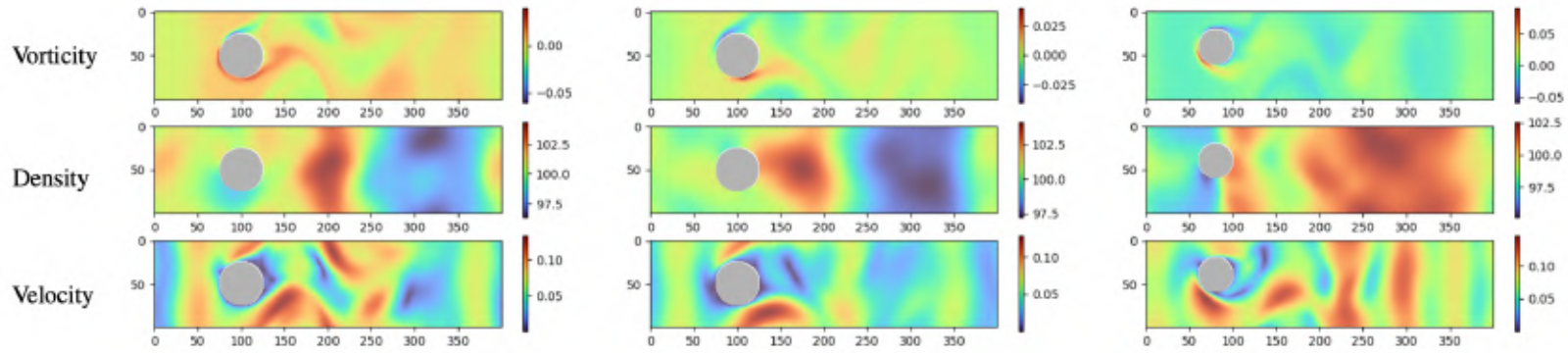

Figure 4.4: LBS samples from three different ensemble members at $t = 9000$: vorticity, density, and velocity magnitude fields.

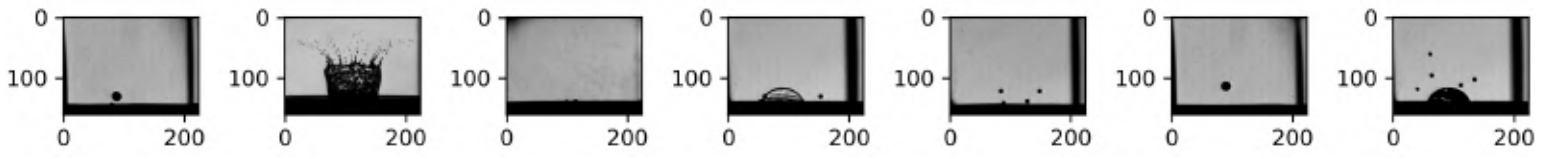

Figure 4.5: Drop Dynamics samples from different ensemble members at different timesteps.

**Nyx (flow-supervised).** The fourth dataset is a 3D+time ensemble based on the compressible cosmological hydrodynamics simulation Nyx, developed by Lawrence Berkeley National Laboratory (Sexton et al., 2021). We consider an ensemble comprising 18 members, each consisting of a maximum of 1600 timesteps with a spatial resolution of $128 \times 128 \times 128$. It contains density and velocity. Akin to InSituNet (He et al., 2019), we vary three parameters for ensemble generation: the total matter density ($\Omega_m \in [0.12, 0.155]$), the total density of baryons ($\Omega_b \in [0.0215, 0.0235]$), and the Hubble constant ($h \in [0.55, 0.7]$). We randomly sample a training subset of 500 timesteps after selecting every 5th timestep across different ensemble members for training, validation, and testing.

To visualize a simulation ensemble field, we use the *Turbo* colormap (Crameri, 2018). We visualize the 2D flow field via arrow glyphs and 3D flow via volume rendering with a transfer function that indicates flow direction. Examples of the LBS and Droplets ensemble fields, as well as 5Jets and Nyx transfer functions, are in Figure 4.19 and Figure 4.20, respectively.

### 4.4.3 *Evaluation*

We evaluate the performance of FLINT for reconstructing the scalar field both qualitatively and quantitatively. For quantitative evaluation, we utilize two different metrics: *peak signal-to-noise ratio* (PSNR) and *learned perceptual image patch similarity* (LPIPS) (Zhang et al., 2018). LPIPS measures similarity between the activations of two images using a pre-defined network; a lower score indicates greater perceptual similarity.

We evaluate the accuracy of the learned flow field using the *endpoint error* (EPE), which measures the average Euclidean distance between estimated and GT flow vectors—a lower EPE indicates higher accuracy. For qualitative assessment, we visualize the flow field outcomes and analyze difference plots for both simulation and experimental ensembles. In our evaluation, we consider different interpolation (or subsampling) rates, where a rate of $x$ means that we only consider each $x^{th}$ timestep of the original data as input.

For the 3D datasets, we utilize 3D PSNR and 3D EPE in the volume domain rather than the image domain. This ensures that the evaluation metrics are appropriately adapted to the spatial characteristics of the 3D data, providing a more accurate assessment of the performance and accuracy of FLINT in handling volumetric datasets.

The 3D PSNR metric used in our evaluation is computed after normalizing the scalar data to [0,1], using the following formulation:

$$\text{MSE} = \frac{1}{mno} \sum_{i=0}^{m-1} \sum_{j=0}^{n-1} \sum_{k=0}^{o-1} \left( D_{t,i,j,k}^{GT} - \hat{D}_{t,i,j,k} \right)^2, \tag{4.10}$$

$$\text{PSNR} = 20 \log_{10}(1.0) - 10 \log_{10}(\text{MSE}), \tag{4.11}$$

where the mean squared error (MSE) is calculated over the entire 3D volume, with $m, n, o$ representing the spatial dimensions of the volume along the $x$-, $y$-, and $z$-axes, respectively.

For flow estimation, we compute the 3D endpoint error (EPE) as:

$$\text{EPE} = \frac{1}{mno} \sum_{i=0}^{m-1} \sum_{j=0}^{n-1} \sum_{k=0}^{o-1} \left\| F_{t,i,j,k}^{GT} - \hat{F}_{t,i,j,k} \right\|_2. \tag{4.12}$$

Here, $F^{GT}$ and $\hat{F}$ represent the ground-truth and predicted flow fields, respectively, with components along the $x$-, $y$-, and $z$-axes. The EPE measures the Euclidean distance between the predicted and true flow vectors across the entire 3D volume.

## 4.5 QUALITATIVE RESULTS

We evaluate FLINT via four different datasets (Section 4.4.2) in different scenarios regarding flow estimation (i.e., supplementing density with flow in both flow-supervised and flow-unsupervised cases) as well as density interpolation (i.e., temporal super-resolution).

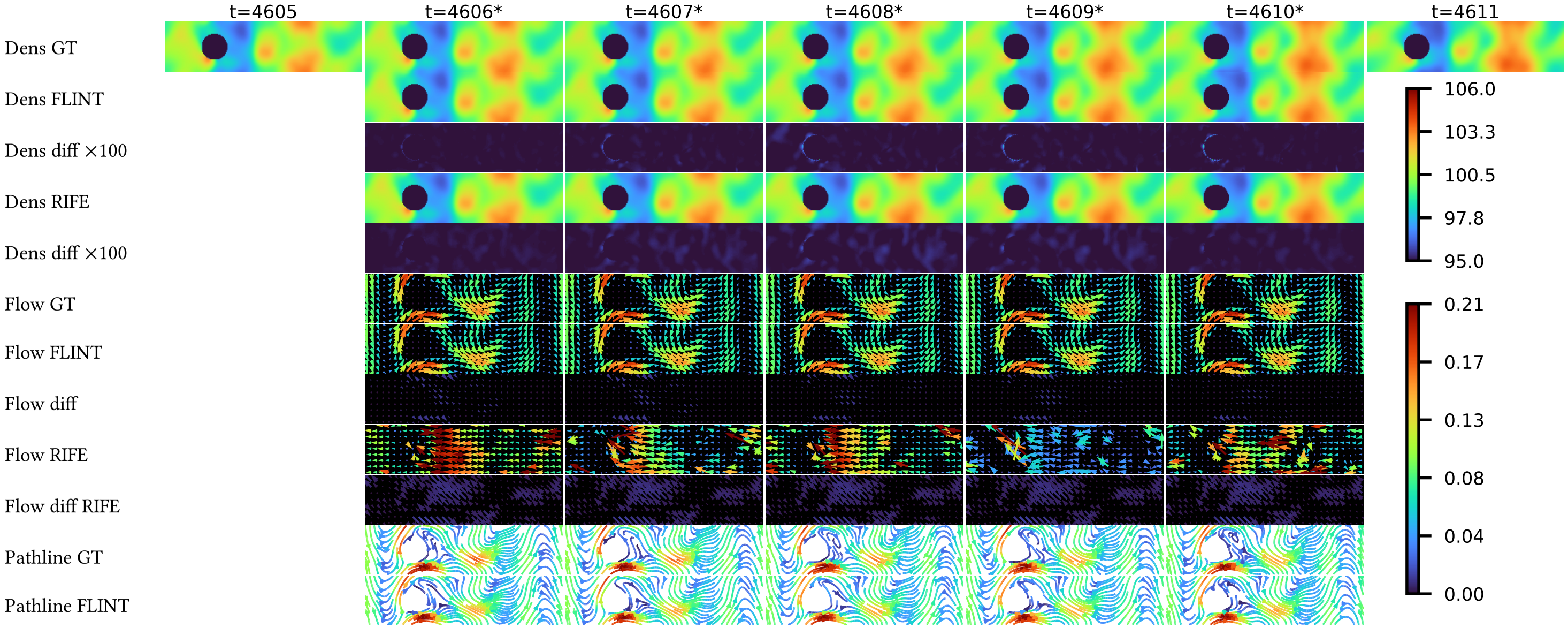

Figure 4.6: LBS ensemble: FLINT flow field estimation and temporal density interpolation at the timesteps with an asterisk (*)— 6× interpolation (between $s = 4605$ and $u = 4611$). From top to bottom: GT density, FLINT interpolated density, difference with GT density (magnified by ×100), RIFE interpolated density, difference with GT density (magnified by ×100); GT flow, FLINT flow estimation, difference with GT flow, flow estimated by RIFE, difference with GT flow, GT pathlines, and FLINT pathlines. The colorbar on the top right maps density, and the one on the bottom right maps flow magnitude.

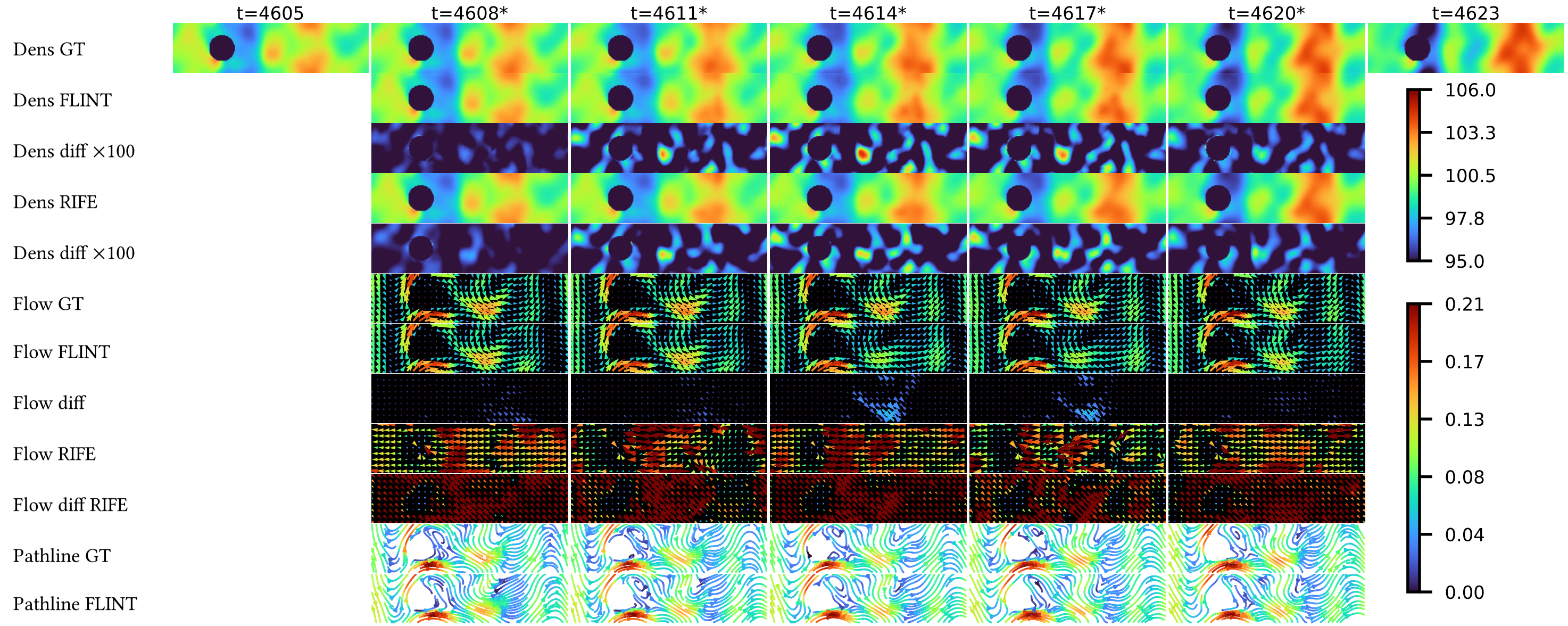

Figure 4.7: LBS ensemble: FLINT flow field estimation and temporal density interpolation at the timesteps with an asterisk (*)— 18× interpolation (between $s = 4605$ and $u = 4623$). From top to bottom: GT density, FLINT interpolated density, difference with GT density (magnified by ×100), RIFE interpolated density, difference with GT density (magnified by ×100); GT flow, FLINT flow estimation, difference with GT flow, flow estimated by RIFE, difference with GT flow, GT pathlines, and FLINT pathlines. The colorbar on the top right maps density, and the one on the bottom right maps flow magnitude.

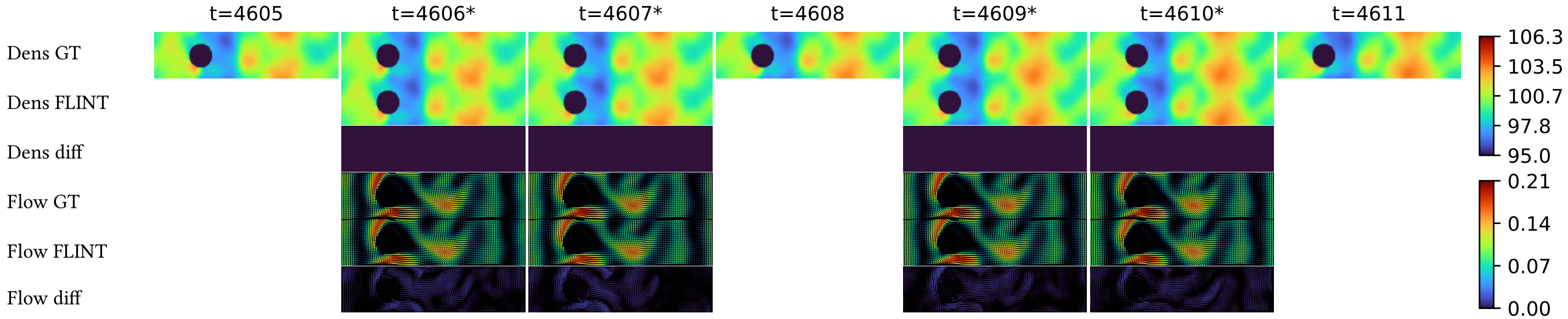

Figure 4.8: FLINT results for LBS ensemble, 3× interpolation: labels of the rows (on the left) from top to bottom: density GT, interpolation, difference; flow GT, estimation, and difference; colorbars: top — density, bottom — flow (on the right). The use of an asterisk symbol (*) following the timestep numbers indicates the density interpolation and flow estimation (second and fifth rows) carried out at that particular timestep. These can be compared against the actual GT fields (first and fourth rows).

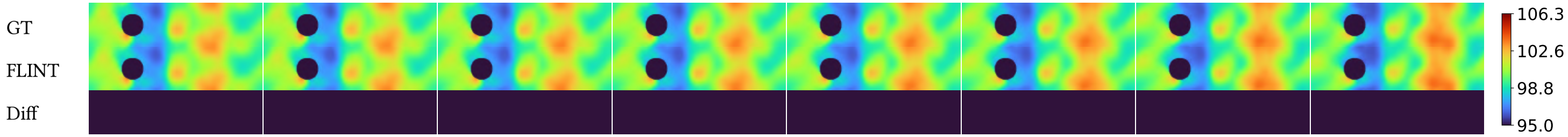

(a) Density interpolation 8×: comparing FLINT estimations against GT, from $t = 4606$.

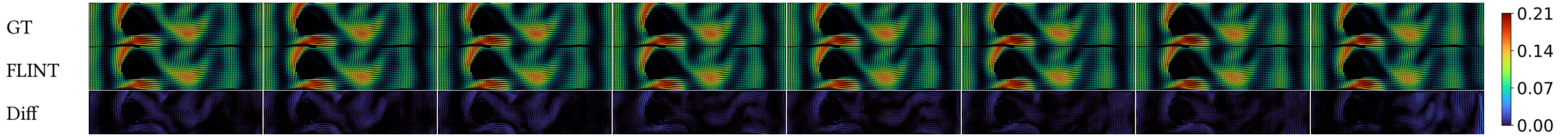

(b) Flow estimation 8×: comparing FLINT estimations against GT, from $t = 4606$.

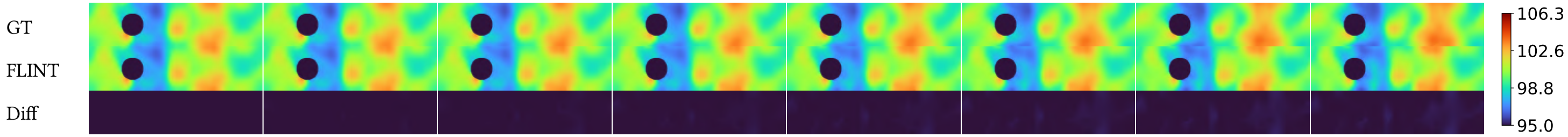

(c) Density interpolation 32×: comparing FLINT estimations against GT, from $t = 4606$.

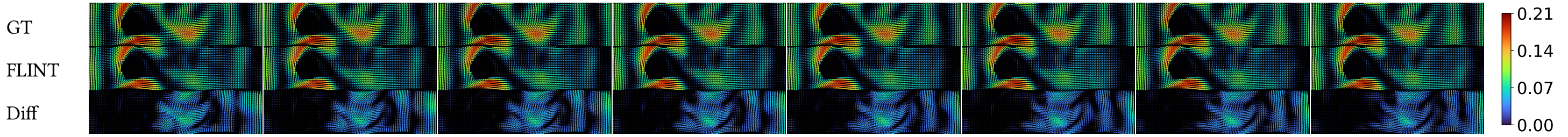

(d) Flow estimation 32×: comparing FLINT estimations against GT, from $t = 4606$.

Figure 4.9: LBS ensemble: Comparison of FLINT estimations at 8× and 32× interpolation. For both density interpolation and flow estimation, results are compared against the ground truth (GT). Each row shows: GT, FLINT estimations, and the difference.

### 4.5.1 *LBS: Flow Field Available for Some Members*

First, we focus on the scenario in which we have GT flow fields available for some ensemble members (i.e., flow-supervised learning scenario). As can be seen in Figures 4.6 and 4.7, in general, our approach achieves accurate performance in terms of density field interpolation and crucially flow field estimation. In this example, results were obtained at the interpolation rates of 6× and 18×. For both cases, the error between the reconstructed density field and its GT is small (note the magnification by a factor of 100 in the FLINT difference plots, third row). Moreover, it shows that the model is able to effectively learn a flow field that closely resembles simulated flow, resulting in a low error when compared to GT data. Comparing our FLINT method to RIFE, we can see that interpolation of the density field worked well for both, however, the flow learned by RIFE is not accurate and is far away structurally from the GT flow. (See Section 4.3.4 for a summary of the differences between the FLINT and RIFE methods.) Upon closer examination of the density difference error between FLINT and GT in Figure 4.7, it becomes apparent that the highest error occurs in the middle of the frame. This aligns with an observed error between the GT flow and FLINT flow estimation in the same region. This observation suggests that as the model begins to produce flow with a certain degree of error in this example, the density also experiences a similar error.

In general, going from an interpolation rate of 6× (Figure 4.6) to 18× (Figure 4.7), the density interpolation remains quite accurate (differences are only visible due to significant contrast enhancement). Naturally, there are more errors in the flow estimations of both FLINT and RIFE when increasing the interpolation rate, but FLINT is still able to maintain comparably high accuracy. Further examples including different interpolation rates can be found in Figures 4.8 and 4.9. This includes evaluations at very high interpolation rates, such as 32×, where the quality of flow estimation begins to degrade due to the significant structural differences between widely separated timesteps. Despite this, the scalar field interpolation remains relatively robust, demonstrating FLINT's ability to handle challenging conditions effectively.

Additionally, we perform a pathline analysis to further assess the accuracy of the flow estimated by FLINT. Pathlines were generated to visualize the flow field. Results for FLINT are shown in the last row and GT pathlines in the last but one row in Figures 4.6 and 4.7. It can be seen that the pathline patterns produced by FLINT align closely with the GT for an interpolation rate of 6× (Figure 4.6), indicating a high level of accuracy in the flow field estimation. Even at an increased interpolation rate of 18× (Figure 4.7, more detailed in Figure 4.10), the pathline patterns closely reflect the GT, although some small deviations become noticeable (e.g., for $t = 4608$, $t = 4614$, and $t = 4617$). These findings further validate the robustness of FLINT in capturing complex flow dy-namics across varying interpolation rates, demonstrating the utility of reconstructing this otherwise missing flow information for analyzing flow around a cylinder.

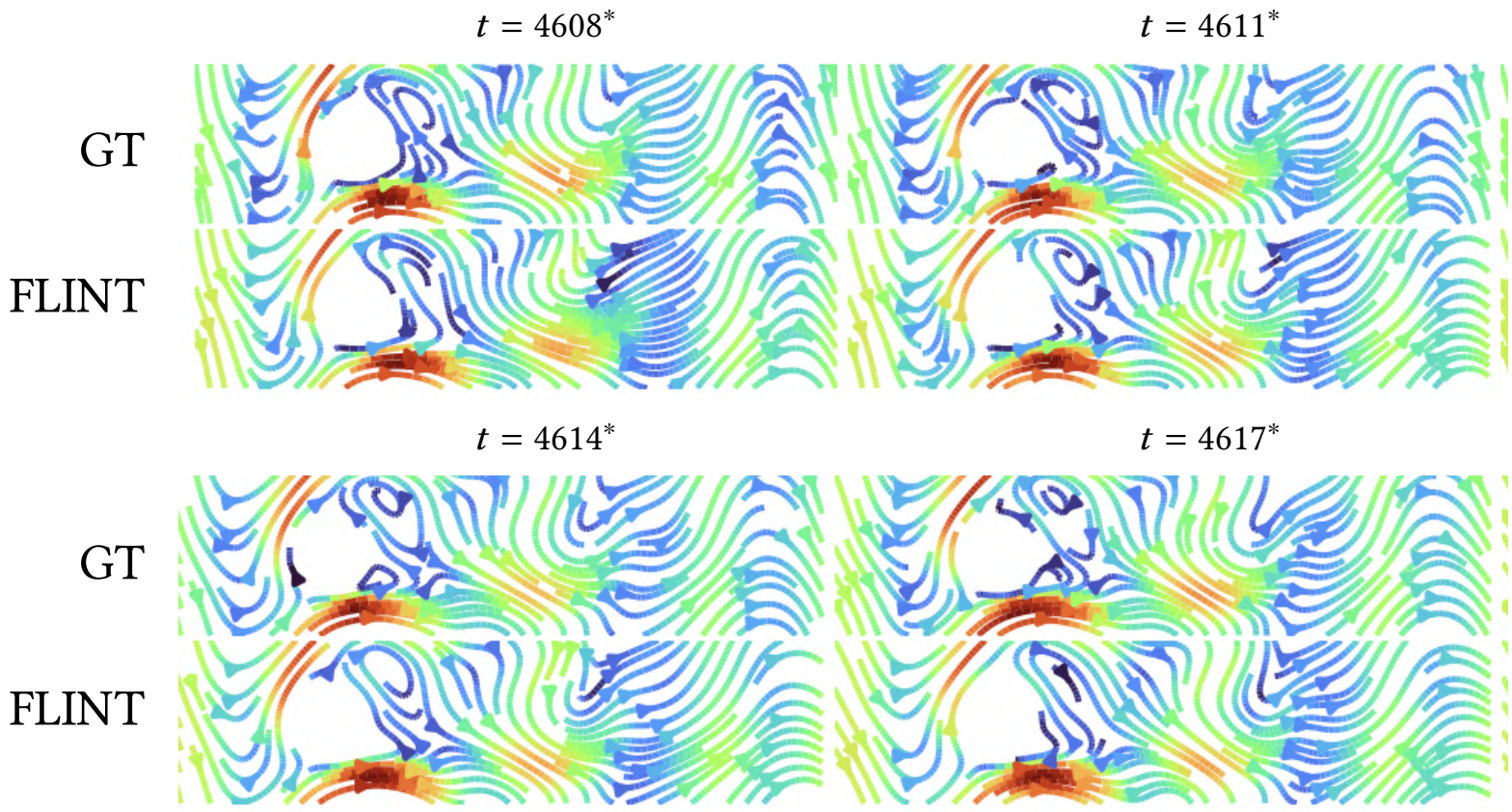

Figure 4.10: Pathline analysis, LBS, 18×.

More illustrative results showcasing FLINT's performance at various interpolation rates for the LBS ensemble are provided in Figures 4.8 and 4.9. These results affirm the positive outcomes in terms of density interpolation and flow estimation achieved at the interpolation rates of 3× and 8×, indicating the method's reliability in these scenarios. However, it is worth noting that at a very high rate of 32×, as can be seen in Figure 4.9d, although density interpolation can still maintain satisfactory quality, there appears a noticeable degradation in the quality of flow estimation, which is not surprising due to the large variation in structure between the two fields. In Figure 4.9d, we observe significantly larger errors for flow estimation compared to lower interpolation rates, highlighting the challenges posed by such high rates.

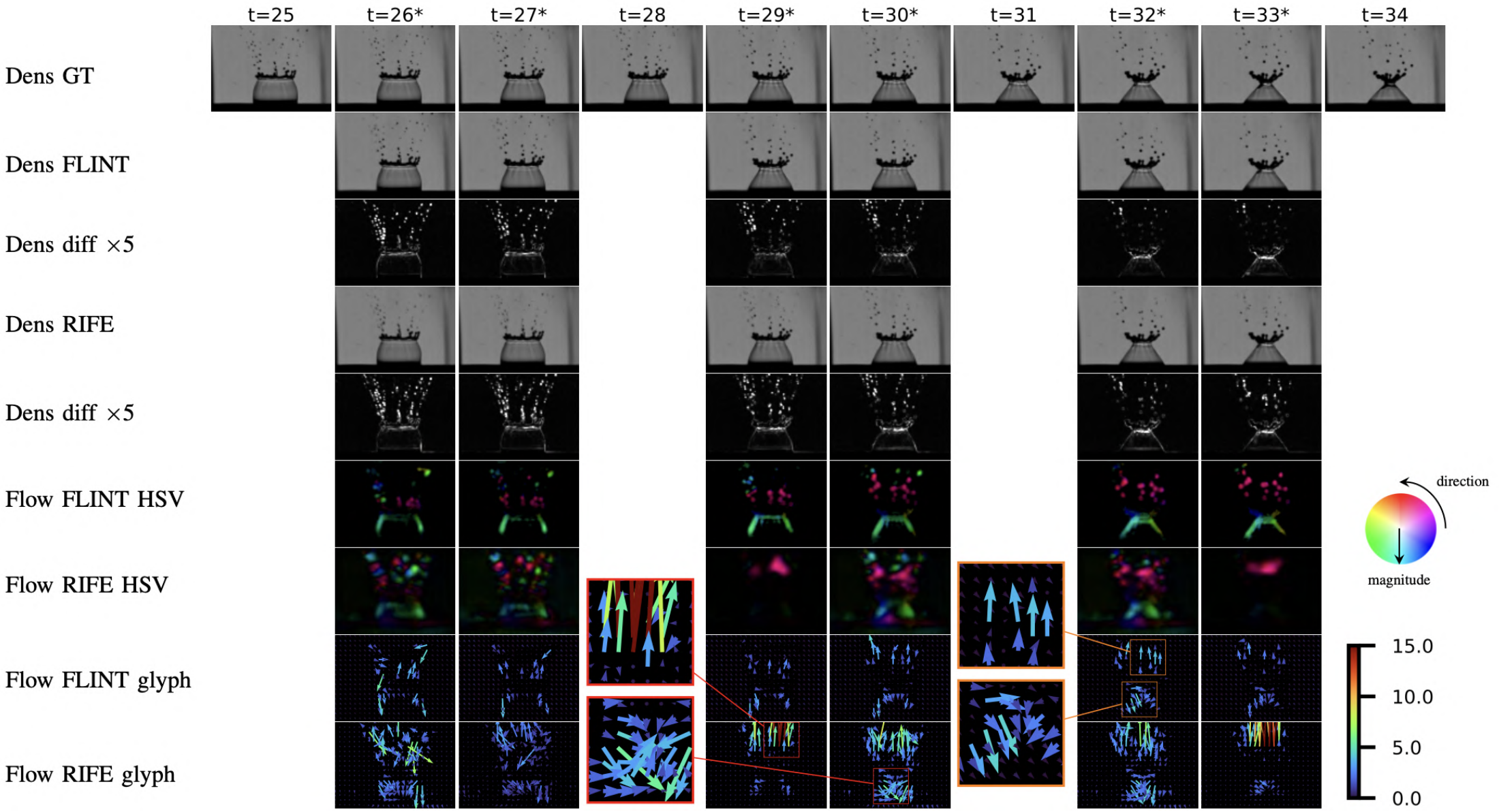

Figure 4.11: Droplets ensemble: FLINT flow field estimation and temporal density interpolation during inference—at the timesteps with an asterisk (*)—3× interpolation. From top to bottom, the rows show GT density, FLINT interpolated density, difference to GT density (magnified by ×5), RIFE interpolated density, difference to GT density (magnified by ×5), FLINT flow estimation in HSV (see bottom right), RIFE flow estimation in HSV, flow glyphs for FLINT, and flow glyphs for RIFE.

### 4.5.2 *Droplets: No GT Flow Field Available*

The effectiveness of FLINT in interpolating the luminance field—captured by cameras during an experiment—based on the estimated optical flow, even in the absence of GT flow information, is evident from Figure 4.11. The results demonstrate camera image interpolation by FLINT of high quality for the Droplets ensemble, in the second row. We compare FLINT's temporal interpolation performance to the one achieved with RIFE in Figure 4.11, fourth row. RIFE's interpolation exhibits slightly higher variation compared to the GT, as can be seen from the difference plots, in the third and fifth rows. It is noteworthy that this concerns experimental data that naturally contains noise. For instance, in the top row of Figure 4.11, the noise is visible as small, scattered inconsistencies in the background throughout the luminance field, giving it a slightly grainy appearance. Additionally, shadows can be observed on the right side of the images, where darker regions obscure portions of the luminance field. Despite this, FLINT demonstrates robust performance, effectively handling these imperfections and delivering high-quality interpolation results.

When examining the optical flow results generated by FLINT in Figure 4.11 (sixth row), it is clear that the model successfully captures meaningful flow patterns that correspond to the direction of fluid particle movements in the majority of cases. The visualization shows droplets moving upward with a distinct red hue, while the parts of the splashes collapsing downward are characterized by a green hue. The optical flow estimated by RIFE, in comparison, is significantly less accurate as it fails to capture the finer details of particle movements, resulting in a blurrier flow. This indicates that FLINT is able to learn and predict flow information that aligns with the underlying dynamics of the fluid. In Figure 4.11 (the second-to-last row for FLINT and the last row for comparison with RIFE), we further present arrow glyphs depicting flow direction and magnitude. With FLINT, these visualizations accurately reflect the movements of the bubble, crown, and droplets, and with this directly capture the evolution of the underlying physical phenomena in a static image (complementing the experimental images in the top row(s)). For example, in the orange zoom-ins ($t = 32$), the arrow glyphs effectively illustrate how the bottom part of the crown collapses downward, while the droplets predominantly move upward as they splash. This provides a clear visualization of the contrasting dynamics within the scene. When comparing FLINT to RIFE, it becomes evident that our method offers superior accuracy in capturing the flow dynamics. RIFE tends to generate excessive flow in the area of the crown, particularly in timesteps $t = 26, 27, 30$, and $32$. For example, the red bottom zoom-in ($t = 30$) demonstrates such a case where in the original experimental images only the boundaries of the bubble are visibly moving (see top row, Dens GT). Additionally, RIFE produces less accurate flow represen-tations for the droplet splashes in the top part of the scene, as clearly seen in timesteps $t = 26, 29$ (red zoom-in), and 33. In contrast, FLINT adequately captures these flow patterns, providing a clear representation for the analysis of fluid dynamics. In sum, FLINT generates flow fields that align with observed movements for analysis, even without GT flow.

### 4.5.3 *5Jets: Density & Flow Available for Some Timesteps*

Third, we consider the scenario in which we have GT density and velocity fields available for some timesteps sampled from the whole 3D dataset. In this case, the goal is to perform unsupervised density interpolation (i.e., density TSR) and flow estimation via a flow-supervised learning scenario. As Figure 4.12 shows, our approach achieves accurate performance in terms of density field interpolation and flow field estimation. In this example, results were obtained at an interpolation rate of 20×. Visually, the error between the renderings of the reconstructed density field and its GT is small. Moreover, it shows that even at a comparably large interpolation rate of 20×, the model is able to effectively learn a flow field that structurally resembles the GT flow. Comparing our FLINT method to CoordNet, we can see that interpolation of the density field worked well for both (although some differences can be observed, especially highlighted with the red circles). We are using a similar model size of both FLINT and CoordNet for this comparison. Crucially, however, FLINT additionally supplements the density with accurate flow information (as can be seen in Figure 4.12), a feature that CoordNet lacks. In this dataset, the flow reveals the evolution of five distinct jets along the $x$, $y$, and $z$ directions. Observing the flow allows us to interpret how density evolves, particularly in the purple and orange zoom-ins from timesteps $t = 710$ to $t = 790$. Here, we can see that the density concentrates and intensifies along the paths of the jets, forming more pronounced structures as the jets push the material outward. This results in higher density regions aligned with the jet flows, indicating areas where the flow is driving the accumulation of matter. This enhances our understanding of the underlying physical processes and provides a more comprehensive system analysis. Examples of the metric scores for the interpolation rates of 10×, 15×, and 20× can be found in Table 4.1.

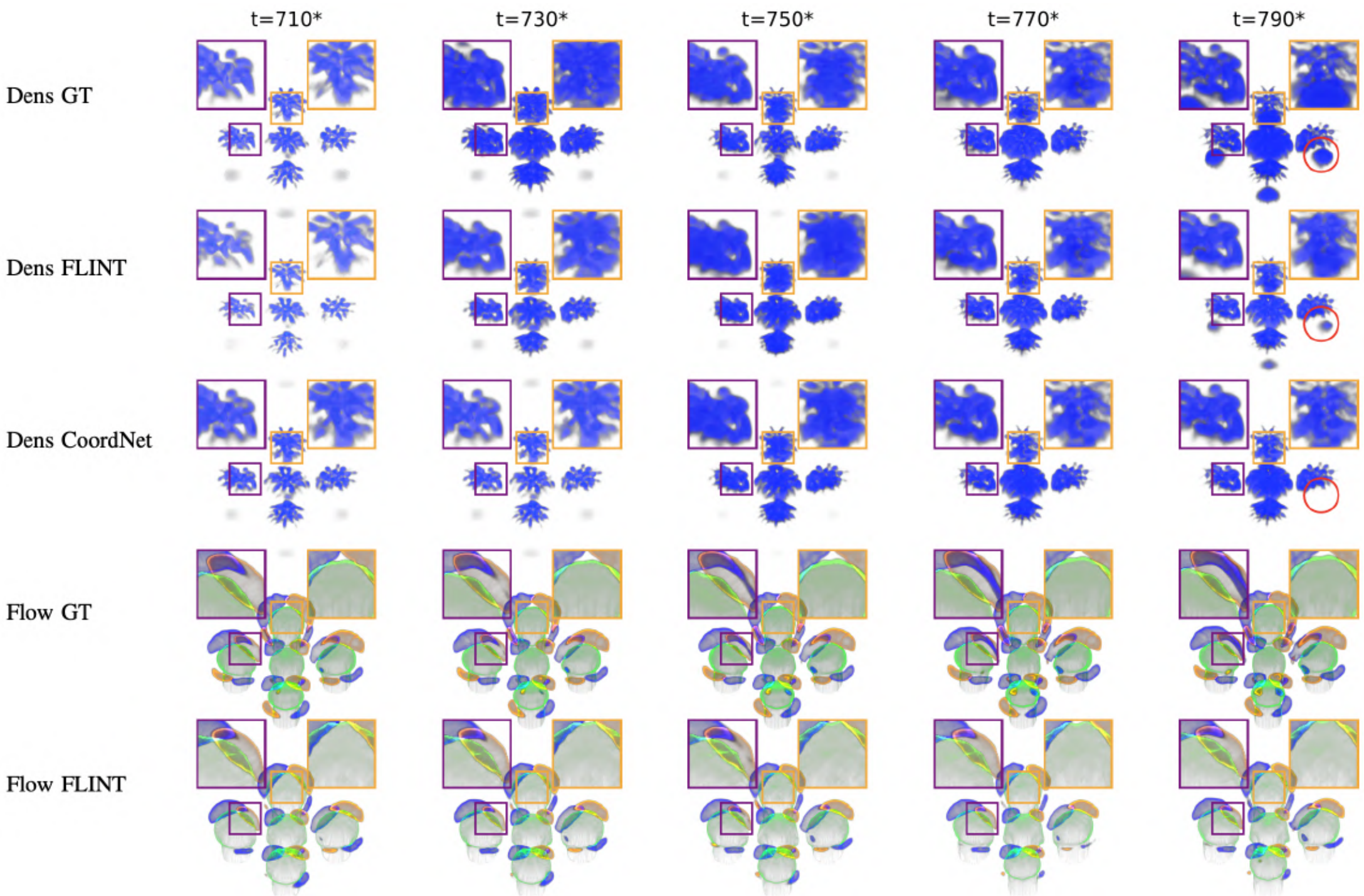

Figure 4.12: 5Jets: FLINT flow field estimation and temporal density interpolation during inference, 20×. From top to bottom, the rows show GT density, FLINT interpolated density, CoordNet interpolated density, GT flow, and FLINT estimated flow. ● ● ● colors represent $x$, $y$, and $z$ flow directions, respectively, transfer functions are in Figure 4.20.

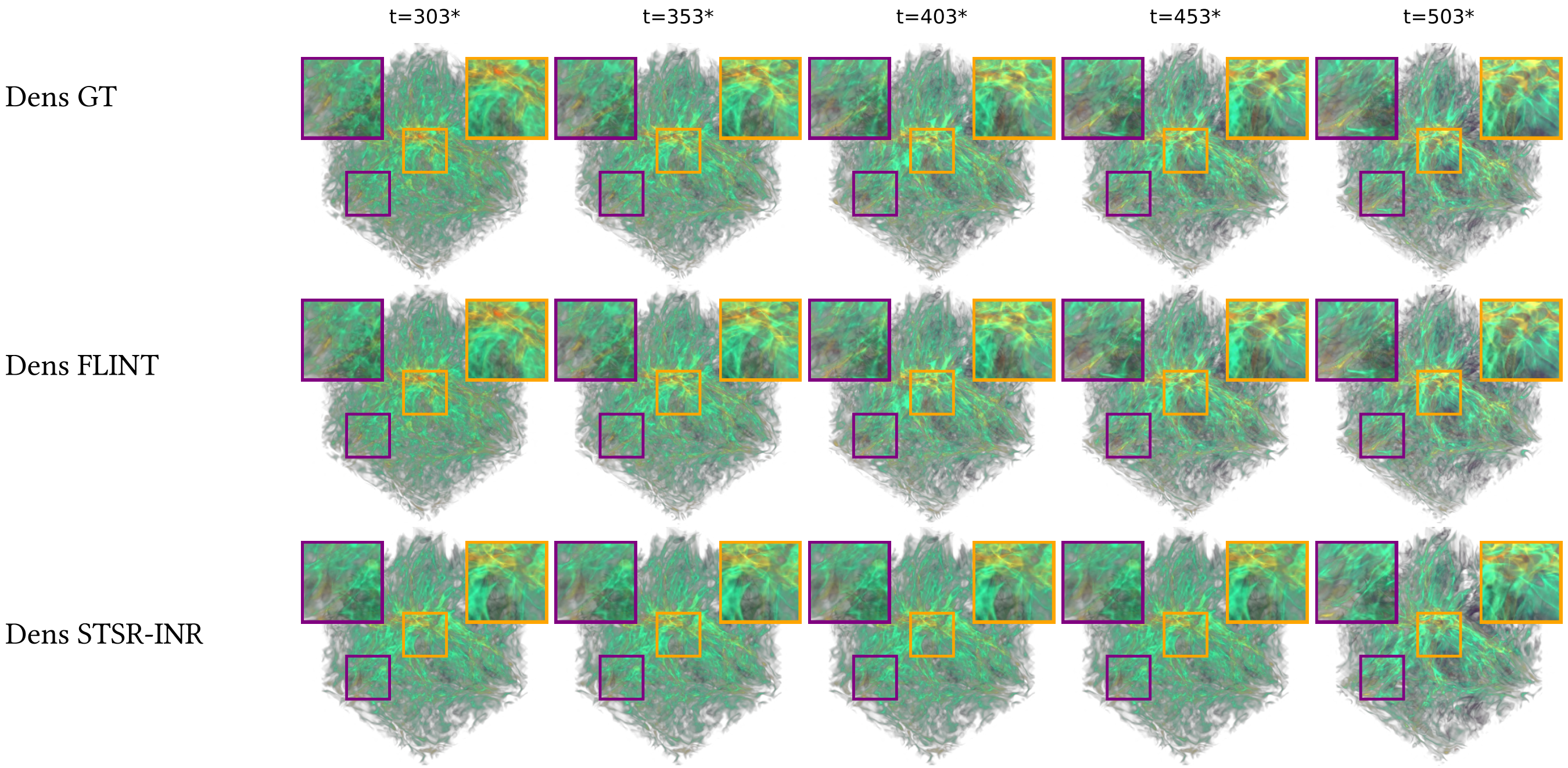

Figure 4.13: Nyx: FLINT temporal density interpolation during inference, 5×. From top to bottom, the rows show GT density, FLINT interpolated density, and STSR-INR interpolation. 3D rendering was used for the density visualization, transfer functions are in Figure 4.19.

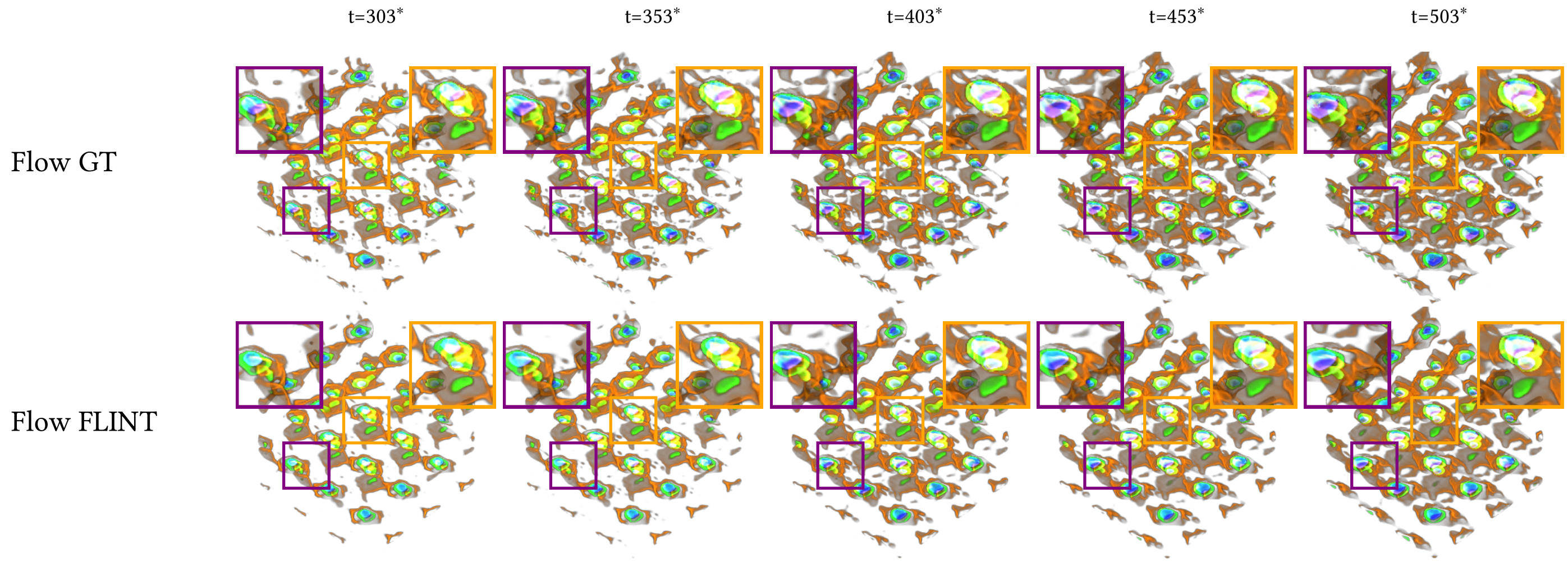

Figure 4.14: Nyx: FLINT flow field estimation during inference, 5×. From top to bottom, the rows show GT flow and FLINT flow estimation. 3D rendering was used for the flow field visualization (● ● ● colors represent $x$, $y$, $z$), transfer functions are in Figure 4.19.

### 4.5.4 *Nyx: Density & Flow Available for Some Timesteps*

Fourth, we consider the scenario where GT density and velocity fields are available for some members of the entire 3D ensemble. As Figure 4.13 illustrates, FLINT achieves accurate performance in terms of both density field interpolation and flow field estimation. Visually, the difference between the renderings of the reconstructed density field and its GT is minimal. Moreover, even at a relatively high interpolation rate of 5×, FLINT effectively learns a flow field that structurally resembles the GT flow. When comparing our FLINT method to STSR-INR for temporal density interpolation, it is evident that the interpolation of the density field is more accurate with FLINT. For example, at $t = 503$, STSR-INR shows a different structure and less dark matter density compared to FLINT, which preserves density. This trend is consistently observed across all shown timesteps in Figure 4.13. For this comparison, we used the same model size for both FLINT and STSR-INR. By default, Implicit Neural Representation (INR)-based models like CoordNet and STSR-INR have smaller model sizes compared to FLINT due to their simpler architecture, where data is represented as a continuous function parameterized by a compact set of weights. In contrast, FLINT employs a more complex design with multiple convolutional blocks, enabling it to achieve accurate flow estimation and scalar field interpolation. To ensure a fair comparison, we configured STSR-INR to have the same model size as FLINT, allowing us to directly assess their performance under comparable conditions.

Crucially, FLINT not only reconstructs the density field but also supplements it with accurate flow information, as shown in Figure 4.14, a feature that STSR-INR lacks. When examining the flow, we observe circular swirling patterns, indicating the complex dynamics of the baryonic gas. As these flows intensify, we see evidence of dark matter moving outward—reflected in both the GT and FLINT density, especially in the orange zoom-ins—consistent with an expanding universe. This underlines the utility of FLINT in capturing not just the static density fields but also the dynamic evolution of the cosmic structures. Examples of the metric scores for interpolation rates of 3×, 5×, and 8× can be found in Table 4.2.

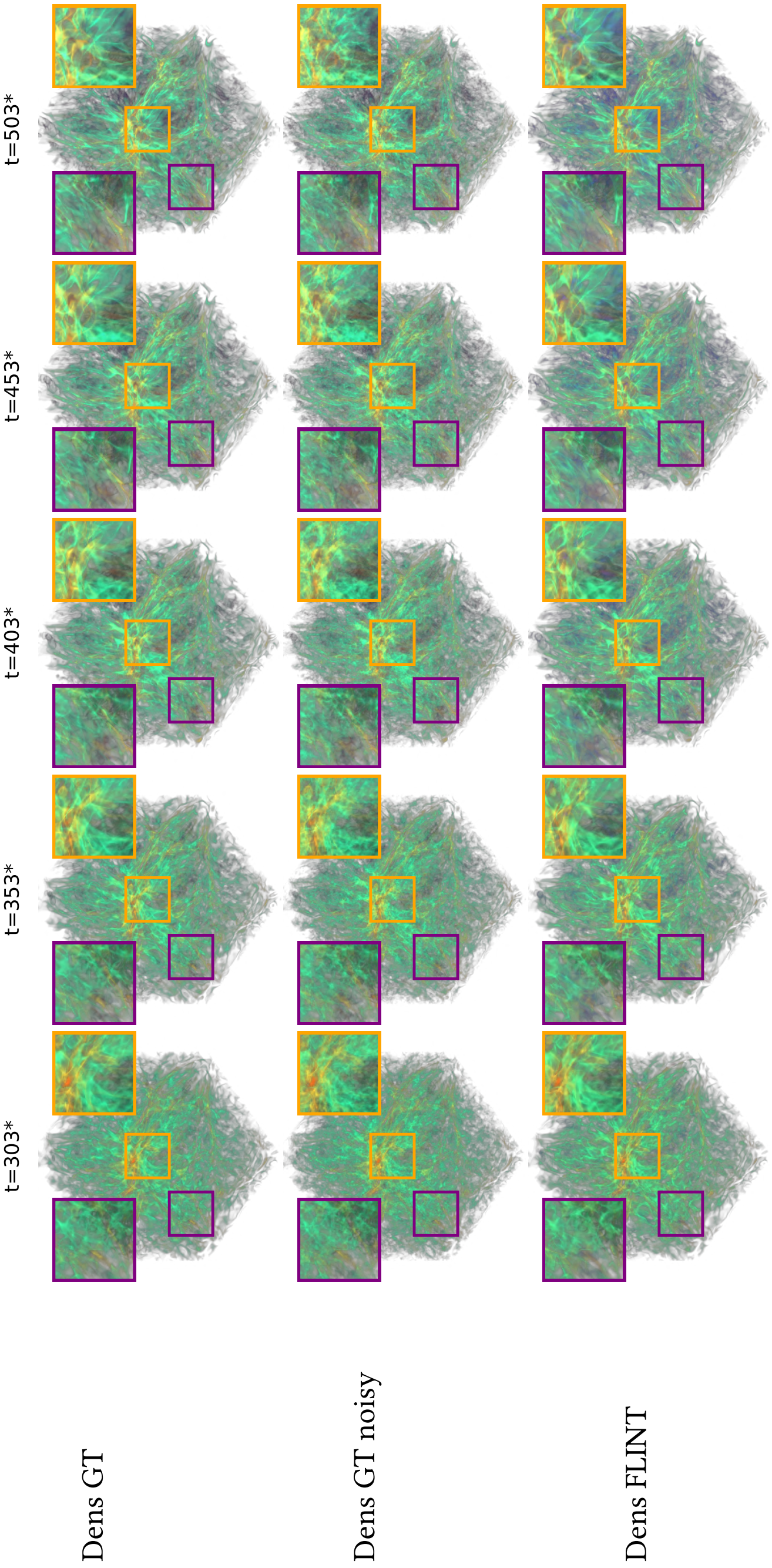

Figure 4.15: Nyx: FLINT temporal density interpolation during inference under noisy conditions, 5×. From top to bottom, the rows show GT density, FLINT interpolated density, and STSR-INR interpolation.

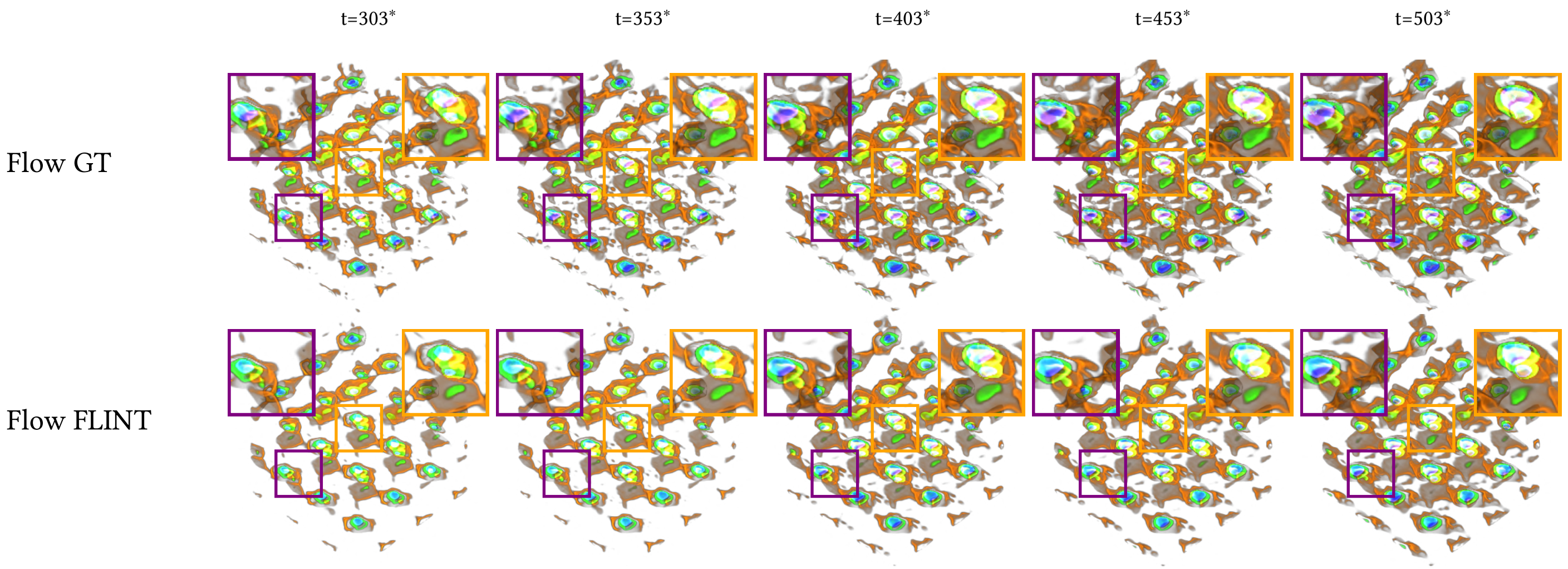

Figure 4.16: Nyx: FLINT flow field estimation during inference under noisy conditions, 5×. From top to bottom, the rows show GT flow and FLINT flow estimation. 3D rendering was used for the flow field visualization (● ● ● colors representing $x$, $y$, and $z$ flow directions respectively).

### 4.5.5 *Noise Handling*

FLINT's robustness to noisy data was evaluated in both 2D and 3D scientific ensembles. The 2D Droplets ensemble naturally contains noisy input data, as evidenced by the density plots in Figure 4.11 top row, where background noise is clearly visible. Despite this inherent noise, FLINT successfully reconstructs density and flow fields, maintaining high accuracy and capturing the essential dynamics of the dataset.

Additionally, to further test FLINT's robustness, we introduced random Gaussian noise into the 3D Nyx simulation ensemble. Specifically, we added noise with a mean of zero and a standard deviation of 0.025 to the density field (ranging within $[0, 1]$ after normalization). These parameters were selected to simulate realistic imperfections without overwhelming the dataset, allowing us to assess FLINT's ability to maintain accuracy under noisy conditions. The results, illustrated in Figure 4.15 (second row corresponds to the GT density with added noise), demonstrate that FLINT's performance does not significantly degrade under these conditions. The model continues to produce accurate density and flow estimations (Figures 4.15 and 4.16, respectively), even in the presence of added noise. These findings highlight FLINT's capability to handle real-world imperfections in data, making it a reliable solution for noisy scientific ensembles.

### 4.5.6 *Evaluation on Temperature and Flow Fields*

In Figure 4.17, we present qualitative results for the Nyx simulation ensemble, 5×. The temperature interpolation demonstrates good performance, with the overall structure and spatial coherence of the field being well preserved across timesteps. For the flow fields estimated from the temperature field, some deviations are observed compared to the GT (Figure 4.18). These deviations are expected, as predicting flow from temperature is inherently more challenging due to the weaker relationship between temperature distributions and physical motion. Despite this, the results still show FLINT's capability to generate reasonable flow estimations even from scalar fields like temperature, which are less directly correlated with movement.

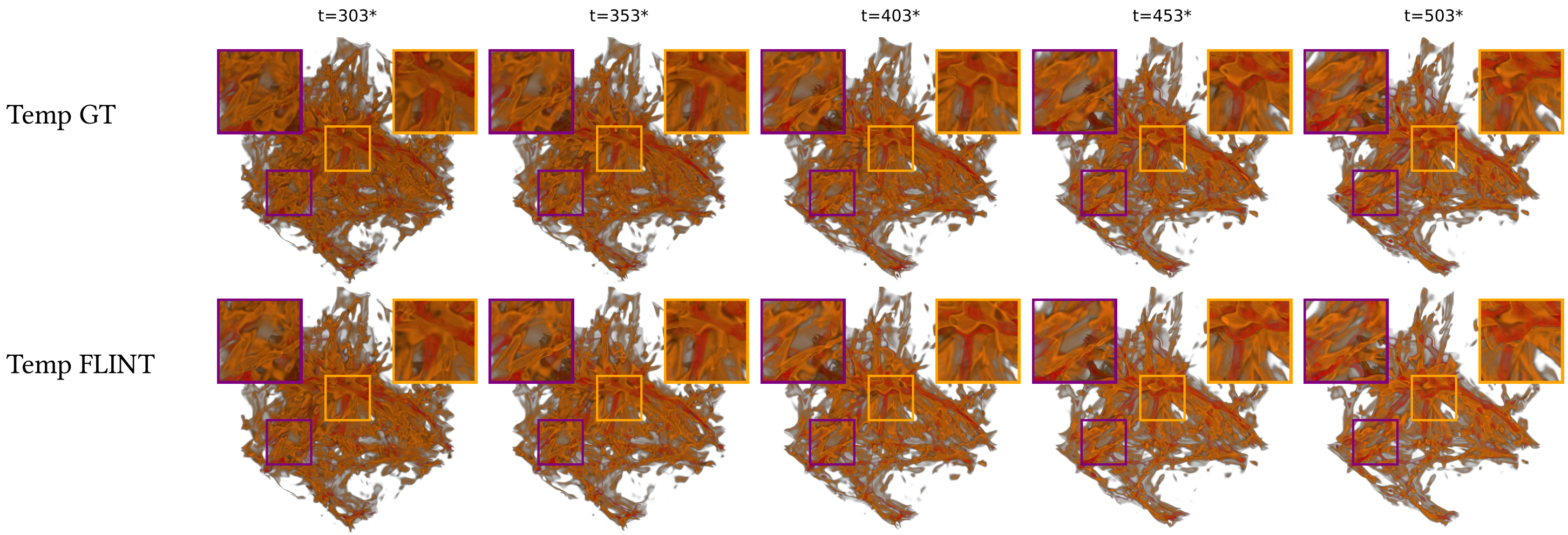

Figure 4.17: Nyx: FLINT temporal temperature field interpolation during inference, 5×. From top to bottom, the rows show GT temperature and FLINT interpolated temperature. 3D rendering was used for the temperature field visualization.

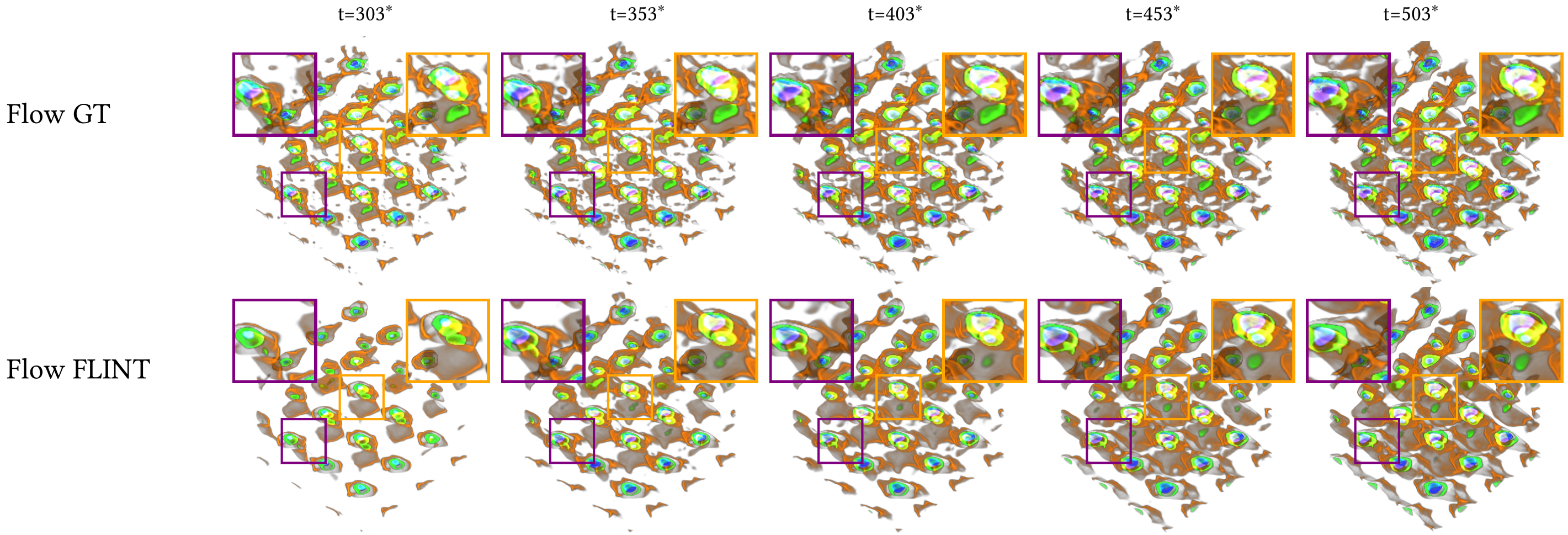

Figure 4.18: Nyx: FLINT flow field estimation during inference, 5×. From top to bottom, the rows show GT flow and FLINT flow estimation. 3D rendering was used for the flow field visualization (● ● ● colors representing $x$, $y$, and $z$ flow directions respectively).

### 4.5.7 *Nyx: Domain Expert Evaluation*

A domain expert—an assistant professor in astronomy with a research focus on cosmological simulations and observational data evaluation—provided insights into FLINT's utility for analyzing scalar fields and velocity information in the Nyx dataset. He highlighted that FLINT's ability to estimate velocity fields alongside density fields offers significant advantages. The expert remarked, "compared to other reconstruction methods, with FLINT you get information not only on the *density* in Nyx simulations but also on the *velocity*". This dual capability is particularly important in cosmology, where velocity data derived from "redshift" measurements are critical for estimating distances and analyzing spatial relationships between objects.

The expert stated that "having a way to estimate accurately the peculiar velocities would help when comparing with observations". He further underlined the relevance of velocity information in creating "lightcones", which combine simulation timesteps to mimic observations of the universe at different epochs. These lightcones are crucial for studying phenomena such as the intergalactic medium and the 3D distribution of galaxies, where "the distance information comes really from a velocity shift". He emphasized that "having a way to estimate accurately the velocities would help when comparing with observations". By accurately reconstructing both density and velocity fields, FLINT enhances the fidelity of simulation outputs aligned with observational data, bridging gaps between simulations and real-world measurements, and enabling a deeper understanding of cosmological dynamics.

While the expert's feedback highlights clear cosmological applications, our current evaluation does not directly employ galaxy-scale cosmology datasets with features such as redshift-based velocity shifts or lightcone structures. This choice was driven by the computational and storage demands of running and processing such simulations, as well as the focus of this work on validating FLINT 's methodology across representative scalar and vector field datasets. Future work might involve applying FLINT to full cosmological simulation outputs to directly demonstrate its utility in the scenarios described by the domain expert.

### 4.5.8 *5Jets and Nyx Transfer Functions*

The transfer functions used for visualizing the density field and the $x$, $y$, $z$ components of the flow field for the 5Jets and Nyx dataset are illustrated in Figures 4.20 and 4.19. These transfer functions were selected experimentally to ensure that the rendered volumes are the most representative of the underlying data, effectively highlighting the key structures and dynamics.

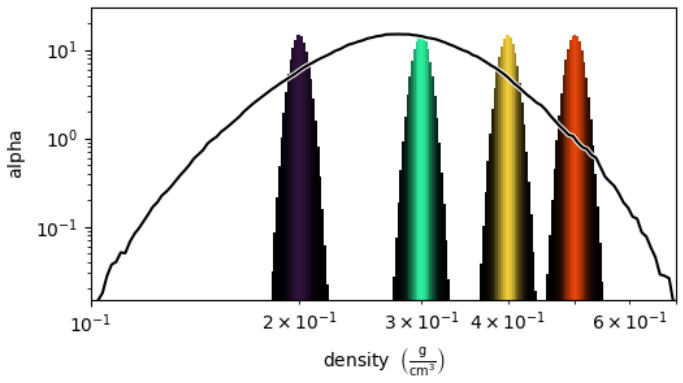

(a) TF for density field.

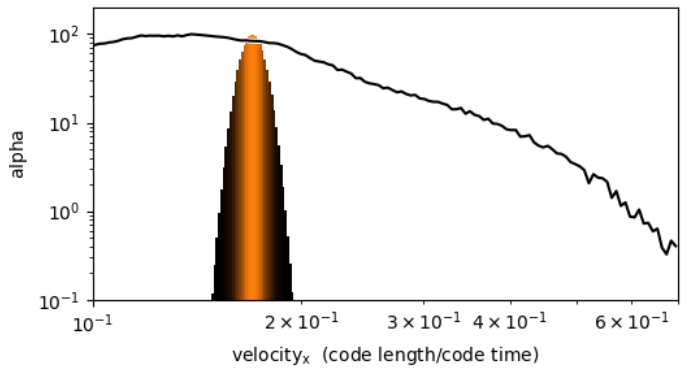

(b) TF for flow field in $x$ direction.

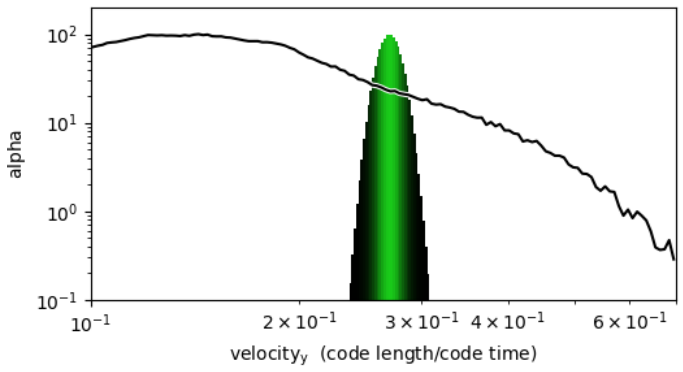

(c) TF for flow field in $y$ direction.

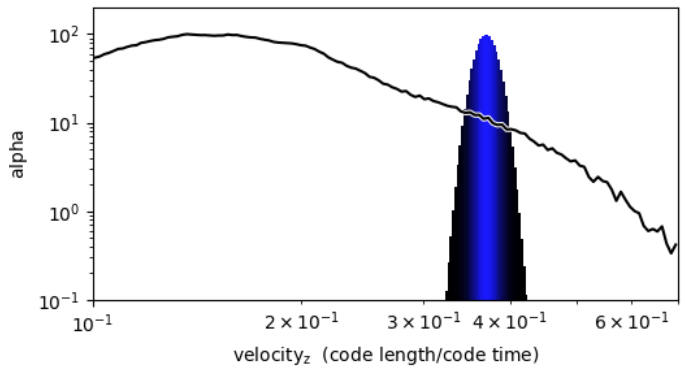

(d) TF for flow field in $z$ direction.

Figure 4.19: Transfer functions for density and $x$, $y$, $z$ components of the flow field, Nyx.

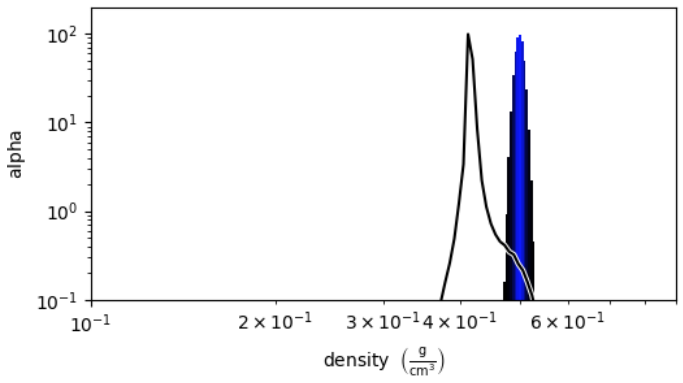

(a) TF for density field.

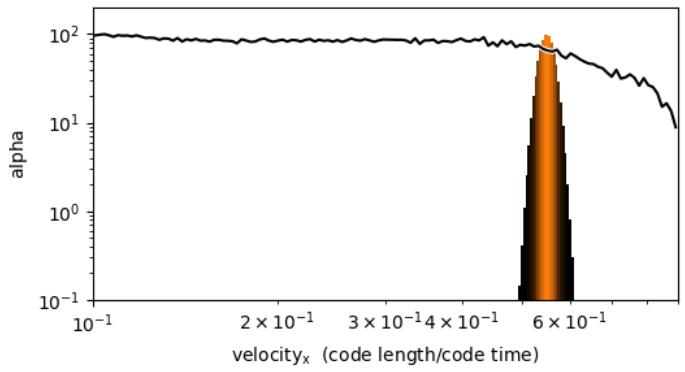

(b) TF for flow field in $x$ direction.

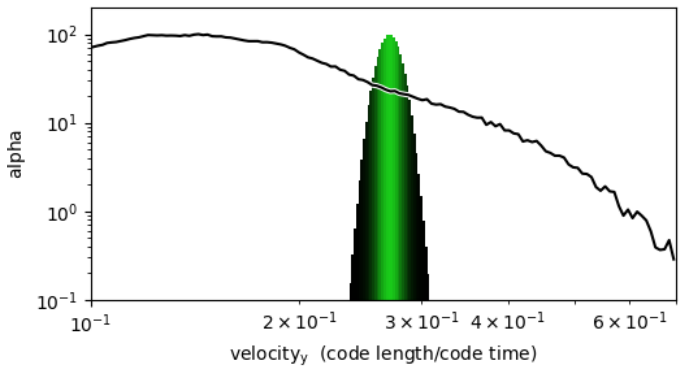

(c) TF for flow field in $y$ direction.

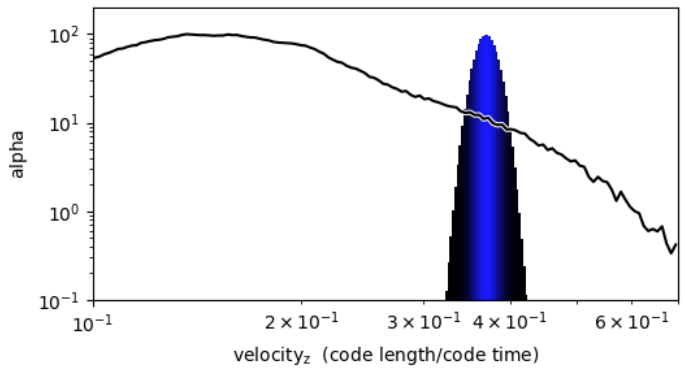

(d) TF for flow field in $z$ direction.

Figure 4.20: Transfer functions for density and $x$, $y$, $z$ components of the flow field, 5Jets.

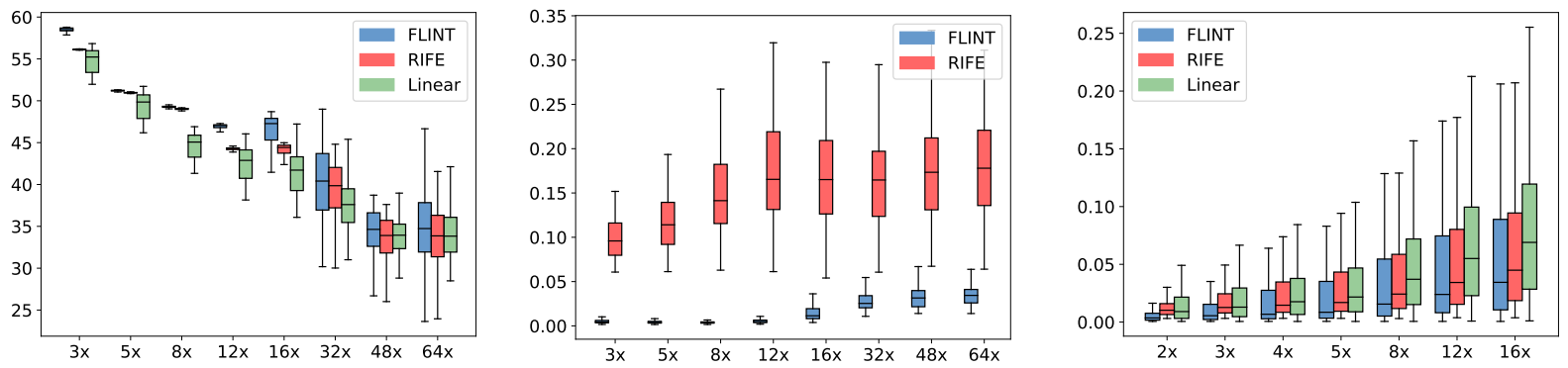

(a) LBS: PSNR ↑ scores (b) LBS: EPE ↓ scores (c) Droplets: LPIPS ↓ scores

Figure 4.21: Comparison of FLINT, RIFE, and linear interpolation at various interpolation rates. For PSNR plots for the Droplets dataset and LPIPS scores for the LBS dataset, see Figures 4.22a and 4.22b, respectively.

## 4.6 QUANTITATIVE AND COMPARATIVE EVALUATION

In this section, we present quantitative results and compare against baseline methods to demonstrate the improvement achieved with our proposed method, followed by ablation and parameter studies to explore different configurations and hyperparameters.

### 4.6.1 *Comparison Against Baselines*

We illustrate quantitative results using boxplots in Figure 4.21 for LBS and Droplets ensembles. FLINT (blue) is compared against two baselines, RIFE (red) and linear interpolation (green), across various interpolation factors ranging from 2 to 64, showcasing representative outcomes for both ensembles. Similarly, in LPIPS-based evaluation for the Droplets, FLINT demonstrates better performance. Furthermore, examining the flow estimation (supplementing density with flow information during inference) results for the LBS ensemble in Figure 4.21b reveals that FLINT achieves a significantly lower endpoint error with minimal variance when compared to RIFE. This underscores FLINT's proficiency in learning an accurate flow representation that closely aligns with the GT physical flow, particularly evident in the LBS ensemble scenario. As linear interpolation cannot provide flow estimation results, no statistics are shown for it, in the case of the LBS ensemble.

In the case of the Droplets ensemble, FLINT outperforms linear interpolation for both LPIPS and PSNR metrics, confirming our expectations, as shown in Figure 4.21c and Figure 4.22a. This superior performance can be attributed to FLINT's capacity to learn robust optical flow, which in turn enhances its ability to interpolate and reconstruct complex density fields. FLINT consistently performs better than RIFE as well, across all presented interpolation rates, for both PSNR and LPIPS metrics. This performance reaffirms that our proposed model is well-suited for the challenging task of reconstructing density fields within scientific ensembles. These findings underscore the potential of FLINT in spatiotemporal data interpolation tasks, with application for scientific visualization and data analysis.

In the context of the Droplets ensemble, the results presented in Figure 4.22a provide evidence of FLINT's superior performance compared to linear interpolation. This observation holds true across a spectrum of interpolation rates, according to our expectations. FLINT also consistently surpasses the RIFE method in terms of both PSNR and, previously shown in Figure 4.21c, LPIPS metric scores.

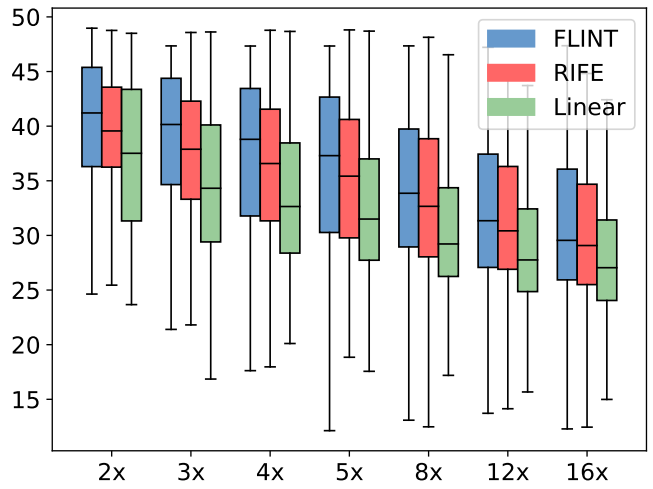

(a) Droplets ensemble: comparison of PSNR scores of FLINT, RIFE, and linear interpolation.

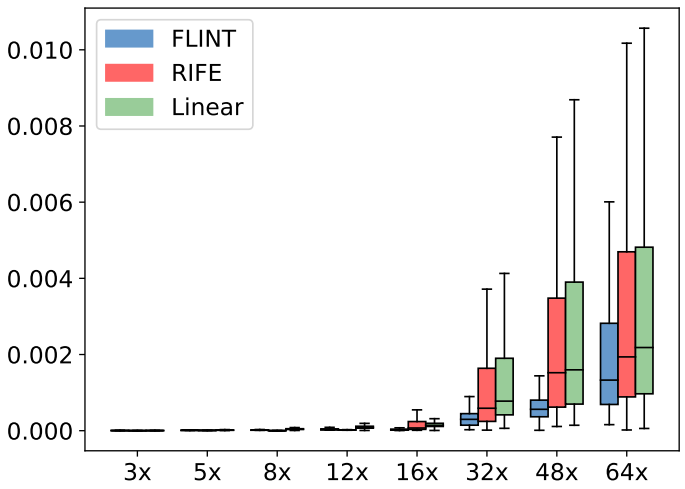

(b) LBS ensemble: comparison of LPIPS scores of FLINT, RIFE, and linear interpolation.

Figure 4.22: Comparison of PSNR and LPIPS scores for the Droplets and LBS ensembles at various interpolation rates.

In the context of the LBS ensemble, the results presented in Figure 4.22b demonstrate FLINT's superior performance compared to both linear interpolation and the RIFE method, also when evaluated using the LPIPS metric. Across varying interpolation rates, FLINT consistently achieves lower LPIPS scores, signifying better perceptual similarity to the ground truth. These findings underscore FLINT's ability to preserve fine details and produce visually coherent interpolations, even under challenging conditions.

CoordNet (Han and Wang, 2022), another method we considered for comparative evaluation, offers an advanced framework for various tasks in time-varying volumetric data visualization, including TSR. It has shown improvements over the TSR-TVD method (Han and Wang, 2019) discussed in Section 4.2, positioning it as a relevant benchmark for temporal interpolation performance. We trained CoordNet for the 3D+time 5Jets dataset and compared across various interpolation rates, see Table 4.1. FLINT demonstrates competitive performance in terms of PSNR score for density interpolation while also serving the dual purpose of enabling flow estimation. While CoordNet facilitates TSR, it does not extend to flow field estimation, which we consider to be the main contribution of our work.

STSR-INR (Tang and Wang, 2024), another method we considered for comparative evaluation, was applied to the 3D+time Nyx dataset and compared across various interpolation rates, as shown in Table 4.2.

Table 4.1: Comparison against baselines, 5Jets

| **Method** | **10×** | | **15×** | | **20×** | |
|---|---|---|---|---|---|---|
| | **PSNR ↑** | **EPE ↓** | **PSNR ↑** | **EPE ↓** | **PSNR ↑** | **EPE ↓** |
| FLINT | 46.72 | 0.7271 | **45.41** | 0.7286 | **44.93** | 0.7292 |
| Linear | 44.70 | – | 42.28 | – | 38.11 | – |
| CoordNet | **48.45** | – | 44.89 | – | 43.74 | – |

STSR-INR is an INR-based approach that employs a variable embedding scheme to learn latent vectors for different variables. This method utilizes a variational auto-decoder to optimize the learnable latent vectors, enabling latent-space interpolation. STSR-INR shows improvements over both STNet (Han et al., 2021) and CoordNet (Han and Wang, 2022), positioning it as a relevant benchmark for TSR performance. FLINT demonstrates superior performance in terms of PSNR score for density interpolation while also serving the dual purpose of enabling flow estimation, a capability that extends beyond the functionalities provided by STSR-INR.

Table 4.2: Comparison against baselines, Nyx

| **Method** | **3×** | | **5×** | | **8×** | |
|---|---|---|---|---|---|---|
| | **PSNR ↑** | **EPE ↓** | **PSNR ↑** | **EPE ↓** | **PSNR ↑** | **EPE ↓** |
| FLINT | **53.17** | 0.0310 | **52.31** | 0.0310 | **49.39** | 0.0309 |
| Linear | 47.49 | – | 41.92 | – | 37.51 | – |
| STSR-INR | 49.63 | – | 44.21 | – | 41.09 | – |

**Evaluation on Additional Scalar Field.** In addition to evaluating FLINT for density and flow fields, we extended our analysis to another scalar field to demonstrate its versatility. Specifically, we evaluated FLINT's performance on the temperature field from the Nyx simulation ensemble. We examined FLINT's ability to interpolate the temperature field and estimate the corresponding flow fields. Qualitative results are shown in Figures 4.17 and 4.18, and quantitative evaluations are provided in Table 4.3. The temperature with flow results indicate that the temperature interpolation performance remains robust across different interpolation rates, although the PSNR values are slightly lower compared to density results. This difference reflects the increased complexity of interpolating temperature fields due to their distinct data characteristics. Similarly, the flow results show lower EPE scores. This is expected, as estimating flow from temperature is inherently more challenging in comparison to density due to its weaker correlation with movement or physical flow dynamics.

Our findings align with observations by Gu et al. (2022), where the performance of vector field reconstruction can vary depending on the input scalar field. Similarly, FLINT's performance showed variations when applied to the temperature field, reflecting the challenges associated with different data characteristics.

These results further demonstrate FLINT's adaptability across diverse scalar fields while highlighting the influence of field-specific dynamics on reconstruction accuracy.

Table 4.3: Comparison of different fields, Nyx

| **Field** | **3×** | | **5×** | | **8×** | |
|---|---|---|---|---|---|---|
| | **PSNR ↑** | **EPE ↓** | **PSNR ↑** | **EPE ↓** | **PSNR ↑** | **EPE ↓** |
| Dens. + Flow | 53.17 | 0.0310 | 52.31 | 0.0310 | 49.39 | 0.0309 |
| Temp. + Flow | 49.32 | 0.0446 | 46.91 | 0.0451 | 44.43 | 0.0467 |

**Impact of Noise.** To evaluate FLINT's robustness to noise, we introduced random Gaussian noise into the 3D Nyx simulation ensemble. Even with the added noise in the GT density, FLINT achieved reliable results, maintaining an average PSNR of 46.36 dB and an EPE of 0.0356 at a 5× interpolation rate. In comparison, without noise, the model achieved a higher PSNR of 52.31 dB and a slightly lower EPE of 0.0310, demonstrating a modest decline in performance due to noise. This highlights FLINT's resilience to noisy data and reinforces its applicability to real-world scenarios where imperfections in scientific datasets are common.

**Large Interpolation Rates.** At large interpolation rates, such as 32×, we observe that while density interpolation (Figure 4.23) maintains acceptable quality (average PSNR of 37.71 dB), the accuracy of flow estimation (Figure 4.24) noticeably degrades (average EPE is 0.0479). In comparison, at 8× interpolation, the model achieved a PSNR of 49.39 dB and an EPE of 0.0309, while at 16×, the PSNR was 43.42 dB with an EPE of 0.0338. The degradation at a large interpolation rate is expected due to the increased structural differences between flows at widely separated time steps, which make flow estimation inherently more challenging. Nevertheless, FLINT delivers satisfactory results for scalar field interpolation even at these demanding rates. Qualitatively, while the main structure, such as the distribution of dark matter, is preserved and remains visually consistent with the GT, finer details and intricate patterns become less accurate.

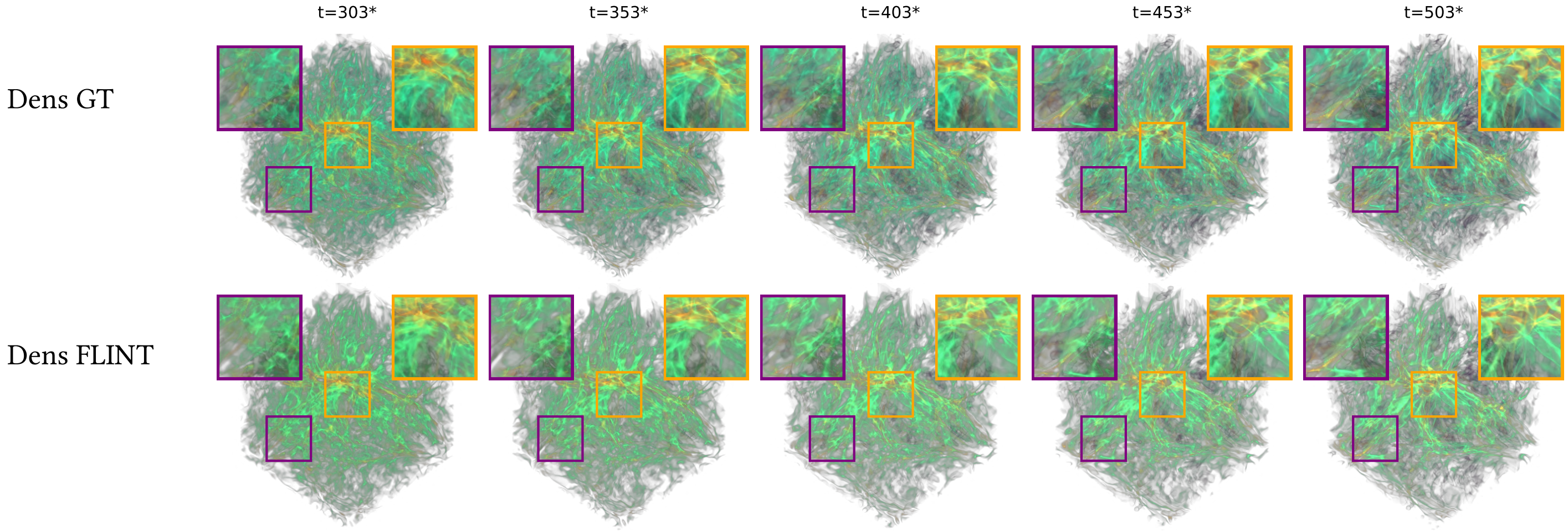

Figure 4.23: Nyx: FLINT temporal density interpolation during inference, 32×. From top to bottom, the rows show GT density and FLINT interpolated density. 3D rendering was used for the density field visualization.

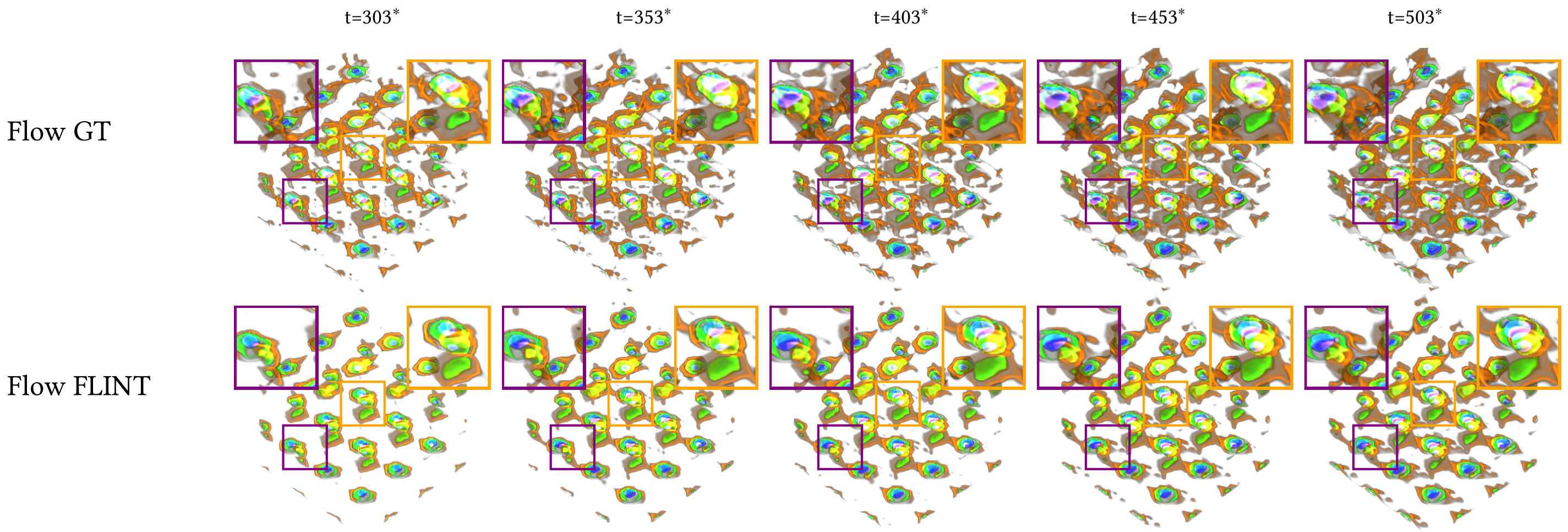

Figure 4.24: Nyx: FLINT flow field estimation during inference, 32×. From top to bottom, the rows show GT flow and FLINT flow estimation. 3D rendering was used for the flow field visualization (● ● ● colors representing $x$, $y$, and $z$ flow directions respectively).

**Inference time comparison.** We conducted an inference time comparison of our model, FLINT, against RIFE, CoordNet, and STSR-INR across all datasets. Specifically, we compared FLINT with RIFE on the 2D ensembles (LBS and Droplets) and with CoordNet and STSR-INR on the 3D datasets (5Jets and Nyx). For each evaluation, we measured the inference time over 3K timesteps for the 2D ensembles and 300 timesteps for the 3D ensembles from the test set. Our findings indicate that FLINT achieves faster inference time compared to RIFE, with an average of 0.025 seconds per timestep on the 2D datasets, as opposed to RIFE's 0.035 seconds. On the 3D datasets, FLINT significantly outperforms both CoordNet and STSR-INR in inference speed, even achieving interactive rates with an average time of 0.2 seconds per timestep for FLINT. In comparison, it takes 2.1 seconds for CoordNet and 1.5 seconds for STSR-INR. The faster inference time of FLINT compared to INR-based approaches like CoordNet and STSR-INR can be attributed to the inherent efficiency of convolutional neural networks (CNNs), which excel in parallel processing and optimized memory access, while INR-based methods require solving implicit functions at each point, adding computational complexity.

### 4.6.2 *Ablation Studies*

Our proposed FLINT method is a deep neural network that has various loss components for training the network. To assess the impact of these in achieving optimal results, we conducted a series of ablation studies.

The ablation studies conducted on FLINT provide valuable insights into the impact of different loss components on the overall performance. In Table 4.4, which presents the results for the LBS, 5Jets, and Nyx datasets with available GT flow fields for the model training, four variants of FLINT are compared: "FLINT no *flow*", "FLINT no *rec*", "FLINT w/o s-t", and "FLINT". "FLINT no *flow*" refers to FLINT without the inclusion of the flow-related loss component. As expected, this variant demonstrates considerably less accurate physical flow, measured by the EPE metric, despite a relatively high density interpolation score, measured by PSNR metric. "FLINT no *rec*" refers to FLINT without the reconstruction-related loss component. In this case, the flow learned by the model is relatively good, however, the density interpolation suffers due to the absence of its related loss component. "FLINT w/o s-t" refers to FLINT where the teacher block (see Figure 4.2) and the teacher-related terms in the loss functions are removed. This scenario leads to slightly less accurate interpolation and flow estimation due to the fact that the model does not receive GT densities during the training. Finally, "FLINT" represents the complete FLINT model with all loss components. It achieves the highest scores across all metrics, including PSNR and EPE, outperforming all other variants. This demonstrates the signifi-cance of incorporating the proposed loss components in achieving superior interpolation results.

Table 4.4: Ablation of FLINT (flow-supervised)

| Method | LBS, 3× | | 5Jets, 10× | | Nyx, 5× | |
|---|---|---|---|---|---|---|
| | PSNR ↑ | EPE ↓ | PSNR ↑ | EPE ↓ | PSNR ↑ | EPE ↓ |
| FLINT no *flow* | 57.41 | 0.0842 | 37.80 | 3.4985 | 51.97 | 0.1461 |
| FLINT no *rec* | 49.89 | 0.0119 | 32.32 | 0.7332 | 44.67 | 0.0497 |
| FLINT w/o s-t | 56.37 | 0.0064 | 41.99 | 0.7323 | 52.08 | 0.0334 |
| FLINT | **58.44** | **0.0051** | **46.72** | **0.7271** | **52.31** | **0.0310** |

Table 4.5: Ablation of FLINT, Droplets, 2× (flow-unsupervised)

| Method | PSNR ↑ | LPIPS ↓ |
|---|---|---|
| FLINT no *rec* | 32.09 | 0.0373 |
| FLINT no *dis* | 39.87 | 0.0129 |
| FLINT no *photo* | 40.47 | 0.0105 |
| FLINT no *reg* | 40.71 | 0.0098 |
| FLINT no *dis, photo* | 40.36 | 0.0116 |
| FLINT no *dis, reg* | 40.37 | 0.0117 |
| FLINT no *reg, photo* | 40.11 | 0.0110 |
| FLINT no *dis, reg, photo* | 39.81 | 0.0119 |
| FLINT w/o s-t | 40.25 | 0.0202 |
| FLINT | **41.16** | **0.0087** |

Similar ablation studies were performed for the Droplets ensemble, for which no GT flow fields are available, as shown in Table 4.5. Multiple variants of FLINT were evaluated, including "FLINT no *rec*", "FLINT no *dis*" (without distillation), "FLINT no *photo*" (without photometric loss), "FLINT no *reg*" (without regularization), ablation of combinations of these losses, and "FLINT w/o s-t". Each variant is compared to the complete FLINT method. The results indicate that each loss component contributes to the overall performance of FLINT. The complete FLINT model outperforms all the variants in terms of PSNR and LPIPS scores. This showcases the importance of reconstruction, distillation, photometric, and regularization loss components in achieving the best possible reconstructions for the Droplets ensemble.

### 4.6.3 *Parameter Studies*

Our proposed FLINT method involves several hyperparameters, and we performed extensive parameter studies to understand their impact on achieving optimal results. Table 4.6 displays the outcomes of these studies regarding 2D+time and 3D+time datasets.

Table 4.6: Hyperparameter search for FLINT

| **Method** | **LBS, 3×** | | **5Jets, 10×** | | **Nyx, 5×** | | **Droplets, 2×** | |
|---|---|---|---|---|---|---|---|---|
| | **PSNR ↑** | **EPE ↓** | **PSNR ↑** | **EPE ↓** | **PSNR ↑** | **EPE ↓** | **PSNR ↑** | **LPIPS ↓** |
| FLINT *128* | 49.86 | 0.0135 | 43.57 | 0.7397 | 51.68 | 0.0315 | 36.89 | 0.0162 |
| FLINT *Lapl* | 57.23 | 0.0112 | 42.82 | 0.7387 | 51.73 | 0.0319 | 39.44 | 0.0192 |
| FLINT $\lambda_{flow} = 0.3$ | 58.03 | 0.0109 | 44.89 | 0.7308 | 52.29 | 0.0311 | – | – |
| FLINT $\lambda_{flow} = 0.1$ | 58.17 | 0.0113 | 45.11 | 0.7549 | 52.19 | 0.0320 | – | – |
| FLINT $\gamma = 0.9$ | 55.81 | 0.0130 | 44.84 | 0.7327 | 52.14 | **0.0310** | – | – |
| FLINT $\gamma = 0.7$ | 56.28 | 0.0137 | 46.08 | 0.7384 | 52.28 | 0.0317 | – | – |
| FLINT $\lambda_{reg} = 10^{-9}$ | – | – | – | – | – | – | 40.80 | 0.0089 |
| FLINT $\lambda_{reg} = 10^{-7}$ | – | – | – | – | – | – | 40.06 | 0.0122 |
| FLINT $\lambda_{photo} = 10^{-7}$ | – | – | – | – | – | – | 40.79 | 0.0107 |
| FLINT $\lambda_{photo} = 10^{-5}$ | – | – | – | – | – | – | 38.98 | 0.0144 |
| FLINT $\lambda_{dis} = 1 \times 10^{-5}$ | – | – | – | – | – | – | 37.86 | 0.0157 |
| FLINT $\lambda_{dis} = 1 \times 10^{-3}$ | – | – | – | – | – | – | 40.34 | 0.0114 |
| FLINT *smooth* | – | – | – | – | – | – | 36.96 | 0.0125 |
| FLINT *stride = 1* | 51.33 | 0.0147 | 43.41 | 0.8116 | 52.19 | 0.0418 | 38.64 | 0.0149 |
| FLINT *3 Blocks* | 53.08 | 0.0136 | 43.09 | 0.7346 | 52.28 | 0.0312 | 39.74 | 0.0144 |
| FLINT *4 Blocks* | **58.44** | **0.0051** | **46.72** | **0.7271** | **52.31** | **0.0310** | **41.16** | **0.0087** |
| FLINT *5 Blocks* | 57.20 | **0.0051** | 45.57 | 0.7420 | 52.15 | **0.0310** | 41.04 | 0.0113 |
| FLINT *6 Blocks* | 56.17 | 0.0106 | 44.61 | 0.7691 | 51.98 | 0.0313 | 39.79 | 0.0134 |
| FLINT *two-stage* | 56.87 | 0.0069 | 45.78 | 0.7409 | 52.07 | **0.0310** | 40.95 | 0.0108 |

The parameter search investigates various configurations of FLINT, each labeled with a specific identifier. To determine the model's optimal architecture for our task, we conducted a series of experiments, adjust-ing the model's capacity by varying its width and depth. For example, "FLINT *128*" represents a configuration with 128 channels in all convolutional layers across all blocks. Through hyperparameter optimization, we found that the best-performing model, named "FLINT" or "FLINT *4 Blocks*", increases both the channel capacity and the number of blocks. This configuration, with 4 blocks and channel counts in the convolutional layers ranging from 256 to 128, is designed to capture more intricate and detailed features within the data. Another configuration is "FLINT *smooth*", where a smoothness loss component was incorporated into the training process for Droplets, however, not yielding the best results. Additionally, "FLINT *stride = 1*" represents an architecture where each block uses convolutional and deconvolutional layers with a stride of one, instead of varying strides. Furthermore, we explored two different loss functions for the interpolation part of the model. The configuration "FLINT *Lapl*" indicates the use of the Laplacian loss, which measures the $L_1$-loss between two Laplacian pyramid representations (pyramidal level is 5) of the reconstructed density field and GT density field. The configuration "FLINT" employs the simple $L_1$-distance as the loss function, as described in Section 4.3.3. In addition, we varied the number of blocks of FLINT (*Conv Block* in Figure 4.2), ranging from three to six. As can be seen from Table 4.6, the most optimal configuration is the one with four blocks ("FLINT *4 Blocks*"). Models with fewer than four blocks lack sufficient capacity, while those with more begin to overfit. The configuration "FLINT *two-stage*" employs a two-stage optimization approach where a teacher model is trained first, followed by a student network trained to align with the teacher's outputs. In contrast, "FLINT" achieves slightly better performance overall while reducing training time by nearly a half (12 vs. 20 hours).

The results in Table 4.6 demonstrate optimized FLINT's superior performance over other configurations, with the highest scores in PSNR and EPE metrics. This underscores the importance of channel capacity, number of blocks, and optimized loss functions in capturing the intricate dynamics of fluid ensembles, enabling FLINT's best density interpolation and flow estimation capabilities.

## 4.7 CONCLUSION AND FUTURE WORK

In this work, we proposed FLINT, a learning-based approach for the estimation of flow information and scalar field interpolation for 2D+time and 3D+time scientific ensembles. FLINT offers flexibility in handling various data availability scenarios. It can perform flow-supervised learning to estimate flow fields for members when partial flow data is available. In cases where no flow data is provided, common in experimental datasets, FLINT employs a flow-unsupervised approach, generating flow fields based on optical flow concepts. Additionally, FLINT generates high-quality temporal interpolants between scalar fields like den-sity or luminance, outperforming recent state-of-the-art methods. It achieves fast inference and does not require complex training procedures, such as pre-training or fine-tuning on simplified datasets, commonly required by other flow estimation methods (Teed and Deng, 2020; Dosovitskiy et al., 2015; Luo et al., 2021). Its effectiveness in producing high-quality flow and scalar fields has been validated across both simulation and experimental data.

For future work, we aim to improve the FLINT method and expand its application to various problems and domains. FLINT is generally applicable to various fields, including density, flow, energy, as well as to grayscale or RGB datasets, and is valuable for visualization applications, and we aim to study a larger range of applications in future work. FLINT extensions could include expanding support for more timesteps as input, exploring extrapolation in addition to interpolation, or ensemble parameter space exploration. Another promising direction for future work is harnessing the learned flow field for efficient dimensionality reduction (DR). The compressed representation of the estimated flow could be used in order to perform DR of the ensemble members and compared against standard and ML-based DR techniques (Gadirov et al., 2021). Furthermore, in the medical domain, learned optical flow can serve as additional input to classification models (Wang and Snoussi, 2015). This strengthens disease classification tasks by incorporating motion fields, which convey crucial information. As a result, leveraging estimated flow enhances the accuracy and robustness of disease classification models, ultimately contributing to improved healthcare outcomes.

Building on the foundations established with FLINT, the next chapter introduces HyperFLINT—an enhanced version that incorporates hypernetworks to further improve performance and adaptability. While FLINT focused on accurate flow estimation and temporal interpolation, HyperFLINT extends these capabilities with parameter-aware adaptation, enabling dynamic adjustments based on ensemble parameters. This advancement not only boosts reconstruction accuracy but also opens the door to parameter space exploration for deeper insights into scientific simulations.

# 5 HYPERFLINT: HYPERNETWORK-BASED FLOW ESTIMATION AND TEMPORAL INTERPOLATION FOR SCIENTIFIC ENSEMBLE VISUALIZATION

*In the previous chapter, we introduced FLINT, a deep learning-based approach that enables flow estimation and temporal interpolation simultaneously for scientific ensemble data. FLINT demonstrated its ability to reconstruct missing time steps while also predicting flow fields, enhancing the analysis of spatiotemporal datasets. In this chapter, we present its extension, HyperFLINT, which integrates hypernetworks to further improve accuracy and expand the model's capabilities. Beyond the original tasks of flow estimation and interpolation, HyperFLINT introduces a parameter-aware adaptation mechanism, allowing it to dynamically adjust based on ensemble parameters. This not only enhances the accuracy of predictions but also enables parameter space exploration, facilitating deeper insights into scientific simulations. By leveraging hypernetworks, HyperFLINT addresses limitations in previous approaches, making it a powerful tool for analyzing complex ensemble datasets.*

## 5.1 INTRODUCTION

Building on the challenges outlined in the previous chapter, where large-scale spatio-temporal ensembles require efficient reconstruction and analysis due to storage constraints and limited experimental modalities, we now extend our focus to overcoming the limitations of existing approaches in parameter space generalization. While methods such as FLINT (Gadirov et al., 2025a) and other recent techniques (Han and Wang, 2022; Wu et al., 2023; Tang and Wang, 2024) have demonstrated success in reconstructing scalar and flow fields, they primarily focus on data interpolation without explicitly accounting for ensemble parameters. FLINT's student-teacher framework has achieved state-of-the-art results in density interpolation and flow estimation, benefiting applications such as flow visualization (Jänicke et al., 2011), timestep selection (Frey and Ertl, 2017), and ensemble member comparison (Tkachev et al., 2021). However, its reliance on fixed training distributions limits its adaptability to varying simulation conditions. To overcome these limitations, we introduce HyperFLINT, which leverages a hypernetwork to dynamically incorporate ensemble parameters. This enables HyperFLINT to adapt to diverse simulation conditions, enhancing flow estima-

This chapter is based on the paper "HyperFLINT: Hypernetwork-based Flow Estimation and Temporal Interpolation for Scientific Ensemble Visualization" (Gadirov et al., 2025b) tion, improving interpolation, and supporting robust parameter space exploration. By embedding parameter-aware flexibility, HyperFLINT addresses FLINT's constraints, providing a versatile solution for large-scale scientific ensembles.

In this chapter, we introduce HyperFLINT (Hypernetwork-based FLow estimation and temporal INTerpolation), a deep learning approach that uses hypernetworks to estimate missing flow fields in scientific ensembles. HyperFLINT leverages the simulation parameters via hypernetworks to generate accurate flow fields for each timestep, even in scenarios where the flow could not be captured or was omitted (see overview in Figure 5.1). Additionally, HyperFLINT is capable of producing high-quality temporal interpolants between scalar fields, providing a comprehensive solution for reconstructing both flow and scalar data. HyperFLINT improves upon existing state-of-the-art methods, such as FLINT (Gadirov et al., 2025a), discussed in the previous chapter, by focusing on scientific ensembles and incorporating hypernetworks to account for simulation parameters, resulting in better temporal interpolation and flow estimation. Unlike previous approaches, HyperFLINT handles complex spatio-temporal datasets without requiring domain-specific assumptions, pre-training, or fine-tuning on simplified datasets.

The ability of HyperFLINT to incorporate parameter-driven transformations into its model architecture opens up new possibilities for parameter space exploration. By learning the relationships between simulation parameters and output data, HyperFLINT can generate approximations of data for configurations that were not explicitly simulated, allowing for a broader investigation of possible simulation outcomes without rerunning the full simulation for each parameter set. This capacity is particularly valuable for understanding complex dependencies between parameters in ensemble simulations, as it enables researchers to interpolate or extrapolate within the parameter space. With HyperFLINT, users can systematically explore how changes in parameters impact the generated fields, offering insights into the underlying phenomena that would otherwise require extensive computational resources to simulate directly. Moreover, HyperFLINT's parameter-driven capabilities allow for the reconstruction of missing data, including timesteps and individual physical fields (e.g., velocity), which is especially useful in large-scale simulations where only a subset of the data can be saved (Childs et al., 2020). For instance, HyperFLINT enables the application of flow visualization techniques to datasets that originally only contained scalar fields, broadening the scope of analysis for experimental data. In this way, HyperFLINT not only enhances data reconstruction but also supports a comprehensive exploration of simulation parameters, making it a powerful tool for both analysis and parameter space exploration.

In summary, in this chapter we propose HyperFLINT, a flexible and novel method for flow estimation, temporal density interpolation, and parameter space exploration in spatio-temporal scientific ensembles using hypernetworks. The HyperFLINT code is available at: `https://github.com/HamidGadirov/HyperFLINT`. Our key contributions can be summarized as follows:

- To the best of our knowledge, HyperFLINT is the first approach that employs a hypernetwork to adaptively estimate flow fields and produce state-of-the-art temporal interpolations of density fields for scientific ensembles by dynamically conditioning on simulation parameters.
- HyperFLINT effectively handles spatio-temporal ensembles utilizing a hypernetwork, requiring no specific assumptions, making it versatile for diverse scientific applications.
- With the introduction of the hypernetwork, HyperFLINT achieves several key advancements: it dynamically generates simulation parameter-aware weights facilitating the understanding of spatio-temporal scientific ensembles, enhancing model quality and performance by adapting to varying simulation parameters, enabling significantly improved flow estimation and scalar field interpolation, even in scenarios with sparse or incomplete data.

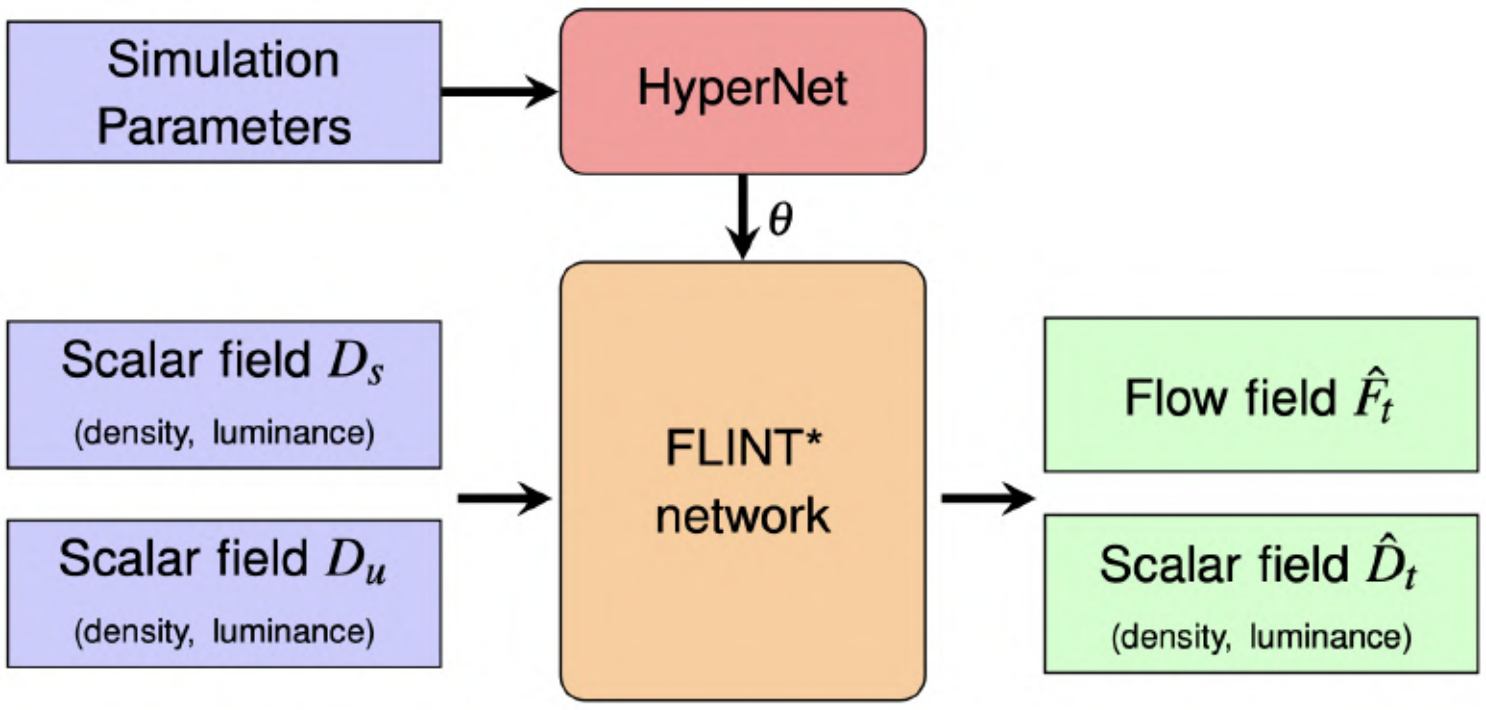

Figure 5.1: Overview of the HyperFLINT pipeline during inference. The FLINT* deep neural network, whose weights are generated by the HyperNet, performs flow field estimation $\hat{F}_t$ and temporal (scalar) field interpolation $\hat{D}_t$, where $s < t < u$, by utilizing the available densities $D_s$ and $D_u$ from the previous and following timesteps, and their simulation parameters that are dynamic across datasets.

## 5.2 RELATED WORK

**Hypernetworks in deep learning.** Hypernetworks (Ha et al., 2016) have gained traction in the deep learning community as an effective mechanism for generating the weights of another neural network, thereby offering greater flexibility in capturing complex relationships between data and model parameters. These networks excel in tasks where parameter space exploration is crucial, as they can dynamically adjust the parameters of a target network based on external inputs, such as simulation parameters. Hypernetworks have been successfully applied in various fields, including neural architecture search (Brock et al., 2017), generative modeling (Skorokhodov et al., 2021), and meta-learning (Przewieźlikowski et al., 2023). By leveraging hypernetworks, HyperFLINT extends the capabilities of traditional neural networks for flow estimation in scientific ensembles, enabling it to adapt to different simulation conditions without requiring retraining. This makes HyperFLINT highly versatile and capable of handling a wide range of scenarios, significantly improving both accuracy and efficiency in flow estimation and temporal interpolation tasks.

**Flow field estimation in scientific visualization.** Several methods have been developed for flow field estimation and visualization in scientific datasets. Kappe *et al.* (Kappe et al., 2015) focused on estimating 3D local flow in microscopy data using a combination of image processing and optical flow techniques, while Kumpf *et al.* (Kumpf et al., 2018) applied ensemble sensitivity analysis with optical flow-based feature tracking to study changes in geo-spatial data. Manandhar *et al.* (Manandhar et al., 2018) proposed dense 3D optical flow estimation for microscopy image volumes using displacement vectors. However, these approaches often rely on specific assumptions, limiting their applicability to real-world scientific ensembles. In contrast, HyperFLINT operates without dataset-specific assumptions, making it versatile for diverse scientific visualization tasks. Sahoo *et al.* (Sahoo and Berger, 2022) introduced Integration-Free Learning of Flow Maps, which estimates flow directly from state observations, enhancing efficiency for large-scale datasets where ground-truth (GT) flow may not be available. FLINT (Gadirov et al., 2025a) emerged as a state-of-the-art approach for reconstructing scalar fields and simultaneously estimating flow fields in spatio-temporal scientific datasets (see Chapter 4). FLINT's student-teacher architecture and flexible loss function helped achieve high accuracy in density interpolation and flow estimation tasks, particularly in 2D+time and 3D+time datasets. FLINT's modular design features a series of convolutional and deconvolutional layers grouped into neural blocks, which iteratively refine scalar field and flow field outputs. Temporal consistency is enforced through specific loss components, ensuring alignment with GT data and improving interpolation accuracy. These capabilities made FLINT particularly effective for addressing missing data in large-scale simulations. However, FLINT exhibits limitations in dynamic simulation scenarios where varying ensemble parameters affect data behavior. Without mechanisms to explicitly incorporate these parameters, FLINT's adaptability and generalization remain constrained.

To address these challenges, we extend FLINT's foundational ideas in our proposed HyperFLINT framework. HyperFLINT replaces the student-teacher setup with a hypernetwork that dynamically generates weights for FLINT*, a streamlined adaptation of the original FLINT network (see Section 5.3.5). By conditioning these weights on simulation parameters, HyperFLINT achieves enhanced flexibility and precision, enabling accurate flow estimation, temporal interpolation, and parameter space exploration.

**Machine learning-based upscaling & super-resolution.** Across multiple fields, including image processing, computer vision, and scientific visualization (Ledig et al., 2017; Shi et al., 2016), ML-based upscaling and super-resolution approaches have gained considerable attention. These techniques generally target either spatial, temporal, or combined spatio-temporal enhancements, forming three main categories: spatial super-resolution (SSR), temporal super-resolution (TSR), and spatio-temporal super-resolution (STSR). SSR approaches, exemplified by models such as SRCNN (Dong et al., 2015), SRFBN (Li et al., 2019), and SwinIR (Liang et al., 2021), focus on increasing spatial detail by generating realistic textures and enhancing fine structures. TSR methods, on the other hand, aim to fill in intermediate frames in time-subsampled sequences without degrading spatial quality. Methods like phase-based interpolation (Meyer et al., 2015), SepConv (Niklaus et al., 2017), and SloMo (Jiang et al., 2018) are representative TSR techniques that focus on temporal resolution improvements in videos. While some prior approaches, such as STNet (Han et al., 2021), address either spatial or temporal resolution, they typically do not tackle both simultaneously. For example, TSR-TVD (Han and Wang, 2019), a recurrent generative network proposed by Han and Wang, was designed to temporally upscale a variable different from the given one, without explicitly addressing both spatial and temporal dimensions. Volume scene representation networks (V-SRN) (Lu et al., 2021; Park et al., 2019; Sitzmann et al., 2019; Mildenhall et al., 2021) have significantly advanced neural representations for volumetric data, enabling high-quality rendering and reconstruction through implicit neural representations. Building on these advancements, fV-SRN (Weiss et al., 2022) leverages GPU tensor cores to integrate neural reconstruction into on-chip ray tracing kernels, reducing computational complexity and accelerating training and inference for real-time volume rendering. Recent advancements, such as Filling the Void (Mishra et al., 2024), SSR-TVD (Han and Wang, 2020), FFEINR (Jiao et al., 2024), HyperINR (Wu et al., 2023), CoordNet (Han and Wang, 2022), and STSR-INR (Tang and Wang, 2024), have shown improvements in handling either TSR or SSR of data fields at arbitrary resolutions. FLINT (Gadirov et al., 2025a) can perform temporal interpolation for both 2D+time and 3D+time datasets but does not consider the parameter space of scientific ensembles, which limits its adaptability to different scenarios. oration. In contrast, HyperFLINT introduces hyper-networks to incorporate ensemble parameters directly into the interpolation process, enabling a more adaptive and data-driven approach.

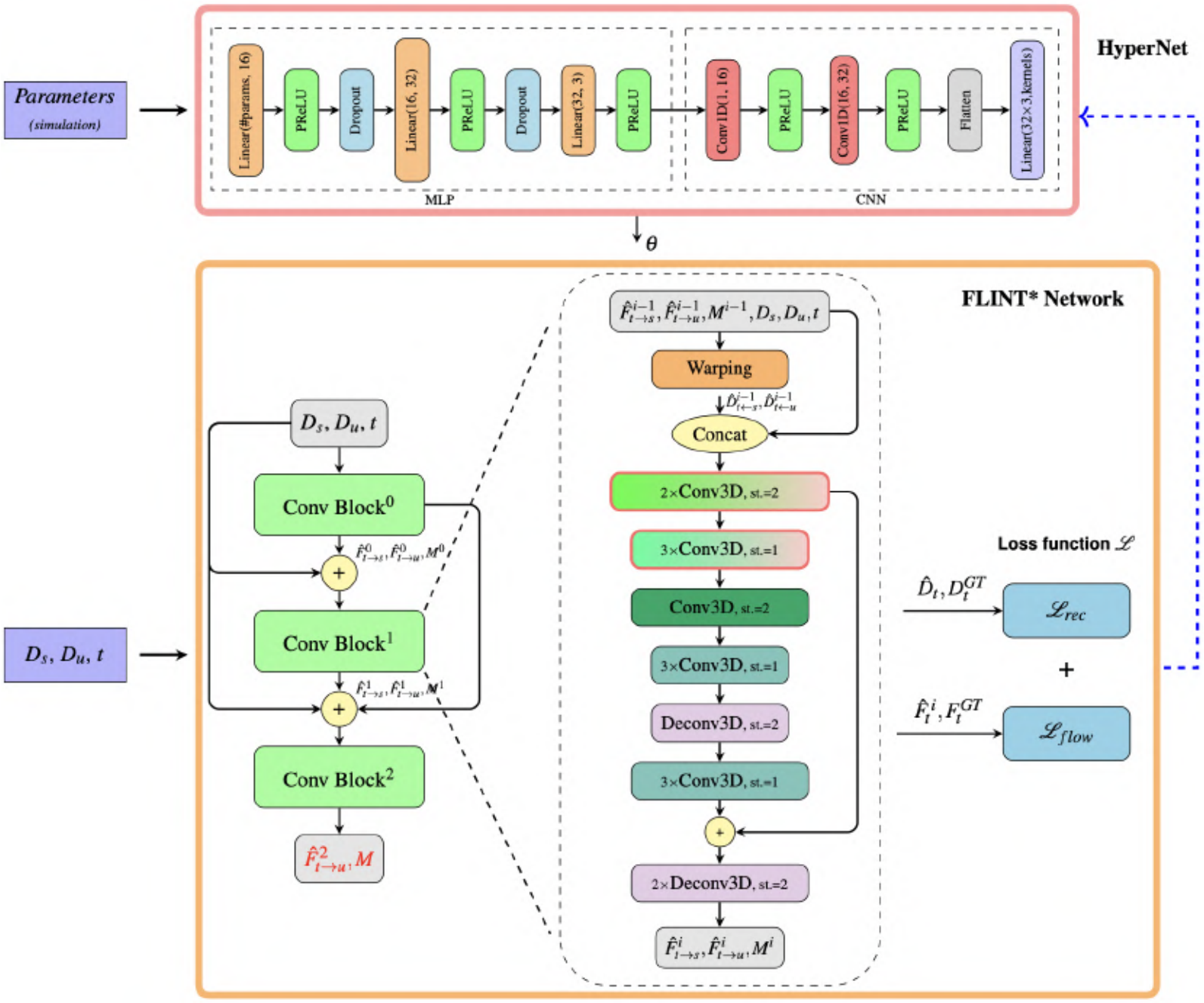

Figure 5.2: HyperFLINT network architecture and pipeline during training: Given the input fields $D_s$ and $D_u$, and their simulation parameters, HyperFLINT predicts the $\hat{D}_t$ scalar field and $\hat{F}_t^i$ flow fields used in the loss function for optimizing network parameters. The HyperFLINT model consists of two key components: the HyperNet and the main network, FLINT*. The HyperNet, depicted within the red box, generates weights for the convolutional layers of FLINT* (with red outlines in the middle column of the orange box). The FLINT* model architecture and loss function are shown in the orange box. The model consists of several stacked blocks of the convolutional network, which takes $D_s$, $D_u$, and $t$ as input and in the $i^{th}$ Conv Block computes estimated flows $\hat{F}^i_{t\to s}$, $\hat{F}^i_{t\to u}$, and fusion mask $M^i$ used for interpolation. The zoomed-in view on the right highlights the structure of a generic Conv Block. The GT density $D_t^{GT}$ and flow $F_t^{GT}$ is only used in the loss function $\mathcal{L}$. The blue dashed arrow in the right of the figure represents the gradient propagation during training, from the output of FLINT* back to the HyperNet.

## 5.3 METHOD

To handle diverse parameter settings in spatio-temporal scientific ensembles, we introduce HyperFLINT, a method that integrates a hyper-network to dynamically adapt the main neural network, FLINT*. The hypernetwork generates weights based on simulation parameters. As Figure 5.2 illustrates, the pipeline consists of three main components: (1) the hypernetwork (HyperNet), which takes simulation parameters as input and produces weights for FLINT*, which is a simplified and streamlined version of the original FLINT network; (2) the FLINT* network, which estimates flow fields and interpolates scalar data across time steps; and (3) a training framework that optimizes both flow estimation and interpolation through a combination of loss functions. Unlike traditional methods, HyperFLINT does not require pre-training or fine-tuning on simplified datasets, enabling efficient adaptation to new scenarios. The following sections elaborate on the neural network architecture of HyperFLINT (Section 5.3.1), describe the temporal interpolation and flow estimation pipeline used in both training and inference (Section 5.3.3), and detail our proposed loss function (Section 5.3.4). A detailed comparison between HyperFLINT and FLINT can be found in Section 5.3.5.

### 5.3.1 *HyperFLINT Network Architecture*

The architecture of HyperFLINT integrates two neural networks: HyperNet and FLINT*. HyperNet (Figure 5.2, top), described in Section 5.3.2, dynamically generates weights for the FLINT* network (Figure 5.2, bottom), which is inspired by the FLINT method (Gadirov et al., 2025a). The architecture of the FLINT* network comprises $N = 3$ stacked convolutional blocks (*Conv Block*), each incorporating convolutional (*Conv*) and deconvolutional (*Deconv*) layers. The inclusion of HyperNet, which provides adaptive, parameter-driven weight adjustments, enables FLINT* to flexibly handle varying simulation parameters and improve generalization across different scenarios. The expanded view of a single convolutional block is presented in the middle column of the orange box in Figure 5.2. The first five *Conv* layers (red outlines, middle column) dynamically adjust their weights based on the output from HyperNet. The FLINT* network starts with 128 feature channels in the first block, reduces to 96 channels in the second block, and finishes with 64 channels in the final block. PReLU activation (He et al., 2015) is applied to all layers except the last one, ensuring efficient learning and non-linear transformations. The training process is driven by loss components illustrated on the right side of Figure 5.2, guiding both flow estimation and scalar field interpolation. FLINT* and HyperNet are trained jointly, ensuring that the hypernetwork optimizes the weights of the main network dynamically throughout training.

### 5.3.2 *HyperNet*

The HyperNet architecture, as depicted in Figure 5.2 top part, is an integral part of HyperFLINT. It takes numerical input parameters that characterize the simulation (such as physical quantities or configuration settings) and processes them through two main components: a Multi-Layer Perceptron (MLP) and a Convolutional Neural Network (CNN). First, the MLP, consisting of three linear layers with PReLU activation functions and two dropout layers, transforms the input parameters into a higher-dimensional representation. This embedding allows the network to interpret the simulation parameters in a way that is beneficial for downstream tasks. Once processed by the MLP, the intermediate representation is passed into the CNN, where additional transformations are applied, refining the information derived from the parameters. Finally, the output of the CNN is reshaped into a one-dimensional vector and passed through a final linear layer to generate the weights ($\theta$) for the convolutional layers of the FLINT* network. The choice of one-dimensional convolutions (Conv1D) within the CNN component is motivated by the fact that the input simulation parameters are represented as one-dimensional sequences. Conv1D layers offer an efficient way to capture local relationships between the parameters while preserving their spatial or sequential structure. Compared to fully connected layers, Conv1D can model these relationships in a way that is more computationally efficient, with fewer parameters, and can capture local patterns in parameters that are missed by dense layers. This structured approach helps HyperNet leverage the inherent relationships among parameters to refine the generated weights. The proposed architecture enables parameter-aware learning, where the simulation's governing parameters directly influence the model's internal weights. This not only improves the model's quantitative and qualitative performance but also facilitates parameter space exploration (Section 5.7).

### 5.3.3 *Flow Estimation and Scalar Field Interpolation*

HyperFLINT takes as input the simulation parameters (for HypetNet), two scalar fields $D_s$ and $D_u$ of the same ensemble member at timesteps $s < u$, and an intermediate timestep $t$, where $s < t < u$ (for FLINT*). The goal is to predict the corresponding flow field $\hat{F}_t$ and generate interpolated scalar fields $\hat{D}_t$ for any intermediate time $t \in [s, u]$. To accomplish this, FLINT* first computes intermediate flow fields, $\hat{F}_{t \to s}$ and $\hat{F}_{t \to u}$. The *time-backward* flow field, $\hat{F}_{t \to s}$, represents the flow vectors from the frame at time $t$ to an earlier frame at $s$. Conversely, the *time-forward* flow, $\hat{F}_{t \to u}$, represents flow vectors from the frame at $t$ to a later frame at $u$. These intermediate flow fields are then used to warp scalar fields towards the target time $t$, generating estimates for that time step. In the final step, HyperFLINT combines the intermediate warped scalar fields using a fusion mask $M$ learned by FLINT*, where $M(i, j) \in [0, 1], \forall i, j$, which ensures smooth blending and high-quality interpolation.

**Warping**. In HyperFLINT, we implement the volumetric *backward warping*, where each target voxel $v_t$ identifies its corresponding source voxel $v_s$ based on the flow fields. This mapping, guided by the intermediate flow fields, ensures continuous resampling using trilinear interpolation around $v_s$ to compute the value for $v_t$. We denote the combined effect of reverse mapping and interpolation by the warping operator $\overleftarrow{W}$. Specifically, the mappings are guided by the flow fields, producing the warped scalar fields $\hat{D}_{t\leftarrow s} = \overleftarrow{W}(D_s, \hat{F}_{t\to s})$ and $\hat{D}_{t\leftarrow u} = \overleftarrow{W}(D_u, \hat{F}_{t\to u})$, which represent the values at time $t$ based on the source volumes $D_s$ and $D_u$, respectively; see Figure 5.3.

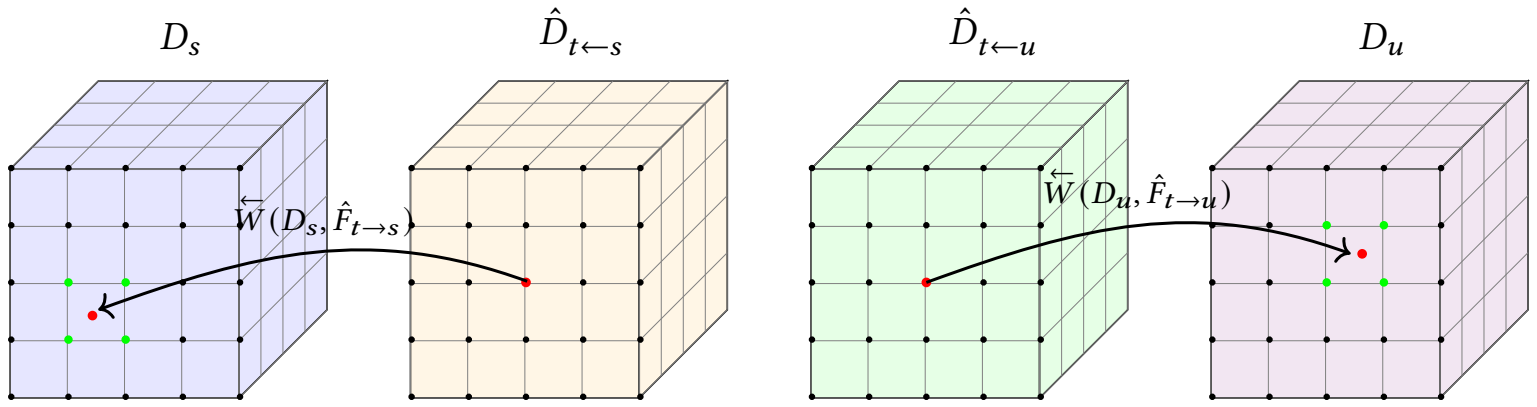

Figure 5.3: Illustration of the 3D backward warping $\overleftarrow{W}$: (scalar) fields $D_s$ and $D_u$ are reversely mapped according to the flow fields $\hat{F}_{t\to s}$ and $\hat{F}_{t\to u}$. The fields $\hat{D}_{t\leftarrow s}$ and $\hat{D}_{t\leftarrow u}$ are then reconstructed using trilinear interpolation considering the values at the coordinates shown with green dots (for the visible front surface of the cube).

In HyperFLINT, iterative refinement within FLINT*'s convolutional blocks is applied similarly to the original FLINT method (Gadirov et al., 2025a). However, FLINT* is optimized specifically for use with a hypernetwork (see Section 5.3.5), the architecture of which was constructed based on a thorough hyperparameter search (Section 5.6.3) to ensure efficiency and compatibility. The interpolated scalar field $\hat{D}_t$, intermediate flow fields $\hat{F}_t^i$, and estimated final flow field $\hat{F}_t$ are computed as follows:

$$\hat{D}_t = \hat{D}_{t\leftarrow s}^{N-1} \odot M + \hat{D}_{t\leftarrow u}^{N-1} \odot (\mathbf{I} - M), \tag{5.1a}$$

$$\hat{F}_t^{i+1} = \hat{F}_{t\to u}^{i}, \text{ for } i = 0, \ldots, N-2, \quad \hat{F}_t = \hat{F}_t^{N-1} \ (N = 3), \tag{5.1b}$$

where $\odot$ denotes element-wise multiplication, $\mathbf{I}$ is the identity matrix, and $M$ is the fusion mask, similarly to Chapter 4.

**Inference**. During the inference phase, HyperFLINT processes the input scalar fields $D_s$ and $D_u$, along with their associated simulation parameters using the fully trained FLINT* and HyperNet networks. Here, GT information and loss functions are absent, as the network operates solely with its learned parameters. The hypernetwork plays a crucial role by dynamically generating the convolutional layer weights (i.e., kernels) for the FLINT* network based on the provided simulation parameters, as illustrated in Figure 5.2. This allows FLINT* to adapt its operations according to the specific characteristics of each input subset. The final outputs, namely the interpolated scalar field $\hat{D}_t$ and the reconstructed flow field $\hat{F}_t$, are computed according to Equations 5.1a and 5.1b respectively; see Figure 5.1. This adaptive and parameter-aware inference pipeline supports high-quality reconstructions across a diverse range of simulation settings.

### 5.3.4 *Loss Function*

The total HyperFLINT loss is a linear combination of reconstruction loss $\mathcal{L}_{rec}$ and flow loss $\mathcal{L}_{flow}$:

$$\mathcal{L} = \mathcal{L}_{rec} + \lambda_{flow}\mathcal{L}_{flow}, \tag{5.2}$$

where $\lambda_{flow} = 0.2$ for balancing total loss scale w.r.t. the reconstruction component (determined experimentally; see Section 5.6.3).

**Scalar field interpolation.** To interpolate scalar fields temporally, we include a loss component to improve the accuracy of the interpolated density field from Equation 5.1a. The reconstruction loss $\mathcal{L}_{rec}$ measures the $L_1$ distance between the GT density $D_t^{GT}$ and the interpolated field produced by the HyperFLINT network:

$$\mathcal{L}rec = \|D_t^{GT} - \hat{D}t\|_1. \tag{5.3}$$

This helps the interpolated scalar field closely match the ground truth, improving the accuracy of the temporal interpolation.

**Flow estimation**. To improve the quality of the learned flow field, we incorporate a flow loss component calculated as the $L_1$ distance between the estimated flow at each network block and the GT flow (used only during training). Accumulating this measure across all blocks, rather than using only the final one, yields better results. Additionally, we adopt exponentially increasing weights for the loss from RAFT (Teed and Deng, 2020), resulting in the following flow loss equation:

$$\mathcal{L}_{\text{flow}} = \sum_{i=1}^{N} \gamma^{N-i} \|F_t^{GT} - \hat{F}_t^i\|_1, \tag{5.4}$$

where $F_t^{GT}$ is the GT flow at time $t$, $\hat{F}_t^i$ is the flow output from the corresponding $i^{th}$ block of the FLINT* network (Equation 5.1b), and $N = 3$ is the number of blocks in the model. We experimentally established the value of $\gamma$ as 0.8, aligning with the RAFT loss and validating this choice through our hyperparameter search.

### 5.3.5 *Comparison between HyperFLINT and FLINT methods*

The key innovation of HyperFLINT is its integration of a hypernetwork that dynamically generates weights for the FLINT* network, a streamlined variant of the original FLINT model. By removing the student-teacher setup, FLINT* reduces complexity while being optimized to work with the hypernetwork. This enables HyperFLINT to condition convolutional layers on simulation parameters, aligning the network more precisely with data characteristics. The result is enhanced accuracy in flow estimation and scalar field interpolation, capturing physical phenomena with greater fidelity than the original FLINT approach (see Section 5.5 for details). Furthermore, the inclusion of a hypernetwork introduces a novel capability to HyperFLINT: parameter space exploration. By conditioning the model on input parameters, HyperFLINT generates predictions for unseen configurations, enabling researchers to estimate outcomes for unsimulated parameter sets. This approach is particularly valuable in computationally expensive scientific studies, offering insights without requiring exhaustive simulations. This flexibility distinguishes HyperFLINT from traditional methods like FLINT, which lack parameter-driven adaptability.

## 5.4 STUDY SETUP

In this section, we describe the training setup, provide an overview of the datasets used in our experiments and discuss the evaluation methods employed to assess the results obtained by HyperFLINT.

### 5.4.1 *Training*

We normalize scalar fields to the [0, 1] range and vector fields component-wise to the [-1, 1] range before training. HyperFLINT is optimized using AdamW (Loshchilov et al., 2017) with early stopping, similarly to FLINT (Gadirov et al., 2025a) to prevent overfitting on the training data. We use an experimentally determined learning rate of $10^{-4}$ with a cosine annealing scheduler that gradually decreases the learning rate to $10^{-5}$ by the end of the training. We train HyperFLINT with mini-batches of size 4 for 3D ensemble datasets used in our study (see Section 5.4.2). We split the set of all available data into training, validation, and test subsets. To support arbitrary interpolation and flow estimation during training, $t \in [s, u]$ is chosen randomly within a maximum time window of size 12, similarly to FLINT (Gadirov et al., 2025a). This window, confirmed through our hyperparameter search, efficiently determines the maximum time gap between sampled timesteps $s$ and $u$ for constructing the training set, allowing for effective interpolation. Our proposed HyperFLINT model is trained for up to 200 epochs on a single Nvidia Titan V GPU (12GB VRAM), which typically takes around 8 hours per dataset. We monitor the validation loss during training, and in all datasets, the model consistently reaches convergence—defined as a plateau in validation loss or early stopping criteria—within this training budget.

### 5.4.2 *Datasets and Evaluation*

We consider two scientific ensemble datasets in our study.

**Nyx.** The first dataset is a 3D+time ensemble (similarly to Section 4.4.2) based on the compressible cosmological hydrodynamics simulation Nyx, developed by Lawrence Berkeley National Laboratory (Sexton et al., 2021). We consider an ensemble comprising 36 members, each consisting of a maximum of 1600 timesteps with a spatial resolution of $128 \times 128 \times 128$. It contains density and the $x$, $y$, $z$ components of velocity. Akin to InSituNet (He et al., 2019), we vary three parameters for ensemble generation: the total matter density ($\Omega_m \in [0.1, 0.2]$), the total density of baryons ($\Omega_b \in [0.0215, 0.0235]$), and the Hubble constant ($h \in [0.55, 0.75]$). We randomly sample a training subset of 500 timesteps and utilize different ensemble members for training, validation, and testing.

**Castro.** The second dataset consists of a 3D+time ensemble based on astrophysical hydrodynamics simulation Castro, simulating the merger of two white dwarfs, developed by the Lawrence Berkeley National Laboratory (Almgren et al., 2010). This ensemble contains 12 members, each with up to 800 timesteps, and a spatial resolution of $128 \times 128 \times 128$. The dataset includes the density field and the $x$, $y$, and $z$ velocity components. For ensemble generation, we vary two parameters based on the dataset authors' suggestions, such as the masses of the primary ($M_P$) and secondary ($M_S$) white dwarfs ($M_P, M_S \in [0.8, 0.95]\, M_\odot$), where $M_\odot$ represents the solar mass. A training subset of 400 timesteps is randomly sampled, with different ensemble members designated for training, validation, and testing.

We evaluate HyperFLINT's performance in reconstructing scalar and vector fields both qualitatively and quantitatively. For density field evaluation, we use *peak signal-to-noise ratio* (PSNR), while the accuracy of the flow field is assessed with *endpoint error* (EPE), which calculates the average Euclidean distance between estimated and GT flow vectors—lower EPE values indicate higher accuracy. For qualitative assessment, we visualize the predicted flow fields produced by our model for the simulation ensembles. Given the 3D nature of our datasets, PSNR and EPE are calculated in the volume domain, ensuring the metrics align with the spatial characteristics of the data and provide a meaningful and representative evaluation of HyperFLINT's performance.

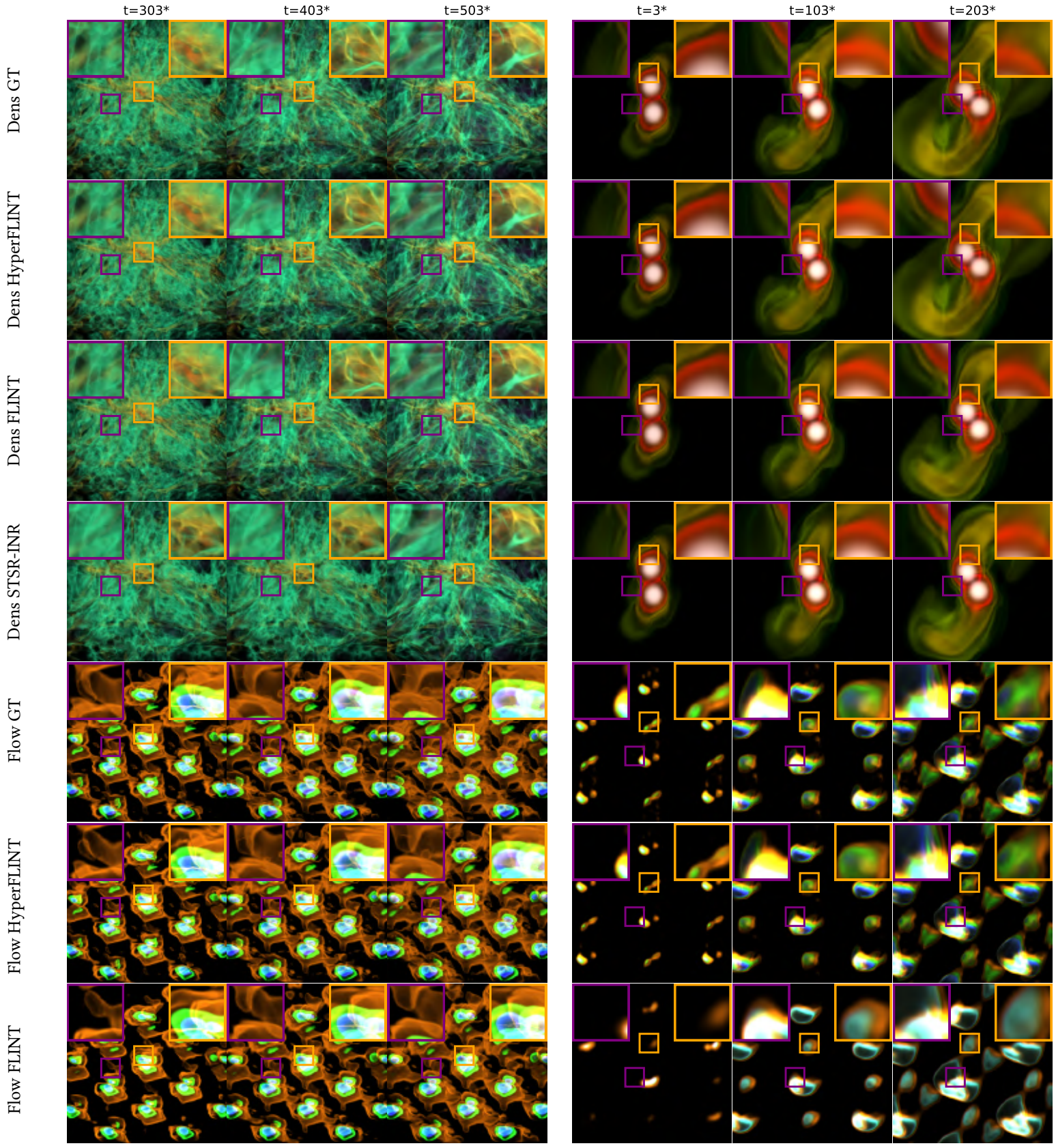

(a) Nyx results

(b) Castro results

Figure 5.4: Nyx and Castro: HyperFLINT flow field estimation and temporal density interpolation, 5×. From top to bottom, the rows show GT density, HyperFLINT interpolated density, FLINT interpolated density, STSR-INR interpolated density, GT flow, HyperFLINT flow estimation, and FLINT flow estimation. 3D rendering was used for the density and flow visualization (● ● ● colors representing $x$, $y$, and $z$ flow directions respectively). The corresponding transfer functions are shown in Section 5.5.3.

## 5.5 QUALITATIVE RESULTS

We evaluate HyperFLINT on two different datasets (Section 5.4.2) with respect to flow estimation and density interpolation (i.e., temporal super-resolution), and compare it with FLINT and STSR-INR.

### 5.5.1 *Nyx*

First, we consider the scenario where GT density and velocity fields are available for some members of the entire 3D ensemble. As Figure 5.4a illustrates, HyperFLINT achieves accurate performance in terms of both density field interpolation and flow field estimation. Visually, the difference between the renderings of the reconstructed density field (second row) and its GT is minimal. Moreover, even at a relatively high interpolation rate of 5×, HyperFLINT effectively learns a flow field that structurally resembles the GT flow. Both HyperFLINT and FLINT (third row) produce results that are visually very close to the GT for density interpolation; however, when compared to STSR-INR (fourth row), HyperFLINT demonstrates more accurate density field interpolation. At $t = 303$, $t = 403$, and $t = 503$, the purple and orange zoom-ins clearly reveal a different structure and less dark matter density in STSR-INR's results compared to HyperFLINT, which reconstructs density more effectively.

Moreover, HyperFLINT not only reconstructs the density field but also supplements it with accurate flow information, as shown in Figure 5.4a (sixth row), a feature that STSR-INR lacks. When examining the flow, we observe circular swirling patterns, indicating the complex dynamics of the baryonic gas. As these flows intensify, we see evidence of dark matter moving outward—reflected in both the GT and HyperFLINT density, especially in the orange zoom-ins—consistent with an expanding universe. This underlines the utility of HyperFLINT in capturing not just the static density fields but also the dynamic evolution of the cosmic structures. Furthermore, HyperFLINT outperforms the FLINT model (last row) in capturing flow information, producing more accurate representations of the underlying dynamics. For example, at $t = 303$ and $t = 503$, in the purple and orange zoom-ins, the FLINT flow exhibits structural differences compared to the GT, whereas HyperFLINT more effectively preserves the intricate flow dynamics.

A domain expert from astronomy specializing in cosmological simulations and observational data highlighted the advantages of estimating both density and velocity fields (similarly to Section 4.5.7). This capability is crucial in cosmology for estimating distances between astronomical objects and analyzing their spatial relationships. This would help in bridging simulations with real-world observations. Velocity estimation is particularly valuable for constructing “lightcones”, in which simulation data are used to model the evolution of the universe.

### 5.5.2 *Castro*

For the Castro ensemble dataset, as Figure 5.4b illustrates, HyperFLINT achieves results that are visually very close to the GT for both density field interpolation and flow field estimation. When comparing Hyper-FLINT to FLINT and STSR-INR for temporal density interpolation (third and fourth row), HyperFLINT demonstrates superior accuracy. For instance, at $t = 103$ and $t = 203$, STSR-INR and FLINT show noticeable deviations in structure and reduced matter density around the merging white dwarfs. In contrast, HyperFLINT preserves these density structures more effectively, aligning well with the GT. For flow estimation, HyperFLINT produces more coherent and spatially accurate flow fields than FLINT. This is particularly evident at $t = 3$ and $t = 203$ in Figure 5.4b, where FLINT introduces artifacts and inconsistencies in motion direction. In contrast, HyperFLINT better captures the developing flow dynamics, especially in the detailed areas in the purple and orange zoom-ins, where it maintains a more accurate representation of the evolving flows.

### 5.5.3 *Transfer Functions*

In Figure 5.5 we present transfer functions for visualizing the density and velocity components of the Castro ensemble. These transfer functions are applied to the density field and the $x$, $y$, and $z$ components of the flow field, offering an intuitive means to explore and analyze the spatial and dynamic characteristics of the simulation data. The utilized transfer functions help to better understand the structural patterns and flow dynamics inherent to each ensemble. For the Nyx ensemble, we use the same transfer functions as shown in Figure 4.19, Section 4.5.8.

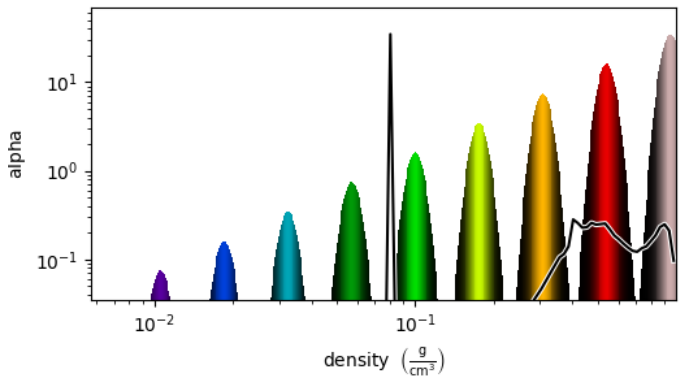

(a) TF for density field.

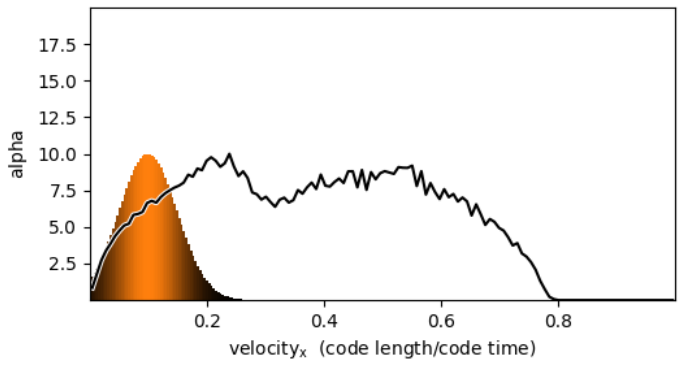

(b) TF for flow field in $x$ direction.

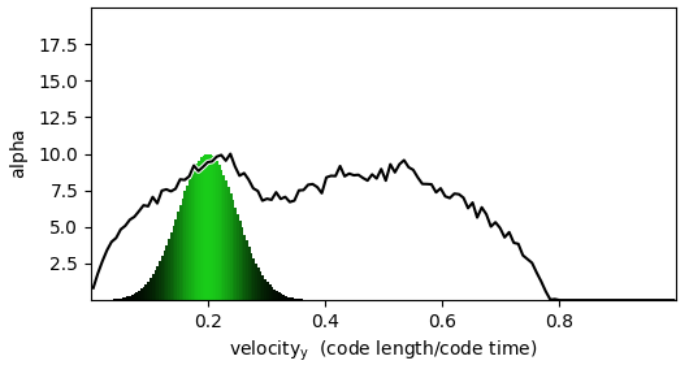

(c) TF for flow field in $y$ direction.

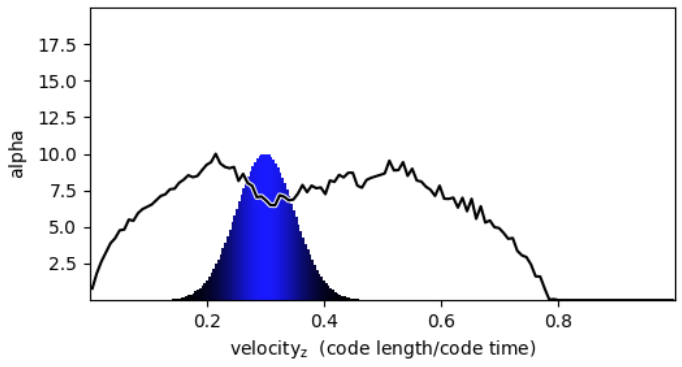

(d) TF for flow field in $z$ direction.

Figure 5.5: Castro ensemble: transfer function for density and $x$, $y$, $z$ components of the flow field.

## 5.6 QUANTITATIVE AND COMPARATIVE EVALUATION

In this section, we present quantitative results and compare against baseline methods to demonstrate the improvement achieved with our proposed method, followed by ablation and parameter studies to explore different configurations and hyperparameters.

### 5.6.1 *Comparison Against Baselines*

STSR-INR (Tang and Wang, 2024) and CoordNet (Han and Wang, 2022) were evaluated as key benchmarks for temporal super-resolution (TSR) tasks on the 3D+time Nyx and Castro datasets, with results shown in Table 5.1 and Table 5.2. STSR-INR employs a variational auto-decoder to optimize latent vectors for variable interpolation in the latent space, building on prior works like STNet (Han et al., 2021). It has demonstrated notable improvements in density interpolation but focuses solely on scalar field interpolation, lacking capabilities for flow field estimation. Similarly, CoordNet, a benchmark for visualizing time-varying volumetric data, also improves upon TSR-TVD (Han and Wang, 2019) as detailed in Section 5.2. While CoordNet achieves strong results in PSNR scores for density interpolation across various rates, it shares STSR-INR's limitation in addressing only TSR of scalar fields without extending to flow field estimation. In contrast, HyperFLINT not only outperforms both methods in density interpolation but also introduces flow estimation, a capability absent in both CoordNet and STSR-INR.

FLINT (Gadirov et al., 2025a), based on a pure CNN architecture with a student-teacher training mechanism, serves as another valuable benchmark for comparison. FLINT demonstrated improved results over STSR-INR and CoordNet in temporal interpolation and provided reasonable flow estimation, making it a strong competitor (Chapter 4). Our evaluation against FLINT shows that HyperFLINT slightly outperforms FLINT in terms of temporal super-resolution for scalar fields. More notably, HyperFLINT demonstrates superior performance in flow estimation, yielding significantly better results. This is evidenced by the lower EPE scores reported for both the Nyx and Castro datasets, as shown in Table 5.1 and Table 5.2. Furthermore, HyperFLINT offers the added advantage of enabling parameter space exploration, a feature neither FLINT nor other baselines can achieve. This makes HyperFLINT a more versatile and powerful tool for analyzing scientific ensembles.

**Inference time comparison.** We conducted an inference time comparison of HyperFLINT against FLINT, CoordNet, and STSR-INR across 3D ensemble datasets, measuring inference time over 300 timesteps from the test set. HyperFLINT significantly outperforms CoordNet and STSR-INR in speed, achieving an average of 0.18 seconds per timestep compared to 2.1 seconds for CoordNet and 1.5 seconds for STSR-INR. Remarkably, HyperFLINT even slightly outpaces FLINT, which averages 0.2 seconds per timestep. This improvement arises from HyperFLINT's efficient CNN utilization, which, despite the added hypernetwork complexity, maintains a performance advantage. Unlike INR-based methods that solve implicit functions at each point, HyperFLINT leverages CNNs' parallel processing capabilities and optimized memory access, resulting in faster inference.

Table 5.1: Comparison against baselines, Nyx ensemble

| Method | 3× | | 5× | | 8× | |
|---|---|---|---|---|---|---|
| | PSNR ↑ | EPE ↓ | PSNR ↑ | EPE ↓ | PSNR ↑ | EPE ↓ |
| HyperFLINT | **53.32** | **0.0237** | **52.70** | **0.0238** | **51.07** | **0.0242** |
| FLINT | 53.17 | 0.0310 | 52.31 | 0.0310 | 49.39 | 0.0309 |
| STSR-INR | 49.63 | — | 44.21 | — | 41.09 | — |
| CoordNet | 49.37 | — | 44.02 | — | 40.78 | — |
| Linear | 47.49 | — | 41.92 | — | 37.51 | — |

Table 5.2: Comparison against baselines, Castro ensemble

| Method | 3× | | 5× | | 8× | |
|---|---|---|---|---|---|---|
| | PSNR ↑ | EPE ↓ | PSNR ↑ | EPE ↓ | PSNR ↑ | EPE ↓ |
| HyperFLINT | **49.48** | **0.0275** | **47.39** | **0.0276** | **45.41** | **0.0278** |
| FLINT | 47.89 | 0.0502 | 46.16 | 0.0506 | 43.83 | 0.0513 |
| STSR-INR | 44.83 | — | 42.38 | — | 40.26 | — |
| CoordNet | 44.52 | — | 42.29 | — | 40.08 | — |
| Linear | 42.11 | — | 37.28 | — | 33.64 | — |

**Resource and efficiency comparison.** To further highlight the practical trade-offs of HyperFLINT compared to the baseline FLINT model, we summarize key computational metrics including model size, memory usage, training time, and hardware requirements in Table 5.3. While HyperFLINT incorporates a hypernetwork component to generate weights dynamically for several convolutional layers, it achieves a reduced parameter count and faster training due to its streamlined architecture, fewer convolutional blocks, and the removal of FLINT's teacher block used in its student-teacher setup. Although HyperFLINT requires more memory during both training and inference—due to the additional computation and storage needed to dynamically generate convolutional weights per timestep from the hypernetwork—the overall training time is shorter, and the number of trainable parameters is smaller. Both models were trained and evaluated on the same hardware (NVIDIA Titan V with 12 GB VRAM) for a fair comparison.

Table 5.3: Comparison between FLINT and HyperFLINT in terms of inference time, parameter count, memory usage, and training performance.

| Metric | FLINT | HyperFLINT |
|---|---|---|
| Inference time (s/timestep) | 0.20 | **0.18** |
| Trainable parameters | 9.8M | **7.9M** |
| Memory usage (inference) | **4.8 GB** | 6.6 GB |
| Memory usage (training) | **10.2 GB** | 11.2 GB |
| Total training time | ~12 h | **~8 h** |

### 5.6.2 *Ablation Studies*

Our proposed HyperFLINT method is a deep neural network that has various loss components. To assess the impact of these in achieving optimal results, we conducted a series of ablation studies.

Table 5.4: Ablation of HyperFLINT

| **Method** | **Nyx, 5×** | | **Castro, 5×** | |
|---|---|---|---|---|
| | **PSNR ↑** | **EPE ↓** | **PSNR ↑** | **EPE ↓** |
| HyperFLINT no *flow* | 51.94 | 0.1348 | 46.92 | 0.7432 |
| HyperFLINT no *rec* | 44.78 | 0.0254 | 37.89 | 0.0335 |
| HyperFLINT w/o hyper | 50.89 | 0.0357 | 46.04 | 0.0516 |
| HyperFLINT | **52.70** | **0.0238** | **47.39** | **0.0276** |

The ablation studies on HyperFLINT reveal the impact of various loss components on performance, as shown in Table 5.4 for the Nyx and Castro datasets. Four variants are compared:

- HyperFLINT no *flow*: omits the flow loss, leading to poor flow estimation (EPE) despite decent density interpolation (PSNR).
- HyperFLINT no *rec*: omits the reconstruction loss, resulting in decent flow learning but compromised density interpolation.
- HyperFLINT w/o hyper: removes the hypernetwork, leading to subpar interpolation and flow estimation due to the absence of dynamic adaptation enabled by simulation parameters.
- HyperFLINT: the full model with all components, achieves the best scores in both PSNR and EPE, outperforming all other variants.

### 5.6.3 *Hyperparameter Studies*

Our proposed HyperFLINT method involves several hyperparameters, and we performed extensive hyperparameter studies to understand their impact on achieving optimal results. Table 5.5 displays the outcomes of these studies regarding spatio-temporal datasets.

Table 5.5: Hyperparameter search for HyperFLINT

| **Method** | **Nyx, 5×** | | **Castro, 5×** | |
|---|---|---|---|---|
| | **PSNR ↑** | **EPE ↓** | **PSNR ↑** | **EPE ↓** |
| HyperFLINT *64* | 51.68 | 0.0315 | 47.12 | 0.0430 |
| HyperFLINT *Lapl* | 51.73 | 0.0319 | 44.95 | 0.0480 |
| HyperFLINT *hyper all* | 48.19 | 0.0281 | 42.79 | 0.0419 |
| HyperFLINT *w teacher* | 49.31 | 0.0265 | 44.43 | 0.0393 |
| HyperFLINT $\lambda_{flow} = 0.3$ | 52.29 | 0.0311 | 45.39 | 0.0316 |
| HyperFLINT $\lambda_{flow} = 0.1$ | 52.19 | 0.0320 | 47.01 | 0.0325 |
| HyperFLINT $\gamma = 0.9$ | 52.14 | 0.0310 | 46.98 | 0.0345 |
| HyperFLINT $\gamma = 0.7$ | 52.28 | 0.0317 | 47.07 | 0.0371 |
| HyperFLINT *stride = 1* | 52.19 | 0.0418 | 46.94 | 0.0348 |
| HyperFLINT *2 Blocks* | 49.72 | 0.0245 | 46.87 | 0.0407 |
| HyperFLINT *3 Blocks* | **52.70** | **0.0238** | **47.39** | **0.0276** |
| HyperFLINT *4 Blocks* | 52.51 | 0.0249 | 46.98 | 0.0314 |
| HyperFLINT *5 Blocks* | 52.09 | 0.0256 | 46.97 | 0.0341 |
| HyperFLINT *hyper no MLP* | 52.27 | 0.0261 | 46.06 | 0.0363 |
| HyperFLINT *hyper no CNN* | 51.74 | 0.0282 | 45.46 | 0.0378 |
| HyperFLINT *hyper no dropout* | 52.48 | 0.0259 | 46.61 | 0.0298 |

We conducted extensive hyperparameter optimization to identify the optimal configuration for HyperFLINT, testing variations in model width, depth, and loss functions. The best-performing setup, labeled "HyperFLINT *3 Blocks*", includes three blocks with convolutional layers having channel counts decreasing from 128 to 64, balancing capacity and avoiding overfitting. Alternatives like "HyperFLINT *hyper all*" (all layers affected by the hypernetwork) and "HyperFLINT *w teacher*" (using a teacher block similar to FLINT) resulted in convergence issues and degraded performance. Similarly, the "HyperFLINT *stride = 1*" variant underperformed due to reduced expressiveness from fixed strides.

We explored loss functions, comparing "HyperFLINT *Lapl*" using a Laplacian pyramid, and "HyperFLINT" which applies $L_1$ loss. The simpler $L_1$ loss proved more effective. Additionally, we show the roles of HyperNet components, as configurations like "HyperFLINT *hyper no MLP*," "HyperFLINT *hyper no CNN*," and "HyperFLINT *hyper no dropout*" exhibited reduced performance. These findings underscore the importance of the proposed architecture and individual components. In Table 5.6, we present the hyperparameter search results for Nyx at $64^3$ resolution, where "HyperFLINT *3 Blocks*" remains the best-performing model, consistent with what has been determined in Table 4.6 at $128^3$. This suggests good generalization across resolutions, with a slight performance drop at higher resolutions due to increased data complexity.

Table 5.6: Hyperparameter search for HyperFLINT, Nyx, 5×, $64^3$

| **Method** | **PSNR ↑** | **EPE ↓** |
|---|---|---|
| HyperFLINT *hyper all* | 48.87 | 0.0279 |
| HyperFLINT *3 Blocks* | **52.93** | **0.0220** |
| HyperFLINT *4 Blocks* | 52.81 | 0.0231 |
| HyperFLINT *hyper no MLP* | 52.33 | 0.0254 |
| HyperFLINT *hyper no CNN* | 51.65 | 0.0277 |

## 5.7 SIMULATION PARAMETER SPACE EXPLORATION

Our proposed HyperFLINT model achieves improved performance in flow estimation and density interpolation (Sections 5.5 and 5.6) while introducing novel capabilities for exploring simulation parameter spaces. Leveraging its HyperNet architecture, which dynamically conditions the FLINT* network on specific simulation parameters, HyperFLINT facilitates efficient analysis of how parameter variations influence simulation outcomes. This section highlights two key examples demonstrating HyperFLINT's utility in parameter space exploration. First, Section 5.7.1 examines the correlation between HyperNet-generated weights and simulation data, showcasing a strong alignment with simulation characteristics. Second, Section 5.7.2 demonstrates HyperFLINT's ability to interpolate within the parameter space, accurately predicting outputs for configurations beyond the training set. These functionalities are validated using experiments with the Nyx simulation ensemble.

### 5.7.1 *Hypernetwork weights as Proxy for Data Similarity*

To better understand the broader capabilities of HyperNet beyond generating weights for the FLINT* network, we explored its potential for parameter space analysis. Specifically, we constructed similarity matrices to examine how the weights generated by HyperNet correlate with the underlying data dynamics, reflecting the influence of simulation parameters. This analysis highlights how HyperNet weights, apart from serving as weight generation for the main network, can provide insights into parameter-driven variations within the dataset. Specifically, we compute similarity matrices for (i) the weights produced by HyperNet and (ii) the original dataset volumes (both in Figure 5.6), both of which are naturally influenced by the simulation parameters. To compute these matrices, we measure pairwise distances between samples using the Euclidean distance. For HyperNet, we flatten the generated output weights into vectors; for the raw data, we flatten the volumetric fields—specifically, the density and velocity components—into vectors as well. This enables a direct comparison of structural similarities driven by simulation parameters across both the learned rep-resentation space and the original data space. For this analysis, we considered 24 members from the Nyx ensemble under variation of $\Omega_m = [0.1025, 0.12, 0.1375, 0.155, 0.1725, 0.19]$, $h = [0.55, 0.6, 0.65, 0.7]$, while $\Omega_b$ was fixed at 0.0225. Our proposed HyperFLINT model was trained on 25% of the data, specifically on six members with the combinations of parameters $\Omega_m = [0.1025, 0.1375, 0.19]$, $h = [0.55, 0.7]$, with $\Omega_b = 0.0225$.

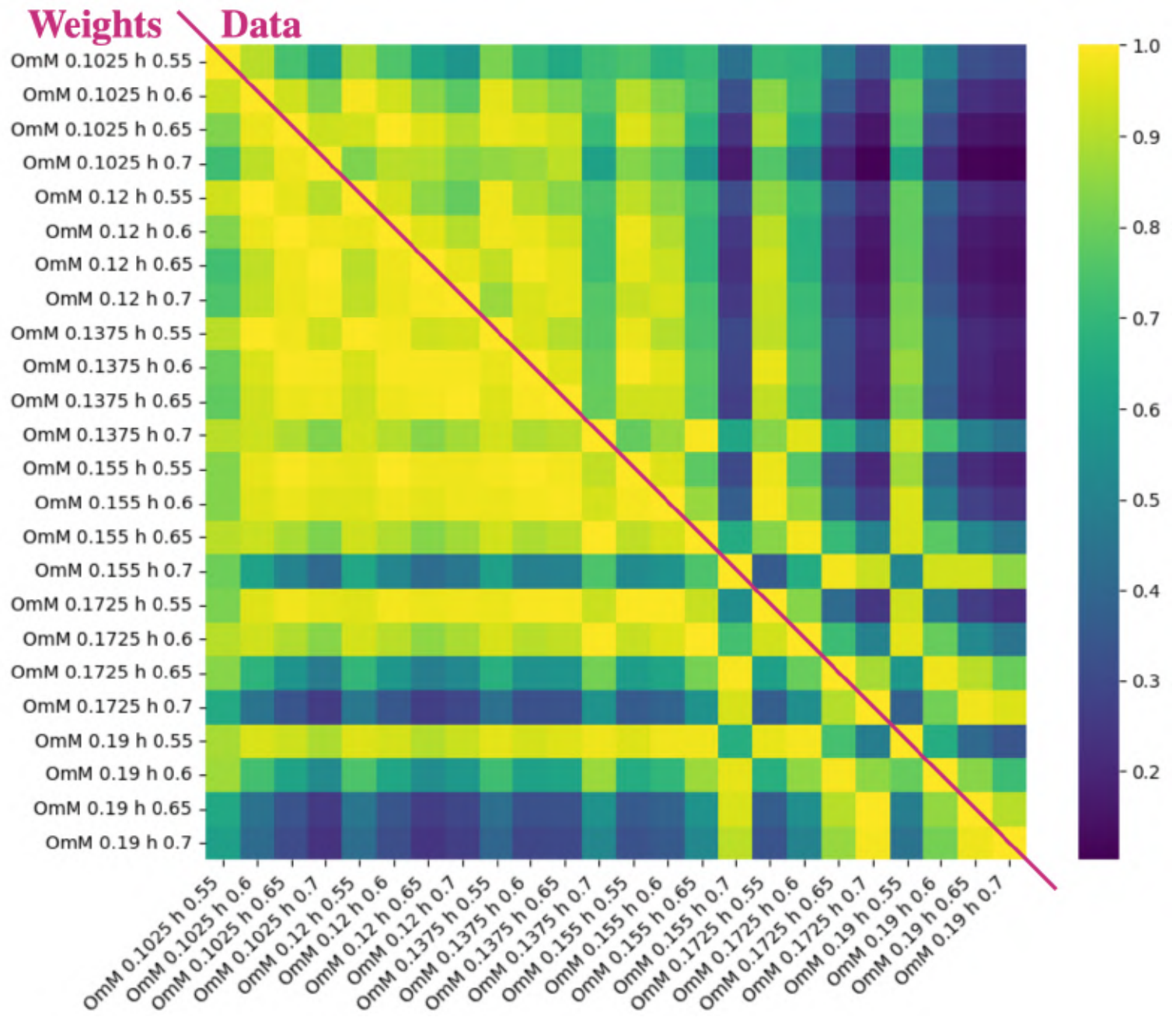

Figure 5.6: Similarity matrix: lower-left—HyperNet's weight similarity, upper-right—Nyx simulation parameter similarity. The diagonal line marks the separation between the two matrices.

The similarity matrices reveal a strong correlation between the HyperNet-generated weights and the dataset volumes, confirming that the HyperNet effectively learns parameter-dependent representations. This correlation suggests that differences in HyperNet weights can serve as a meaningful surrogate for differences in the underlying data. Upon closer examination, certain members stand out with distinguishable characteristics. For instance, for parameter configurations such as $\Omega_m = 0.155$ and $h = 0.7$, and $\Omega_m = 0.1725$ with $h = 0.65$ or $h = 0.7$, the data shows lower similarity compared to other ensemble members. Similarly, for $\Omega_m = 0.19$ with $h = 0.6$, $h = 0.65$, and $h = 0.7$, the matrix highlights notable deviations in the data structure. Despite these variations, HyperNet's weight similarities maintain strong correlations across all configurations, enabling analysis of how simulation parame-ters influence the underlying data dynamics without requiring reconstruction or access to the full volumetric fields (e.g., density or velocity). We also conducted a quantitative evaluation using a *triplet loss* approach (Schultz and Joachims, 2003) to validate the alignment between HyperNet-generated weights and parameter-dependent data dynamics. We analyzed triplets comprising an anchor, a more similar member, and a less similar member to evaluate how well distances in the HyperNet-generated weight space align with those in the data space. Specifically, we computed the distance between the anchor and both the more similar and less similar members in the data space and performed the same calculation in the HyperNet-generated weight space, confirming that weights similarity strongly correlates with data similarity. In both cases, distances were measured using the Euclidean norm between flattened vectors: for the data space, we used flattened volumetric fields; for the HyperNet space, we used the flattened convolutional weights generated for each parameter set. Achieving a 96% triplet correlation, this result underscores HyperFLINT's utility in parameter space analysis, where learned representations guide tasks like constructing characteristic maps of parameter impacts and optimizing sampling strategies.

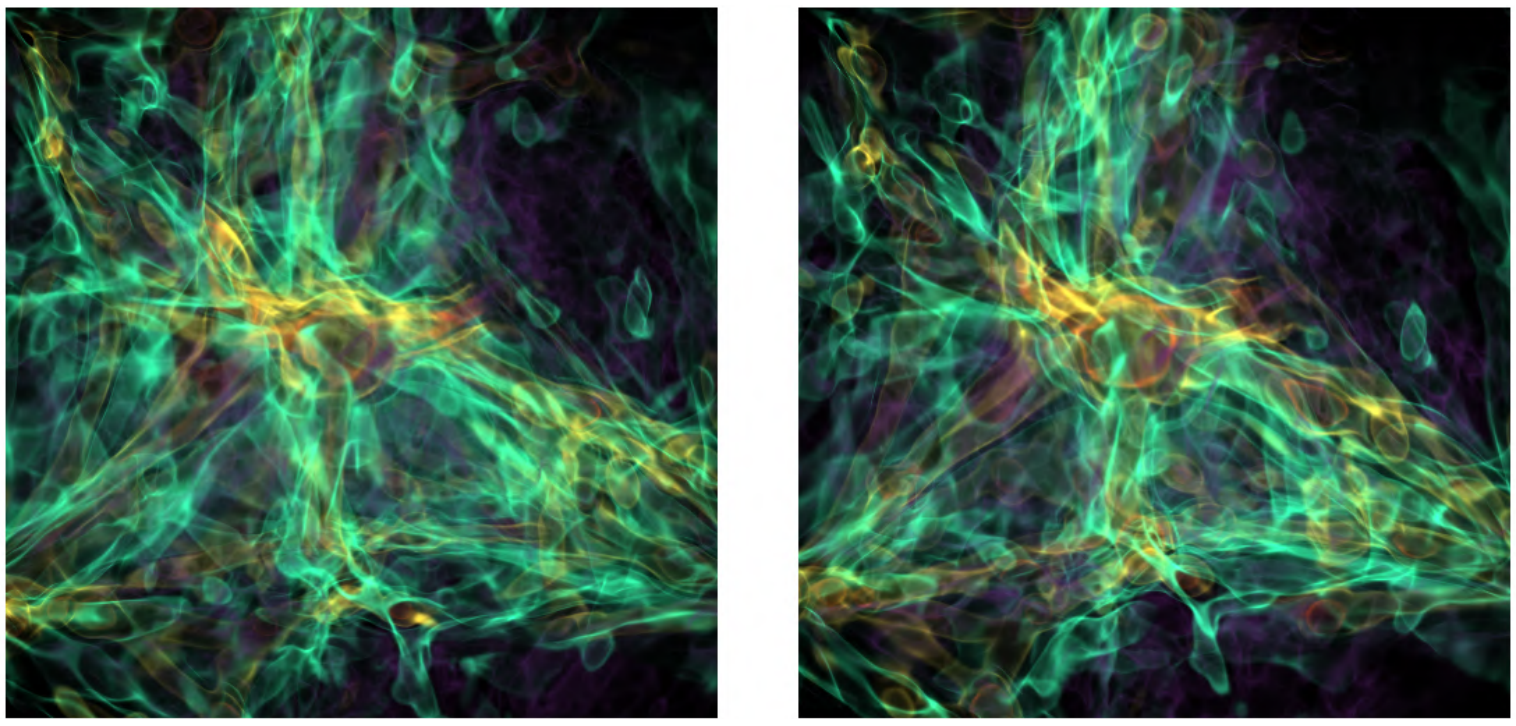

Figure 5.7: The rendering of two volumes that were used for Nyx simulation parameter space exploration.

### 5.7.2 *Parameter-Driven Data Synthesis*

Leveraging the HyperNet architecture, which conditions the FLINT* network on specific simulation parameters, HyperFLINT can dynamically adapt its outputs based on changes in simulation settings. As shown in Figure 5.7, which presents renderings of the input volumes at timesteps $s$ and $u$, these volumes serve as the foundation for interpolating the target timestep while varying only the simulation parameters through HyperNet. As illustrated in Figure 5.8, this approach effectively generates outputs that closely match the ensemble volumes associated with the specified parameter values. Even when starting with identical initial data, HyperFLINT adequately adapts its predictions solely based on the altered parameter inputs. For instance, the density reconstruction and flow estimation are particularly accurate for the parameter setting $\Omega_m = 0.1375$ (third column), corresponding to the data from the ensemble member used as inputs. Notably, HyperFLINT maintains the structural integrity and fine-grained details of the density field (second row) while accurately capturing the dynamic flow patterns (fifth row), demonstrating its capability to preserve both shape and texture. By using the same input volume and altering only the simulation parameters, HyperFLINT generates new data outputs effectively for settings like $\Omega_m = 0.12$, 0.155, and 0.1725, achieving visually close reconstructions, as the difference plots confirm, showcasing its generaliza-tion capabilities. While novel (unseen) simulation features cannot be generated, these results suggest that the model is capable of capturing parameter-dependent relationships, as reflected in the smooth density transitions and flow field predictions that generally follow GT dynamic trends. The reconstruction remains robust despite these novel parameters. As expected, we note a slight increase in error for more distant parameter values such as $\Omega_m = 0.1025$ and 0.19 (first and last columns, respectively). This is likely due to these configurations being further away from the range of parameters the model received as input during inference. Nonetheless, the results remain decent, with distinct features preserved, such as central galaxy density formation and distinct flow patterns, showcasing HyperFLINT's parameter-dependent generalization capabilities.

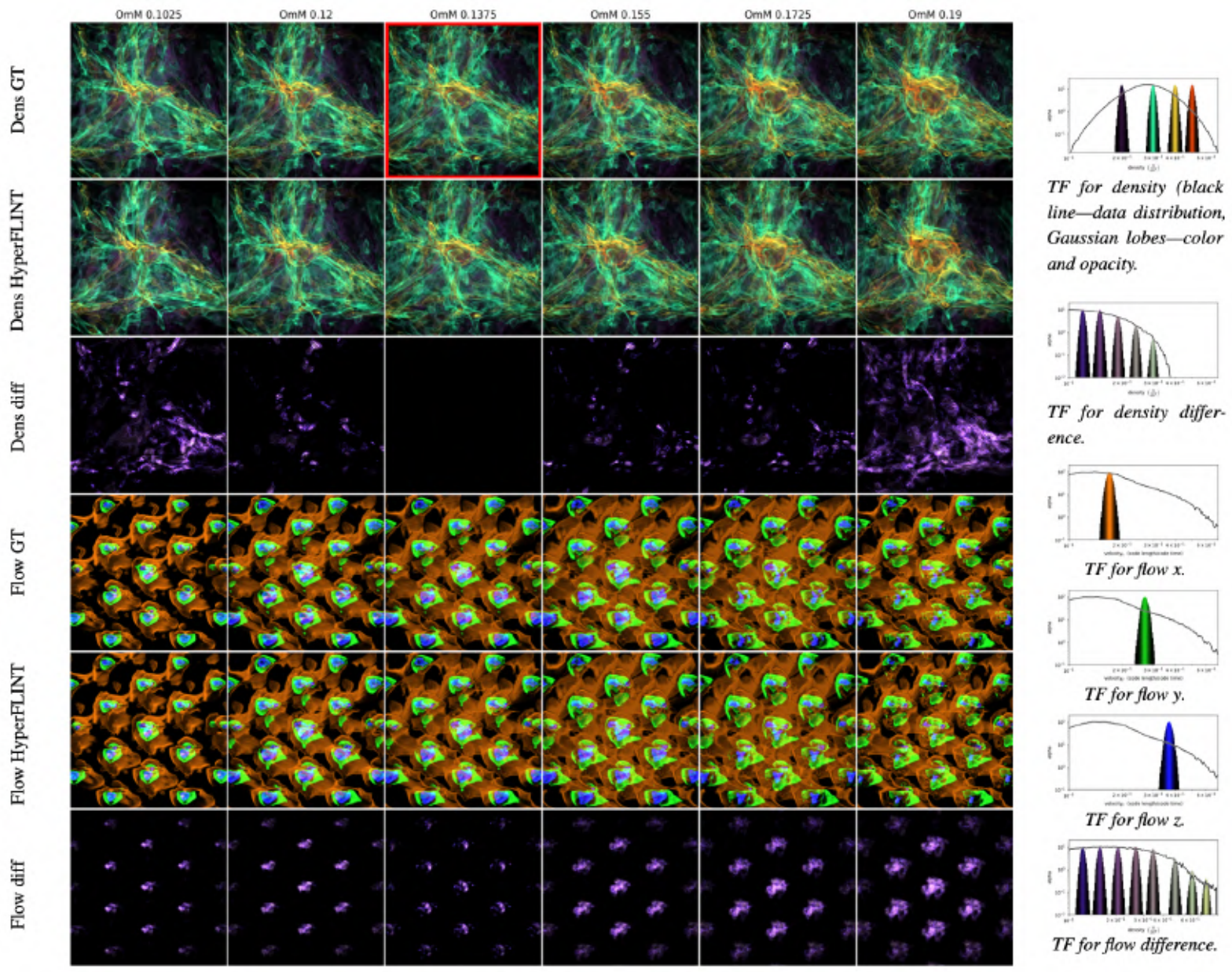

Figure 5.8: Nyx simulation parameter space exploration and transfer functions for density and flow field components. From top to bottom, the rows show GT density, HyperFLINT interpolated density, difference between GT and HyperFLINT density, GT flow, HyperFLINT flow estimation, and difference between GT and HyperFLINT flow. On the right, transfer functions are shown for the density, flow, and their respective difference fields. The volume marked with a red outline (first row, third column) highlights the input volume used for interpolation, which remains fixed during inference. However, the associated input simulation parameters vary, enabling an exploration of how HyperFLINT responds to different parameter configurations within the learned space.

## 5.8 CONCLUSION AND FUTURE WORK

We proposed HyperFLINT, a hypernetwork-based method for flow estimation and scalar field interpolation in spatio-temporal scientific ensembles. By leveraging hypernetworks, HyperFLINT adapts to varying simulation parameters, enabling accurate flow estimation and high-quality temporal interpolants, even with limited data. It outperforms recent state-of-the-art methods while achieving fast inference without requiring extensive pre-training or fine-tuning on simplified datasets. Validated across diverse simulation ensembles, HyperFLINT demonstrates robust and versatile performance, making it a valuable tool for scientific visualization.

The integration of hypernetworks in HyperFLINT establishes a robust framework for analyzing how simulation parameters influence flow and density dynamics, enabling efficient parameter space exploration and deeper insights into complex systems. Integrating skill score metrics that account for spatial and temporal biases, along with adaptivity based on flow characteristics and simulation constraints, could enhance robustness and applicability in future work. Incorporating stable diffusion techniques (Rombach et al., 2022) could further enhance HyperFLINT's ability to generate high-quality vector and scalar fields by leveraging simulation parameters. Known for producing smooth and coherent outputs, stable diffusion methods could complement hypernetworks by refining details and improving generalization in intricate simulations. In particular, their denoising process is well-suited for recovering complex spatial patterns from sparse or noisy inputs, a common scenario in scientific simulations. Moreover, conditioning diffusion models on simulation parameters could allow controllable generation across the parameter space, potentially outperforming deterministic networks in capturing multi-modal behaviors or uncertainty. While diffusion models typically produce a single output per run, each result is a sample drawn from a learned distribution conditioned on simulation parameters. This makes them particularly valuable for systems where the same parameters can yield multiple plausible outcomes—such as turbulent flows or phase transitions—by enabling the generation of diverse outputs through repeated sampling. This contrasts with deterministic models, which tend to collapse to the mean of all plausible behaviors. Additionally, this probabilistic nature supports uncertainty quantification and offers a way to explore variability in data-rich but computationally expensive domains, thereby enhancing model expressiveness and robustness in complex scientific applications.

# 6 CONCLUSION

*This final chapter concludes the thesis by summarizing its key contributions, acknowledging current limitations, and outlining directions for future research. It revisits the progression of novel deep learning methods developed for the analysis of spatiotemporal scientific ensemble data—beginning with autoencoder-based dimensionality reduction, advancing through the FLINT framework for simultaneous flow estimation and temporal interpolation, and culminating in HyperFLINT, an ensemble parameter-aware model powered by hypernetworks. The chapter also reflects on the constraints of these methods and explores future opportunities, including the integration of foundation models, large language models, and advanced generative techniques to create more robust, scalable, and interpretable tools for scientific visualization.*

## 6.1 SUMMARY OF RESEARCH RESULTS AND CONTRIBUTIONS

This dissertation presents novel methodologies for analyzing and visualizing spatio-temporal scientific ensembles using deep learning. Our work builds progressively from fundamental representation learning to more advanced adaptive models, addressing key challenges in dimensionality reduction, flow estimation, and simulation parameter-aware adaptation. We systematically tackle the core research questions introduced in Chapter 1, which we now revisit in the light of our findings.

**Autoencoder-Based Dimensionality Reduction (Chapter 3).** We begin by investigating the role of autoencoders in reducing the dimensionality of ensemble data while preserving essential spatial and temporal structures (**RQ1**). Through a comprehensive study of various autoencoder architectures, we assess their effectiveness in capturing meaningful low-dimensional representations of scientific fields, surpassing traditional dimensionality reduction methods (**RQ4**). We identify trade-offs between compression and reconstruction quality and propose a Pareto-efficient selection strategy based on partially labeled data, ensuring robust and interpretable embeddings. This step establishes the foundation for subsequent tasks by providing a compact yet expressive representation of spatio-temporal ensembles.

**FLINT: Flow Estimation and Temporal Interpolation (Chapter 4).** Building upon the learned representations from autoencoders, we introduced FLINT, a deep learning framework for estimating flow fields and performing high-quality temporal interpolation in scientific ensembles. FLINT is designed to uncover underlying patterns in ensemble experi-ments and simulations (**RQ4**) by directly learning implicit flow dynamics from spatio-temporal data. It addresses both flow-supervised and flow-unsupervised cases, reconstructing missing frames in 2D+time and 3D+time datasets without requiring explicit domain knowledge (**RQ2**). By doing so, FLINT demonstrates that deep learning can capture underlying motion patterns and trends beyond simple interpolation (**RQ4**). This method bridges the gap between static representations and dynamic evolution, enabling more accurate reconstructions of evolving scientific fields.

**HyperFLINT: Hypernetwork-Based Flow Estimation and Interpolation (Chapter 5)**. While FLINT achieves high-quality interpolation, it lacks adaptability to varying simulation conditions. To address this, we introduced HyperFLINT, a hypernetwork-based extension that enables parameter-aware learning and dynamic model adaptation (**RQ3**). By conditioning on simulation parameters, HyperFLINT generates flow fields and interpolations that are more physically consistent across different parameter regimes, expanding the model's generalizability. This approach provides insights into how learned representations evolve with changes in simulation parameters and allows for more flexible scientific exploration. Furthermore, through the use of hypernetworks, HyperFLINT improves both flow estimation and temporal interpolation, achieving significantly more accurate reconstructions compared to conventional techniques such as linear interpolation (**RQ4**). This enables more generalizable and consistent results across diverse scientific domains.

**Contributions and Impact**. The methodologies developed in this dissertation advance machine learning-based ensemble visualization across multiple directions. By integrating insights from dimensionality reduction, flow estimation, temporal super-resolution, and hypernetwork-based adaptation, this work outlines a cohesive set of machine learning strategies that enable high-quality feature extraction, interpolation, reconstruction, and parameter-aware modeling of spatiotemporal scientific data. Our work highlights the potential of deep learning models to adapt to varying simulation and experimental conditions while preserving essential physical structures. These contributions establish a foundation for scalable and robust machine learning techniques in scientific visualization, enabling efficient analysis of large-scale simulation and experimental datasets.

## 6.2 DISCUSSION

This section discusses the limitations of our proposed methods, covering autoencoder-based dimensionality reduction, flow estimation and interpolation with FLINT, and simulation hypernetwork-based parameter-aware reconstruction with HyperFLINT, along with possible solutions and future directions to address these challenges.

### 6.2.1 *Autoencoder-Based Dimensionality Reduction*

Autoencoders have demonstrated strong performance in reducing the dimensionality of high-dimensional ensemble datasets. However, their effectiveness heavily depends on the characteristics of the dataset. For instance, in Markov Chain Monte Carlo (MCMC) ensembles with a high proportion of relevant elements, autoencoders yield a significant improvement over the baseline projection method, UMAP. In contrast, for datasets like Drop Dynamics, where most pixels represent background, the advantage of autoencoders diminishes. Enhancing robustness in such cases requires careful data preprocessing, such as noise reduction or segmentation, to remove unimportant features before training. Additionally, selecting the optimal value of the Lagrangian multiplier $\beta$ in $\beta$-VAEs remains a challenge, as a fixed value does not always yield the best trade-off between reconstruction accuracy and disentanglement. A potential solution is to gradually increase the $\beta$-weighted KL term during training to balance feature separation and reconstruction quality.

### 6.2.2 *FLINT: Flow Estimation and Temporal Interpolation*

FLINT provides high-quality flow estimation and scalar field interpolation, even when flow information is entirely missing. However, at high interpolation rates (e.g., 32×), the accuracy of flow estimation degrades compared to lower interpolation rates. Although the method preserves large-scale structures such as dark matter distributions in astronomical datasets, finer details become less accurate at extreme interpolation rates. Another limitation is that FLINT cannot fully reconstruct missing ensemble members when only two distant timesteps are provided (e.g., beginning and end). Incorporating additional temporal priors or multi-scale feature learning could mitigate this issue and improve interpolation quality for highly sparse datasets.

### 6.2.3 *HyperFLINT: Hypernetwork-Based Flow Estimation and Interpolation*

HyperFLINT extends FLINT by dynamically incorporating ensemble parameters through hypernetworks, enabling simulation parameter-aware flow estimation and scalar field interpolation. However, similar to FLINT, its performance deteriorates at large interpolation rates due to increasing temporal gaps between reference timesteps. Additionally, while HyperFLINT effectively models parameter dependencies, more research is needed to fully explore parameter space variations, particu-larly for extrapolation beyond the observed simulation conditions. The current model is limited in its ability to generalize to unseen parameter ranges when given only a few reference time steps. Further improvements could involve integrating stable diffusion techniques to enhance the generation of high-quality vector and scalar fields or introducing uncertainty quantification mechanisms to assess prediction reliability in parameter space exploration.

## 6.3 DIRECTIONS FOR FUTURE WORK

Building on the methodologies introduced in this dissertation, several promising directions emerge for advancing machine learning-based techniques for spatio-temporal ensemble analysis. These include refining dimensionality reduction strategies, expanding flow estimation and interpolation techniques, and integrating recent developments in generative modeling, large language models (LLMs), and vision language models (VLMs).

**Enhancing Autoencoder-Based Dimensionality Reduction**. While this work explores autoencoder architectures for spatial ensemble projections, the increasing prevalence of foundation models for representation learning suggests new opportunities for improving dimensionality reduction. Instead of training autoencoders from scratch for each dataset, future work could investigate pre-trained vision transformers (ViTs) (Dosovitskiy et al., 2021) and contrastive learning approaches (e.g., CLIP (Radford et al., 2021)) to extract features in a more generalizable manner. Furthermore, integrating self-supervised learning techniques, such as masked autoencoders (MAE) (He et al., 2022), could enable better feature extraction even in cases with limited labeled data. However, creating such universally applicable foundation models for scientific datasets remains highly challenging, as the diversity across domains (e.g., medical imaging, fluid dynamics, climate data) makes it difficult to establish a single representation space comparable to natural language models.

Another direction involves leveraging diffusion models for dimensionality reduction. Given their ability to generate high-quality structured representations (Rombach et al., 2022), diffusion-based latent space learning could provide new ways to improve projection stability and disentanglement. Additionally, exploring transformer-based autoencoders could offer more scalable alternatives to convolution-based feature extraction.

**Extending HyperFLINT for Flow Estimation and Interpolation**. HyperFLINT introduces a hypernetwork-based approach for flow estimation and scalar field interpolation, but several enhancements are possible. One key limitation of current methods is their reliance on explicit numerical simulation data. Given the rapid advancement of neu-ral physics models (Shukla et al., 2021), future work could investigate physics-informed neural networks (PINNs) to integrate domain knowledge directly into flow estimation. By enforcing physical constraints, such models could improve interpolation robustness, particularly in data-scarce regimes.

Additionally, transformer architectures have proven highly effective for long-range dependencies in spatio-temporal data (Vaswani et al., 2017). Replacing convolutional layers in HyperFLINT with spatio-temporal transformers could improve its ability to model complex flow patterns. Furthermore, extending HyperFLINT to autoregressively predict future states (extrapolation) rather than solely focusing on interpolation could enhance its usability for forecasting applications.

**Leveraging LLMs and VLMs for Scientific Data Analysis**. An emerging question in the field is whether LLMs and VLMs can contribute to numerical simulation analysis. While LLMs have primarily been used for text-based applications, recent advances show that they can be adapted for structured data interpretation, including generating and modifying simulation code (Chen et al., 2021). VLMs, which jointly process visual and textual inputs, open new avenues for multimodal interaction with scientific data—e.g., understanding visual trends in simulations via textual queries or aiding model evaluation through visual descriptions.

Future research could explore using LLMs and VLMs to:

1. *Learn implicit simulation priors:* Fine-tuning LLMs or VLMs on scientific datasets may enable them to generate missing simulation frames or summaries by learning high-level patterns from both metadata and visual outputs.

2. *Facilitate adaptive model selection:* These models could assist in choosing optimal architectures for autoencoders or flow models based on dataset characteristics, leveraging techniques like neural architecture search (NAS) guided by textual and visual cues.

3. *Automate parameter tuning and visual diagnostics:* LLMs and VLMs could serve as interactive assistants, recommending hyperparameter adjustments or visually inspecting simulation results, based on learned heuristics from large-scale datasets.

While these applications are still in their early stages, integrating LLMs and VLMs into ensemble data analysis pipelines could significantly enhance interpretability, interactivity, and automation.

**Integration with Generative Models and Diffusion Techniques**. Recent advances in generative models, particularly diffusion models, open up new avenues for ensemble data augmentation. Given their ability to generate realistic high-dimensional outputs, diffusion models could complement HyperFLINT by enhancing the quality of interpolated fields and reducing artifacts in scenarios with limited training data. A key challenge in ensemble interpolation is maintaining temporal coherence—an area where diffusion models excel (Ho et al., 2020). Future work could explore hybrid approaches, where diffusion models refine flow predictions generated by HyperFLINT.

Moreover, generative adversarial networks (GANs) and normalizing flows (Papamakarios et al., 2021) could be used for uncertainty quantification by generating diverse plausible ensemble members. This could enhance robustness in applications where missing data is a challenge, such as climate modeling and astrophysics simulations.

**Real-Time Optimization and Scalability**. As scientific datasets continue to grow, optimizing the efficiency of deep learning models remains a priority. Techniques such as mixed-precision training (Micikevicius et al., 2018; Pandey et al., 2023), low-rank adaptation (LoRA) (Hu et al., 2022b; Dettmers et al., 2023), and model pruning (Han et al., 2016; Cheng et al., 2024) help reduce computational overhead while maintaining performance. Frameworks like vLLM (Rohaninejad et al., 2023) and DeepSpeed (Rasley et al., 2020) offer system-level solutions for efficient inference and training at scale. Switch Transformers (Fedus et al., 2022) further demonstrate the utility of sparse expert models for reducing computation.

Another key avenue is exploring reinforcement learning for adaptive model refinement—allowing models to adjust their complexity dynamically based on available computational resources. Additionally, distributed learning strategies such as federated learning (McMahan et al., 2017) or ZeRO (Rajbhandari et al., 2020) improve the scalability of ensemble analysis. These strategies are particularly beneficial for collaborations across institutions handling large-scale simulation datasets.

## 6.4 FINAL THOUGHTS

At the heart of this work lies a deep curiosity about the nature of reality itself. I have always been fascinated by the idea that the world we perceive—this four-dimensional space-time continuum—may only be a projection of a much higher-dimensional space, a richer, $n$-dimensional structure that our senses and instruments can only partially grasp. Through this lens, scientific visualization and artificial intelligence becomes more than a tool; it becomes a way to explore glimpses of that deeper structure, to simulate and interpret the traces of something fundamentally beyond us. While so much remains unknown—and perhaps unknowable—I believe we are not powerless in the face of that mystery. In fact, I am deeply convinced that the most meaningful experiment we can conduct is to fully engage with the unique space and time we are given, to explore, learn, and create with intention. This work represents my attempt to do just that: to use the tools of science, computation, and visualization to better understand the world we inhabit—and the one that may lie beyond.

# ACKNOWLEDGMENTS

*"I am glad you are here with me. Here at the end of all things."*

— J.R.R. Tolkien, *The Lord of the Rings: The Return of the King*

As I reach the final pages of this journey, I wish to express my deepest gratitude to those who have walked this long road with me. Like the Fellowship that set forth from Rivendell, my PhD was not a path taken alone. Many companions—mentors, colleagues, friends, and family—have stood by me, offering wisdom, encouragement, and strength when the way was unclear.

First and foremost, I would like to express my deepest gratitude to Dr. Steffen Frey for inviting me to pursue a PhD at the University of Groningen, setting the foundation for this incredible journey. His insightful discussions, continuous support, and valuable guidance have been instrumental throughout my research.

I am also sincerely grateful to my advisor, Professor Jos Roerdink, whose high-level suggestions on research direction, paper writing, and academic rigor have significantly shaped my work. His expertise and thoughtful feedback have been invaluable in refining my ideas and strengthening my contributions.

Furthermore, I extend my heartfelt thanks to Professor Kwan-Liu Ma from the University of California, Davis, for enabling my collaboration at his VIDI research group. This opportunity allowed me to engage in a highly productive and enriching academic exchange, which led to a rewarding collaboration and a published paper.

I would also like to express my appreciation to Professor Jiri Kosinka, with whom I had the pleasure of collaborating on one of my projects. His critical insights and constructive suggestions played a key role in advancing the research. Additionally, his efforts in organizing activities for our SVCG research group brought a great sense of community and made the experience even more enjoyable.

I thank the Center for Information Technology of the University of Groningen for their support and for providing access to the Hábrók high performance computing cluster. I also thank the Deutsche Forschungsgemeinschaft (DFG, German Research Foundation) for providing support by funding SFB 1313 (Project Number 327154368). I thank Dr. Maxime Trebitsch (Institut d'Astrophysique de Paris) for his help in interpreting the flow visualizations of the Nyx cosmological simulation and assessing their usefulness for data analysis.

Thanks to my colleagues from the SVCG research group: Sjoerd, Zikai, Dennis, Kuanhao, Heejun, Youngjoo, Yanrui, and Jiamin. Also to Dr. Cara Tursun and Dr. Christian Kehl. I would also like to thank my colleagues from the University of California, Davis—David and Qi. I truly enjoyed my time with you.

To my friends—the mountain squad: Kolya, Zhenya, Pasha, and Nikita—our mountaineering and skiing adventures during the holidays made this PhD journey unforgettable. To my climbing mate, Hylke, for belaying me securely all these years. Warm thanks to friends across the world: Lukas, Kazuhiro, Aidar, for the support and connection despite the distance. And definitely to my longtime friends from school and university in Baku—Ali and Shamo—thank you for always being there.

Lastly, to my family and friends—your unwavering support, love, and encouragement have carried me through every challenge. I am forever grateful for your belief in me.

*With friends surrounded*
*The dawn mist glowing*
*The water flowing*
*The endless river*
*Forever and ever*

— Pink Floyd, *High Hopes, The Division Bell*

# SHORT RÉSUMÉ

## HAMID GADIROV

### PERSONAL DATA

**Name:** Hamid Gadirov
**Email:** gadirovh@gmail.com

### EDUCATION

| | |
|---|---|
| Mar 2021 – Feb 2025 | **University of Groningen, Groningen, the Netherlands**<br>Bernoulli Institute for Mathematics, Computer Science, and AI<br>Ph.D. in Computer Science |
| Oct 2017 – Sep 2020 | **University of Stuttgart, Stuttgart, Germany**<br>M.Sc. in Computer Science |
| Sep 2013 – Jul 2017 | **National Aviation Academy, Baku, Azerbaijan**<br>B.Sc. in Computer Engineering |

### PUBLICATIONS

**H. Gadirov**, G. Tkachev, T. Ertl, and S. Frey. *Evaluation and Selection of Autoencoders for Expressive Dimensionality Reduction of Spatial Ensembles. Int. Symposium on Visual Computing, Springer (2021), pp. 222–234.*

**H. Gadirov**, J. B. T. M. Roerdink, and S. Frey. *FLINT: Learning-based Flow Estimation and Temporal Interpolation for Scientific Ensemble Visualization. IEEE Trans. on Visualization and Computer Graphics (2025).*

**H. Gadirov**, Qi Wu, D. Bauer, K.-L. Ma, J. B. T. M. Roerdink, and S. Frey. *HyperFLINT: Hypernetwork-based Flow Estimation and Temporal Interpolation for Scientific Ensemble Visualization. CGF (2025).*

D. Bauer, Qi Wu, **H. Gadirov**, and K.-L. Ma. *GSCache: Real-Time Radiance Caching for Volume Path Tracing using 3D Gaussian Splatting. IEEE Trans. on Visualization and Computer Graphics* (2025).

Z. Yin, **H. Gadirov**, J. Kosinka, and S. Frey. *ENTIRE: Learning-based Volume Rendering Time Prediction. arXiv preprint arXiv:2501.2501 (2025).*

## COLOPHON

This document was typeset using the typographical look-and-feel `classicthesis` developed by André Miede. The style was inspired by Robert Bringhurst's seminal book on typography "*The Elements of Typographic Style*". `classicthesis` is available for both LaTeX and LyX:

http://code.google.com/p/classicthesis/

*Final Version* as of November 27, 2025 (`classicthesis`).